%% file: main.tex
\RequirePackage{fix-cm}
\documentclass[smallcondensed]{svjour3}     
\smartqed  
\usepackage{graphicx}
\usepackage{latexsym}
\usepackage{amsmath,amsfonts,amssymb}
\usepackage{float}

\usepackage[font=small,skip=0pt]{caption}
\usepackage{caption}
\usepackage{subcaption}
\usepackage{graphbox}
\usepackage{makecell}
\PassOptionsToPackage{hyphens}{url}\usepackage{hyperref}
\begin{document}

\title{Adversarial Canonical Correlation Analysis
}


\author{Benjamin Dutton}


\institute{Benjamin Dutton \at
              North Carolina State University \\
              \email{bcdutton@ncsu.edu} 
}


\maketitle

\begin{abstract}
Canonical Correlation Analysis (CCA) is a statistical technique used to extract common information from multiple data sources (views).  It has been used in various representation learning problems, such as dimensionality reduction, word embedding, and clustering.  Recent work has given CCA probabilistic footing in a deep learning context and uses a variational lower bound for the data log likelihood to estimate model parameters.  Alternatively, adversarial techniques have arisen in recent years as a powerful alternative to variational Bayesian methods in autoencoders.  In this work, we explore straightforward adversarial alternatives to recent work in Deep Variational CCA (VCCA and VCCA-Private) we call ACCA and ACCA-Private and show how these approaches offer a stronger and more flexible way to match the approximate posteriors coming from encoders to much larger classes of priors than the VCCA and VCCA-Private models.  This allows new priors for what constitutes a good representation, such as disentangling underlying factors of variation, to be more directly pursued.  We offer further analysis on the multi-level disentangling properties of VCCA-Private and ACCA-Private through the use of a newly designed dataset we call Tangled MNIST.  We also design a validation criteria for these models that is theoretically grounded, task-agnostic, and works well in practice.  Lastly, we fill a minor research gap by deriving an additional variational lower bound for VCCA that allows the representation to use view-specific information from both input views.
\end{abstract}

\input{sections/introduction.tex}
\input{sections/related_work.tex}

\input{sections/method.tex}
\input{sections/dataset.tex}
\input{sections/experiments.tex}

\input{sections/conclusion.tex}

\bibliographystyle{spmpsci}      
\bibliography{ecml}   

\input{sections/appendix.tex}

\end{document}

%% file: sections/introduction.tex
\section{Introduction}
\label{intro}
In multi-view learning (MVL) problems, multiple data sources or \textit{views} are available at training time and the assumption is made that there is a high degree of information overlap between them.  These data sources can be artificially induced by simply partitioning features into sets or can correspond to distinct physical, real world sensors monitoring a common information source \cite{sun2019multiview}, such as multiple cameras pointing at a common object.  MVL is a generalization of the multi-modal learning problem, which describes learning scenarios where different sensors utilize different \textit{modalities}, such as images, audio or text.  In the most comprehensive survey on multi-view learning methods to date \cite{xu2013survey}, Xu et al. formalize the common assumptions of MVL by identifying two principles universal to all MVL algorithms: 

\begin{enumerate}
    \item \textbf{Consensus principle}: methods operating on different views should largely agree on what they find in each view
    \item \textbf{Complementary principle}: there may exist some information in each view, not found in other views, that can be exploited for learning tasks
\end{enumerate}

Historically, research in multi-view representation learning (MVRL) \cite{li2018survey} has largely focused on the first principle by seeking to exploit the information overlap across views. For instance, Multiview Subspace Learning, largely based on Canonical Correlation Analysis (CCA) \cite{hotelling1936relations}, seeks to maximize agreement between information extracted between views, measured in terms of correlation. Similar approaches to MVRL seek to also maximize agreement, but use different criteria such as distance or alternative similarity measures \cite{li2018survey}.  Together, these form \textit{alignment-based} approaches to MVRL \cite{li2018survey}.  

Recently, \cite{wang2016deep} introduced two probabilistic forms of CCA, one based on the probabilistic graphical model (PGM) of \cite{bach2005probabilistic} but that uses deep networks instead of linear models they call VCCA and, of more interest to us, one that segregates view-specific latent variables from view-common in a PGM they call VCCA-Private.  We believe that models such as VCCA-Private better capture the universal assumptions made in MVL because they segregate view-common latent variables which capture the information overlap between views (addressing the \textit{consensus} principle), from view-specific latent variables which capture information particular to each view (addressing the \textit{complementary} principle).  This allows full information preservation to be sought but in an orderly, disentangled manner which is a promising bias \cite{bengio2013representation} in representation learning at the moment \cite{eastwood2018framework,van2019disentangled,kim2018disentangling,chen2016infogan}.

In \cite{wang2016deep}, Wang et al. also bring together two veins of CCA-based MVRL research: deep learning approaches to CCA and probabilistic interpretations of CCA.  The difficulty with using the models they introduce (both with and without Private information), however, is that there is no closed-form maximum likelihood estimate for the network parameters.  Like \cite{kingma2013auto} do with Variational Autoencoders (VAE), Wang et al. overcome this by deriving a variational lower bound (also known as evidence lower bound or ELBO) that decomposes the loss function into components that seek to a) maximize the log probability of reconstructions and b) minimize the KL-divergence between the aggregated posterior coming from encoders with chosen priors.

However, there are limitations to maximizing the ELBO for a model of this structure: the KL-divergence between the prior(s) and aggregated posterior(s) needs to be a known, differentiable function and there are, arguably, better ways to match aggregated posteriors to priors, such as by using adversaries \cite{goodfellow2014generative,makhzani2015adversarial} or by using Maximum Mean Discrepancy (MMD) \cite{gretton2007kernel,zhao2017infovae}.

Adversarial Autoencoders (AAE) \cite{makhzani2015adversarial} took the existing VAE model and showed how discriminators could be used in place of the KL-divergence terms to allow multiple, straightforward extensions to the model: a) the use of a much larger class of priors that do not require known, differentiable KL divergence terms b) the integration of known label information in a partially supervised setup as well as c) unsupervised clustering d) semi-supervised extensions and e) dimensionality-reduction extensions.

In this work, we introduce Adversarial Canonical Correlation Analysis (ACCA) in two forms which match the model assumptions of VCCA and VCCA-Private, that we correspondingly call ACCA and ACCA-Private.  These models address a gap in the multi-view representation learning (MVRL) research landscape and offer similar extensions to the VCCA model that AAE offers to VAE, showing how adversaries can be used in place of differentiable KL divergence terms to match approximate posteriors to priors.  Although we believe that all of the extensions introduced in AAE are possible and straightforward to use with ACCA, in this work we focus largely on the use of arbitrary priors and the goodness of fit to those priors that adversaries can provide.  We leave other extensions to future work.

We also aim to highlight and motivate a new perspective of analysis largely missing from MVRL that is becoming prominent in other areas of Representation Learning research.  We believe this perspective on the proper aims of representation learning offers new theoretical insights inline with the universal principles of \cite{xu2013survey}: namely, that the purpose of representation learning is to \textit{disentangle} underlying factors of variation.  Conveniently, the VCCA-Private model of \cite{wang2016deep} and ACCA-Private we propose here allows this hypothesis to be explored.  We aim for this to be the first step in multi-view disentangling as a promising research direction, as it has become in general representation learning research.

\subsection{Our contributions}

\subsubsection{We design the ACCA and ACCA-Private algorithms and task-agnostic validation criteria.}
	In section three, we present the ACCA and ACCA-Private models and training algorithms, along with their task-agnostic validation criteria in section 3.4.
\subsubsection{We show that ACCA and ACCA-Private provide increased flexibility in choosing priors over VCCA and VCCA-Private, allowing new biases to be pursued in multiview representations.}
	This follows from work on adversarial density estimation \cite{goodfellow2014generative} and the flexibility that adversarial approaches offer at matching posteriors to priors.  For work in this area specific to autoencoders, see \cite{makhzani2015adversarial}.  We demonstrate this flexibility experimentally in section 5.3.
\subsubsection{We fill a minor research gap in VCCA by deriving variational lower bounds in terms of both views.}
	This is derived in section 2.4.2.
\subsubsection{We demonstrate that ACCA acts as a stronger regularizer on the posterior than VCCA, which comes at the expense of overall information content if the network is not powerful enough but allows better fit of posteriors to priors.}
	We show this experimentally in section 5.1 and 5.2.  For information that is clearly not Gaussian (categorical class information), the variational approach of VCCA fits the rough shape of the posterior to the prior, but allows large fissures, demonstrating a compromise between the discrete underlying aggregate posterior and the continuous prior.  The adversarial approach of ACCA, on the other hand, forces the gaps to close (because those gaps in the posterior get recognized easily by the discriminator), leading to a closer match between the posterior and prior, \textit{even though it comes at the expense of the overall information content}.  This demonstrates that the adversarial approach to matching the distributions can be thought of as a stronger regularizer than the variational approach of VCCA, which can be overcome by increasing the capacity of the network (which we show in the latter half of section 5.1 and 5.2).
\subsubsection{We perform new analysis on VCCA and VCCA-Private from a view-level disentangling perspective.}
	In section 2.4, we propose a new paradigm on disentangling in multiview settings. And, in section 5, we focus on highlighting where underlying factors of variation can be found.  For VCCA and ACCA, in section 5.1, we show how the factors of variation (class, style, angle of rotation for each of the views) is distributed across the dimensions of the latent representation.  In section 5.2, for VCCA-Private and ACCA-Private, we show how this information is distributed \textit{across} the three latent representations in addition to the individual dimensions in each, showing strong multiview disentangling behavior using the paradigm we propose in section 2.4.  Similarly, in section 5.3, we show how class information gets distributed across the posterior.  We take this perspective because we think that the disentangling properties of VCCA-Private and ACCA-Private - both across views and within each view are the most exciting future direction in multiview representation learning at the moment and that these frameworks offer a good theoretical foundation for future research.
\subsubsection{We construct a new dataset, Tangled MNIST, which is more appropriate for evaluating multi-view representation learning algorithms than existing benchmark datasets, such as Noisy MNIST.}
	Because we prioritize the distentangling perspective in this work, we move away from the Noisy MNIST dataset of \cite{wang2015deep,wang2016deep}.  The existing benchmark dataset does not contain low dimensional factors of variation particular to view $y$ - they use independent noise for each dimension of the view, which is incompressible and we have no hope of recovering it in the view-specific representation $h_y$.  We propose the new dataset, Tangled MNIST, in section 4.

%% file: sections/related_work.tex
\section{Background}
\label{related_work}

\subsection{Standard Canonical Correlation Analysis}
Canonical Correlation Analysis (CCA) \cite{hotelling1936relations} forms the basis for much research in MVRL \cite{li2018survey}, including our work.  In CCA, vectors $w_{x,0}$ and $w_{y,0}$ are sought for views $X = [x_1,...,x_N]$ and $Y = [y_1,...,y_N]$ that maximize the correlation between linear projections $A_0 = w_{x,0}^\intercal X$ and $B_0 = w_{y,0}^\intercal Y$, where $N$ is the size of the dataset.  Additional vectors $w_{x,i}$ and $w_{y,i}$ can be sought, subject to the restriction that they are uncorrelated with earlier vectors.  In matrix form, when $m$ projections are sought, all the projection vectors can be combined into matrices $W_x = [w_{x,0}, ..., w_{x,m}]$ and $W_y = [w_{y,0},...,w_{y,m}]$ and CCA rewritten as \cite{hardoon2004canonical}:

\begin{equation}\label{eq:1}
\begin{aligned}
& \underset{W_x, W_y}{\text{max}}
& & \mathrm{Tr}(W_x^\intercal C_{x,y} W_y) \\
& \text{subject to}
& &  W_x^\intercal C_{x,x} W_x = I,\\
& & &  W_y^\intercal C_{y,y} W_y = I,\\
& & &  w_{x,i}^\intercal C_{x,y} w_{y,j} = 0,\\
& & & i,j \in \{1,\ldots,m\}, i \neq j
\end{aligned}
\end{equation}

where $C_{x,x}$, $C_{y,y}$, and $C_{x,y}$ are the covariance matrices of $X$, $Y$ and between $X$ and $Y$, respectively.

\subsection{Nonlinear Canonical Correlation Analysis}

One of the main limitations of standard CCA is the reliance on simple linear projections.  When written in the form above, it is easy to see how kernel matrices could replace the covariance matrices, using the ``kernel trick'' \cite{hofmann2008kernel} and provide a nonlinear extension to the linear projection functions \cite{akaho2001kernel,lai2000kernel,hardoon2004canonical}.

Neural network extensions to CCA similarly allows complex, nonlinear hypothesis spaces for each view's projection function.  In \cite{lai1998canonical}, Lai and Fyfe demonstrate a way to use a neural network to maximize the correlation between individual projections for each view, but the network is simple and the function is still linear.  In \cite{lai1999neural}, they extend their work to multiple projections and introduce simple nonlinearities through the use of activation functions.

In \cite{hsieh2000nonlinear}, Hsieh made two breakthroughs relevant for this work.  They used multilayer perceptrons for each projection function (which relies on a negative correlation loss function) and added networks that try to reconstruct the inputs from the projections, essentially situating CCA in a multiview autoencoder framework, an important precedent for this work.  

The DCCA \cite{andrew2013deep} model of Andrew et al. was arguably the first to explore the use \textit{deep} networks.  They do not use reconstructions, but allow multiple projections and seek to maximize the total correlation between the outputs.  Unfortunately, the loss function requires an expensive correlation calculation across a batch or the entire dataset. They derive the gradient for the loss function but recommend using the entire dataset instead of batches after experimenting with using batches, limiting the utility of the approach.

In \cite{wang2015deep}, Wang et al. devise three variations of DCCA.  All three models use deep networks to generate projections and use the same correlation loss term, but they also include additional decoder networks that seek to reconstruct each view from the projections.  The first model, DCCAE, uses the same loss function and constraints (uncorrelation and normalization) as DCCA, but includes reconstruction terms in the loss function.  The second model, CorrAE, removes the uncorrelation constraint of DCCAE and uses the sum of scalar correlations between the projections.  The third model, DistAE, replaces the correlation term from DCCA with a distance-based criteria.  The authors motivate this using the work of \cite{hardoon2004canonical}, who show that CCA can be understood as minimizing the distances between the projections as long as they meet the uncorrelation and normalization (whitening) constraints.  It is also worth noting, as discussed in \cite{li2018survey}, that distance-based techniques are another type of \textit{alignment}-based approaches to MVRL, so they share some theoretical grounding.  

\begin{figure}
    \centering
    \begin{subfigure}[t]{.2\columnwidth} 
        \centering
        \includegraphics[width=\textwidth]{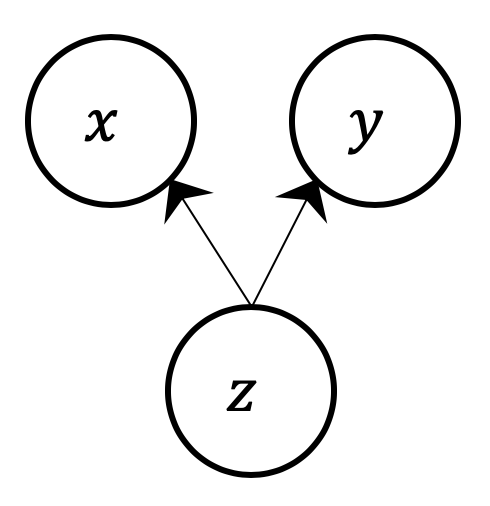}
        \caption{The Probabilistic CCA latent variable model of \cite{bach2005probabilistic}.}
        \label{fig:vcca_pgm}
    \end{subfigure}\hspace{3mm}
    \begin{subfigure}[t]{0.34\columnwidth} 
        \centering 
        \includegraphics[width=\textwidth]{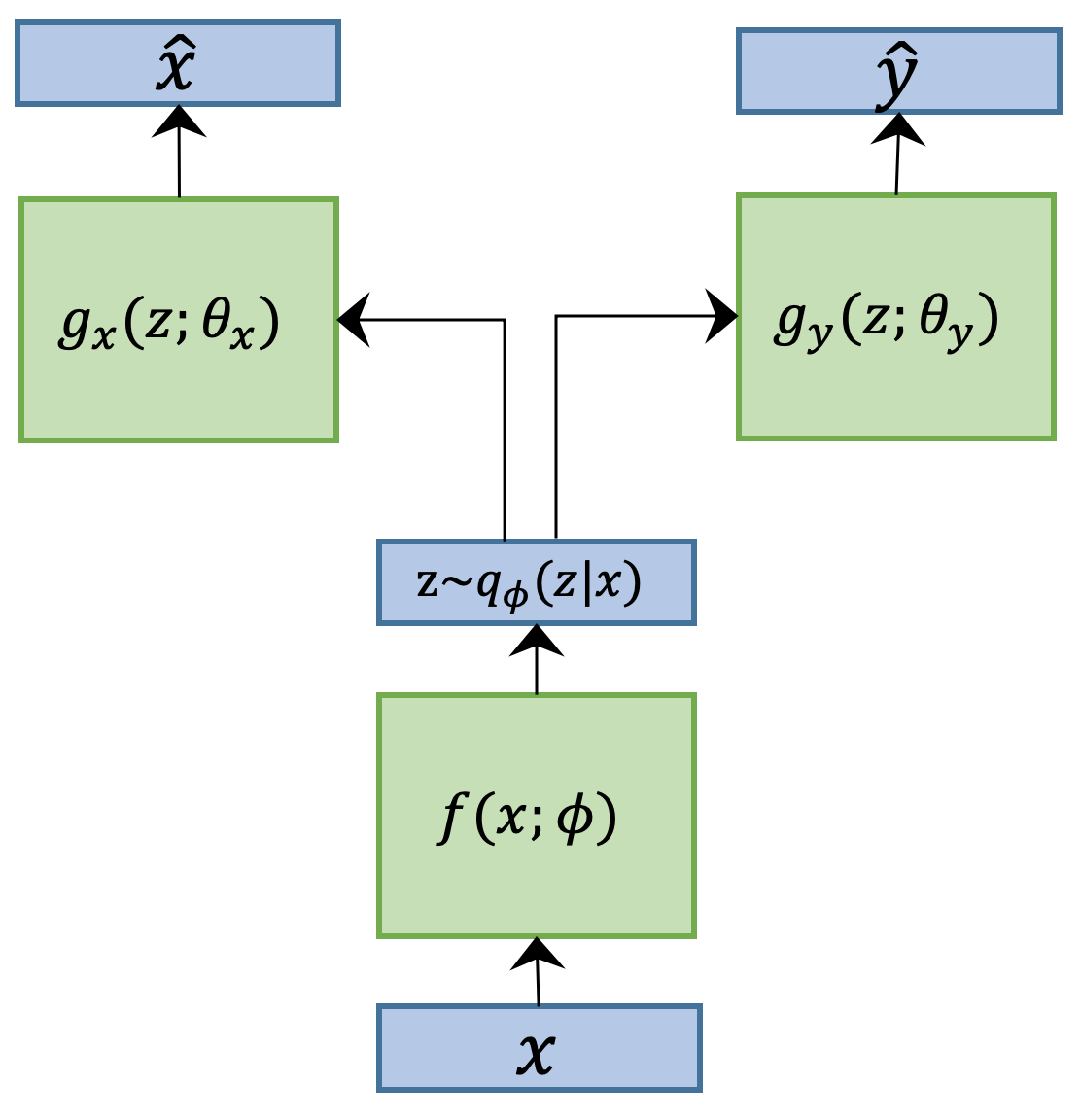}
        \caption{The network structure of VCCA \cite{wang2016deep}.}
        \label{fig:vcca_network_single}
    \end{subfigure}\hspace{3mm}
    \begin{subfigure}[t]{0.34\columnwidth} 
        \centering 
        \includegraphics[width=\textwidth]{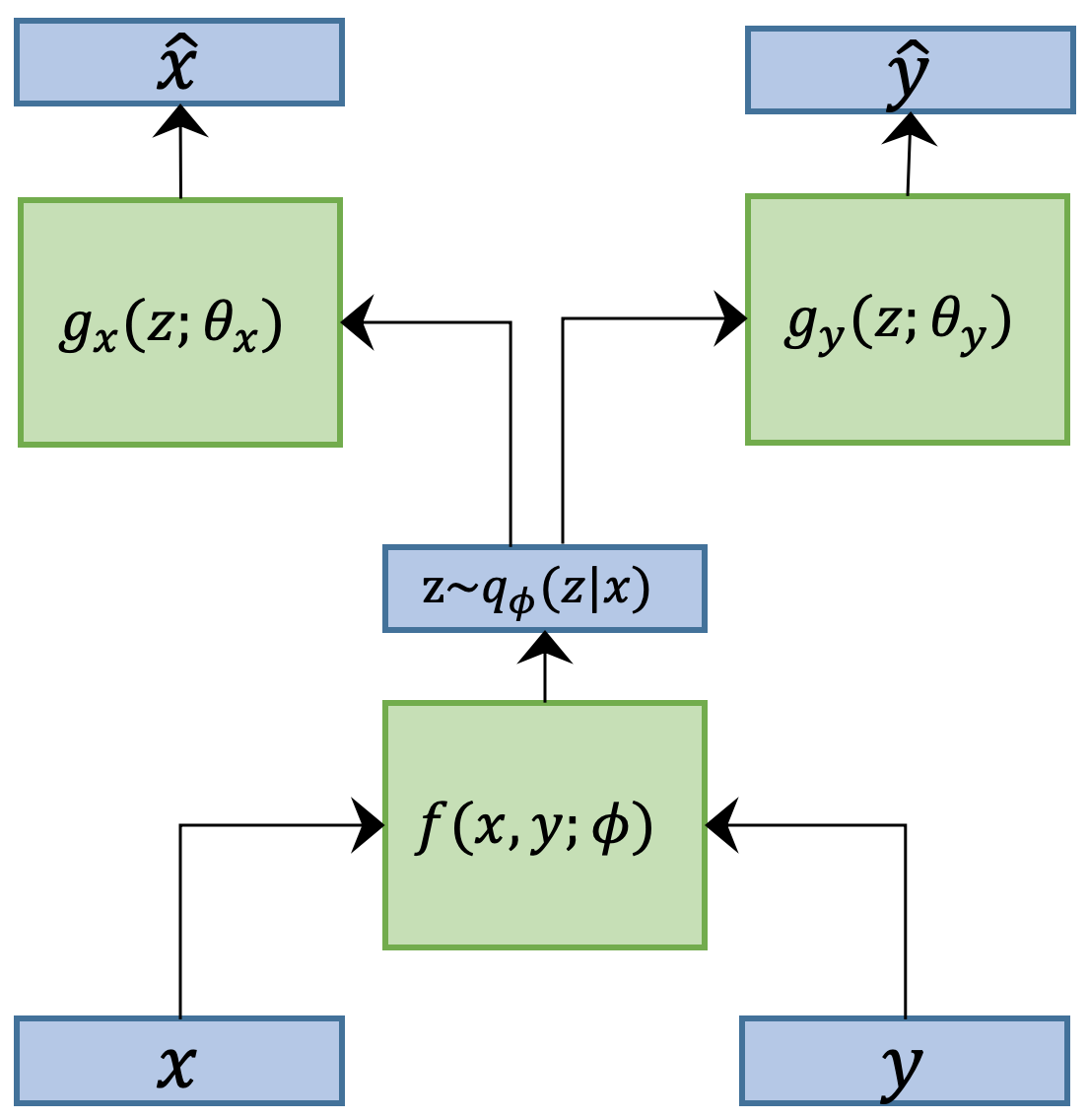}
        \caption{The network structure of VCCA we suggest, which utilizes both views as inputs.}
        \label{fig:vcca_network_double}
    \end{subfigure}
    \caption{The probabilistic CCA model (a) introduced by \cite{bach2005probabilistic} established CCA on probabilistic footing, but found maximum likelihoods for distributions assuming only simple linear projection functions.  The VCCA model (b) of \cite{wang2016deep} replaces these linear projection functions with deep neural networks, making closed form maximum likelihood estimates impossible.  To estimate model parameters, they instead derive a variational lower bound for the data log likelihood requiring the use of an encoder network $f(x;\phi)$ to approximate the single view posterior $q_{\phi}(z|x)$.  However, they only derive a lower bound where the encoder is a function of one view, $x$. We suggest both views should be used (c), which requires derivation of a new ELBO in terms of $q(z|x,y)$. We do so in section 2.4.2.}
\end{figure}

\subsection{Probabilistic Canonical Correlation Analysis}

One limitation of the modern deep CCA methods described above is their reliance on expensive correlation losses that must be computed over, at a minimum, batches.  There is another vein of CCA-based MVRL research, however, based ultimately on the probabilistic interpretation of CCA found in \cite{bach2005probabilistic} (PCCA) that has allowed deep networks to bypass this restriction.  With PCCA, Bach and Jordan offered a latent variable probabilistic interpretation of standard CCA using the PGM found in Fig. \ref{fig:vcca_pgm}, with $z$ as the latent variable for observed views $x$ and $y$.  This model has a factorization for $p(x,y,z)$ of $p(x,y,z)=p(z)p(x|z)p(y|z)$.  They make the following distribution assumptions:

\begin{gather*}
p(z) = \mathcal{N}(0,I_d)\\
p(x|z) = \mathcal{N}(W_xz + \mu_x, \phi_x)\\
p(y|z) = \mathcal{N}(W_yz + \mu_y, \phi_y)
\end{gather*}

Where ${\phi}_x$ and ${\phi}_y$ are positive semidefinite matrices, $min\{m_x,m_y\} \geq d \geq 1$, $x \in \mathbb{R}^m_x$, $y \in \mathbb{R}^m_y$, and $z \in \mathbb{R}^d$.  They show that the maximum likelihood estimates for this model lead to the standard CCA solution.  

\subsection{Variational Canonical Correlation Analysis}

While providing good probabilistic footing for CCA, PCCA suffers from the same linearity limitation for the projection functions as CCA.  Wang et al. overcome this in \cite{wang2016deep} with models VCCA and VCCA-Private.  Together, they form the primary basis for our work so we devote this section to understanding them.  

VCCA uses the same graphical model as PCCA shown in Fig. \ref{fig:vcca_pgm}, but radically changes the distribution assumptions by replacing the linear projection functions with deep neural networks, $g_x(z;\theta_x)$ and $g_y(z;\theta_y)$, where $\theta_x$ parameterizes the neural network for view $x$, $\theta_y$ parameterizes the neural network for view $y$, and when $\theta$ is used without a subscript, it refers to the set of network parameters for both models combined.  The new distribution assumptions VCCA makes are then:

\begin{gather*}
p(z) = \mathcal{N}(0,I) \\
p(x|z) = \mathcal{N}(g_x(z;\theta_x), I) \\
p(y|z) = \mathcal{N}(g_y(z;\theta_y), I)
\end{gather*}

The resulting model, while significantly more expressive, makes straightforward maximum likelihood estimation of the model parameters impossible.  To address this, they use the approach Kingma et al. \cite{kingma2013auto} take with variational autoencoders (VAE) and situate the model within an autoencoder framework (see Fig. \ref{fig:vcca_network_single}) and use the encoder network(s) to help maximize a variational or evidence lower bound (ELBO) on the data log likelihood coming from the generative model provided by the decoder.  

There is one other limitation, though, they address that is of particular interest to us.  CCA naturally exploits the consensus principle (discussed in our introduction) of \cite{xu2013survey} because both observed views $x$ and $y$ rely on a common latent variable, $z$.  While $z$ \textit{can} contain view-specific information for each view (after using the variational lower bound we derive above in terms of both views), it is not clear in what manner and certain VCCA architectures explicitly prevent this possibility (to be discussed shortly).  

Disentangled representation learning, on the other hand, has become a promising research direction in representation learning \cite{bengio2013representation,eastwood2018framework,van2019disentangled,kim2018disentangling,chen2016infogan} where information from underlying \textit{factors of variation} are isolated to individual latent dimensions. This research, however, focuses on single view settings and, to the best of our knowledge, there has been no proposed paradigm extended to multiview settings, so we here propose one. 

\begin{figure}
    \centering
    \begin{subfigure}[t]{.2\columnwidth} 
        \centering
        \includegraphics[width=\textwidth]{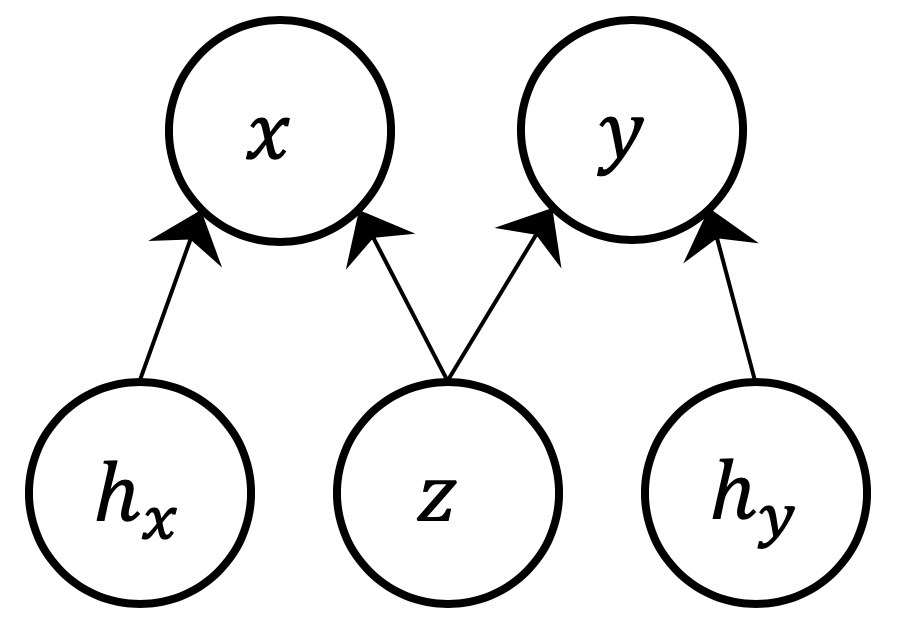}
        \caption{The graphical model of VCCA-Private.}
        \label{fig:vcca_private_pgm}
    \end{subfigure}\hspace{3mm}
    \begin{subfigure}[t]{0.32\columnwidth} 
        \centering 
        \includegraphics[width=\textwidth]{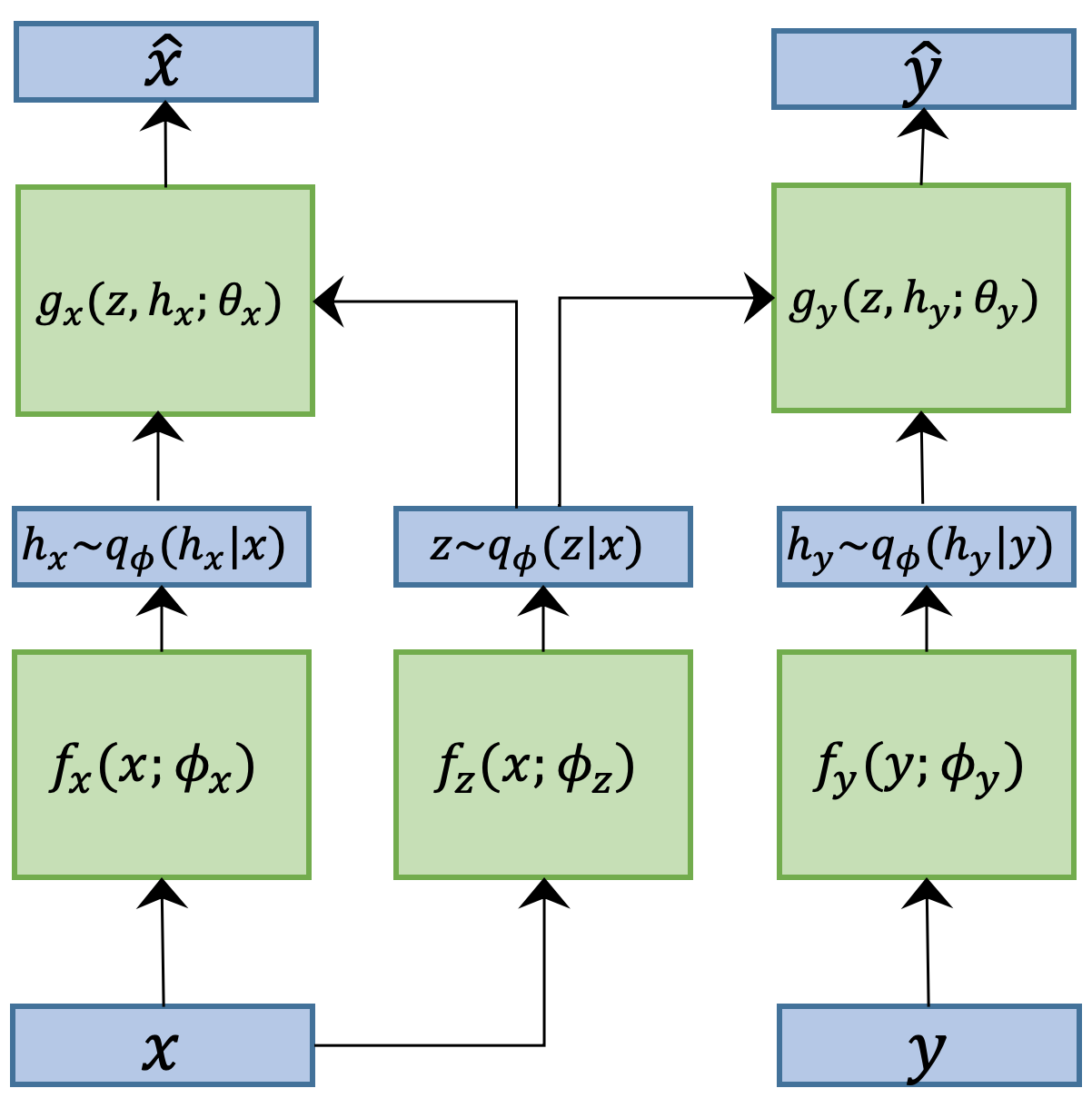}
        \caption{The network structure of VCCA-Private.}
        \label{fig:vcca_private_network_single}
    \end{subfigure}\hspace{3mm}
    \begin{subfigure}[t]{0.32\columnwidth} 
        \centering 
        \includegraphics[width=\textwidth]{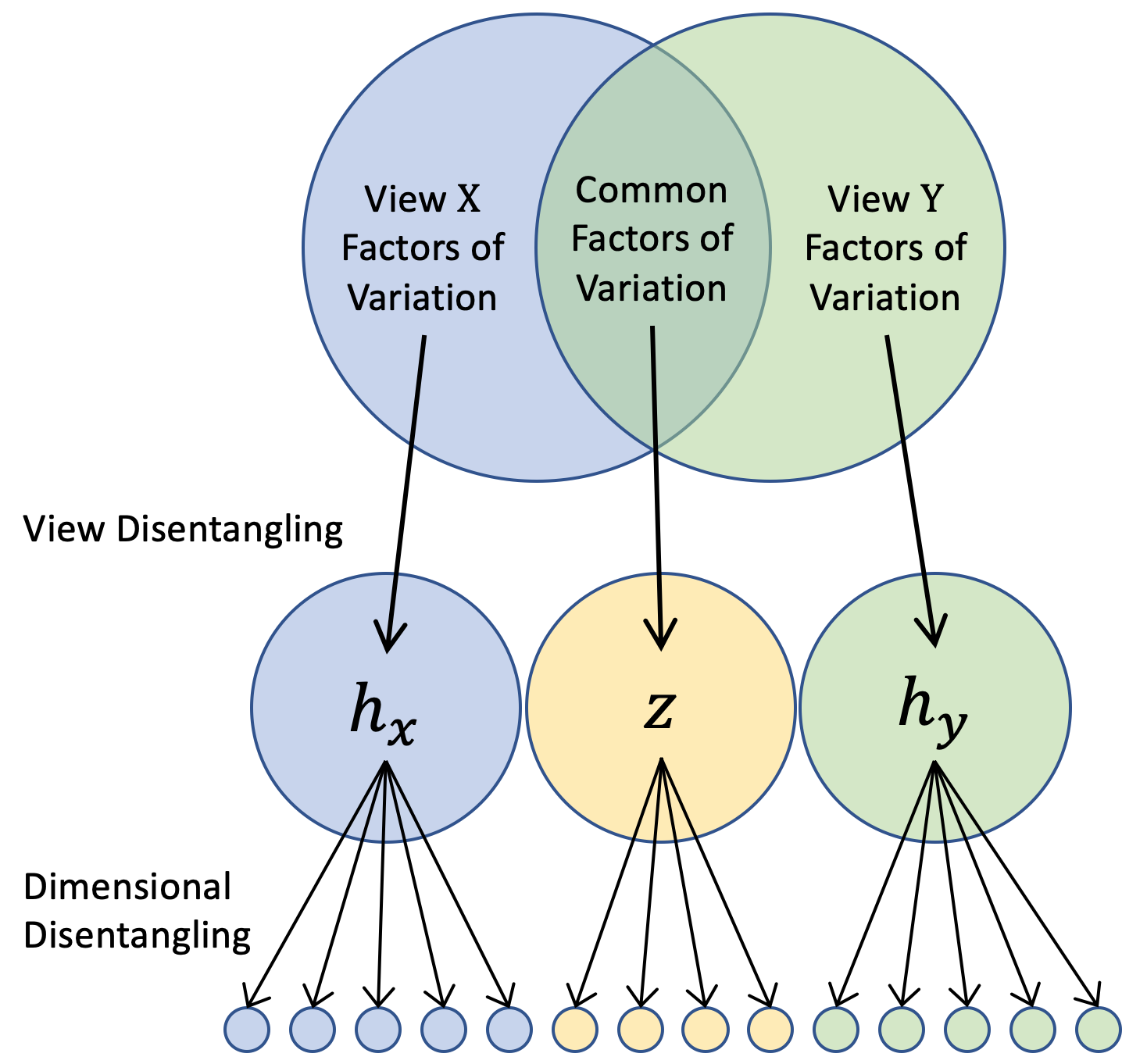}
        \caption{The two stages of the multiview disentangling paradigm we propose.}
        \label{fig:disentangling_perspective}
    \end{subfigure}
    \caption{The VCCA-Private model of \cite{wang2016deep}, with the graphical model of the joint distribution in (a) and the network structure (b) employed to maximize the data likelihood via maximizing the variational lower bound of the data likelihood. Existing disentangling research focuses on dimension-level disentangling in single view settings. We see disentangling in multiview settings as consisting of two levels. In (c), we show our perspective on multiview disentangling: information from factors of variation (FOV) in $X \cap Y$ is represented in multidimensional $z$. Remaining view-specific factors of variation $X \setminus Y$ and $Y \setminus X$ are represented in $h_x$ and $h_y$, respectively. Within each of these representations, the underlying factors of variation they represent may be highly tangled. So, the second level of disentangling seeks to isolate information from each underlying FOV into individual dimensions, ideally in a one-to-one mapping.}
\end{figure}

We propose a perspective on what disentangling means in multiview settings (see Figure \ref{fig:disentangling_perspective}. We argue that disentangling should be understood as occurring at two conceptual layers: view and then dimension. In the first layer, each view is understood as containing a set of factors of variation and we are interested in funneling information into each multivariate learned representation based on whether each of these underlying FOVs are shared or not. We want information from shared factors of variation to reside in $z$ and information from the remaining factors of variation to reside in their view-specific representation ($h_x$ for $x$ and $h_y$ for $y$). Denote the sets of FOVs for views $x$ and $y$ as $X$ and $Y$, respectively. 

In section 1, we describe the consensus and complementary principles of \cite{xu2013survey} and describe how the existence of \textit{information overlap} between $X$ and $Y$ is the governing principle of multiview data. In disentangling language, we argue that this means that $X \cap Y \neq \emptyset$. Furthermore, the complementary principle suggests that $(X \cup Y) \setminus (X \cap Y)$ might be nonempty because either $X \setminus Y \neq \emptyset$ or $Y \setminus X \neq \emptyset$. The consistency principle, therefore, justifies the existence of $z$ and the complementary principle justifies the existence of $h_x$ and $h_y$. This is why we believe that the distribution assumptions of VCCA-Private (which we use in ACCA-Private) shown in Fig. \ref{fig:vcca_private_pgm} are an excellent foundation for future research in view disentangling.

The second stage of multiview disentangling is inline with existing single view disentangling research: tangled factors of variation are isolated to individual dimensions within each of those representations.

VCCA, on the other hand, makes no effort toward view disentangling and seeks only to include information from $X \setminus Y$ in $z$. And, because the ELBO derived in \cite{wang2016deep} for VCCA uses an encoder defined only in terms of view $x$ ($q(z|x$), there is no hope of $Y \setminus Y$ information residing in $z$. We address this minor research gap in section 2.4.2 by deriving an ELBO in terms of $q(z|x,y)$. By choosing $p(z) = \mathcal{N}(0,I)$, VCCA can make strides toward the second stage of disentangling, though there is no reason in VCCA to think that view-specific factors of variation will be isolated from view-common.

\subsubsection{VCCA, Single View Encoder}

In \cite{wang2016deep}, Wang et al. derive an ELBO for VCCA and use an encoding distribution for $z$, $q_{\phi}(z|\cdot)$ defined in terms of just one view as an estimate of the posterior.  In other words, $q_{\phi}(z|x)$ is used as an estimate of $p(z|x)$ instead of using $q_{\phi}(z|x,y)$ as an estimate of $p(z|x,y)$ in the ELBO.  Practically speaking, this means that the encoder for $z$ can only, in theory, be expressed as a function of one view.  They argue that this allows certain uses, such as still allowing inference to be done by the network when only one view is available at test time.  They additionally explore using a convex combination of the losses from encoders over each view independently.

The ELBO they derive is as follows, which we denote $\mathcal{L}_{VCCA_x}$ (all following expectations are with respect to the data distribution unless otherwise specified):


\begin{equation*}
\begin{aligned}
& \log p_{\theta}(x,y) \\
& \geq \underbrace{-D_{KL}(q_{\phi}(z|x) \| p(z))}_\text{Distribution matching term} + \underbrace{\mathbb{E}_{q_{\phi}(z|x)} \big(\log p_{\theta}(x|z) + \log p_{\theta}(y|z) \big)}_\text{Reconstruction loss terms} \\
& = \mathcal{L}_{VCCA_x}(x,y;\theta, \phi) \\ 
\end{aligned}
\end{equation*}

Note that this derivation is general to any choice of distribution for $p(z)$ and $q_{\phi}(z|x)$, with the only requirement being that the KL divergence between the two distributions is closed form and differentiable.  

\subsubsection{VCCA, Two View Encoder}
As Wang et al. discuss \cite{wang2016deep}, in scenarios where only one view is available at test time, it is useful to have the encoder network be a function of only one view.  
However, this places an important restriction on the information $z$ can contain in VCCA.  While it can contain the common information between the views ($x \cap y$) and $x$-specific information ($x \setminus y$), it cannot contain $y$-specific information ($y \setminus x$) due to the structural limitations of the derivation.  While we believe the best approach is to disentangle each of these into separate latent variables, as is done in VCCA-Private and the ACCA-Private method we propose in this work, here we fill a minor research gap by deriving a variational lower bound for VCCA in terms of both views (see Fig. \ref{fig:vcca_network_double}), using $q_{\phi}(z|x,y)$:

\begin{equation*}
\begin{aligned}
& \log p_{\theta}(x,y) = \log p_{\theta}(x,y) \int q_{\phi}(z|x,y) dz\\
& = \int \log p_{\theta}(x,y) q_{\phi}(z|x,y) dz\\
& = \int q_{\phi}(z|x,y) \Big(\log \dfrac{q_{\phi}(z|x,y)}{p_{\theta}(z|x,y)} + \log \dfrac{p_{\theta}(x,y,z)}{q_{\phi}(z|x,y)} \Big) dz\\
& = D_{KL}(q_{\phi}(z|x,y) \| p_{\theta}(z|x,y)) + \mathbb{E}_{q_{\phi}(z|x,y)} \Big(\log \dfrac{p_{\theta}(x,y,z)}{q_{\phi}(z|x,y)} \Big)\\
& \geq \int q_{\phi}(z|x,y) \log  \Big( \dfrac{p_{\theta}(x,y,z)}{q_{\phi}(z|x,y)} \Big) dz\\
& = \int q_{\phi}(z|x,y) \log  \Big( \dfrac{p_{\theta}(x|z)p_{\theta}(y|z)p(z)}{q_{\phi}(z|x,y)} \Big) dz \\
& = \int q_{\phi}(z|x,y)  \Big( \log \dfrac{p(z)}{q_{\phi}(z|x,y)} +  \log p_{\theta}(x|z) + \log p_{\theta}(y|z) \Big) dz \\
& = \underbrace{-D_{KL}(q_{\phi}(z|x,y) \| p(z))}_\text{Distribution matching term} + \underbrace{\mathbb{E}_{q_{\phi}(z|x,y)} \big(\log p_{\theta}(x|z) + \log p_{\theta}(y|z) \big)}_\text{Reconstruction loss terms} \\
& = \mathcal{L}_{VCCA_{x,y}}(x,y;\theta, \phi) \\ 
\end{aligned}
\end{equation*}

\subsubsection{VCCA-Private}
In addition to VCCA, Wang et al. introduce a second model they call VCCA-Private which, we argue, makes much more realistic multi-view assumptions.  The PGM for this model can be found in Fig. \ref{fig:vcca_private_pgm}.  VCCA-Private introduces two new latent variables, one for each view: $h_x$ and $h_y$.  $h_x$ is particular to view $x$ and $h_y$ is particular to view $y$ while $z$ is still shared by both.  We think this model better gets at the heart of the assumptions made in multi-view data \cite{xu2013survey} described in the introduction: consensus information should reside in $z$ and complementary information should reside in $h_x$ and $h_y$.

\begin{gather*}
p(z) = \mathcal{N}(0,I) \\
p(x|z,h_x) = \mathcal{N}(g_x(z,h_x;\theta_x), I) \\
p(y|z,h_y) = \mathcal{N}(g_y(z,h_y;\theta_y), I)
\end{gather*}

Wang et al. similarly derive the variational lower bound for this model and arrive at

\begin{equation*}
\begin{aligned}
& \log p_{\theta}(x,y)\\
& \geq \underbrace{-D_{KL}(q_{\phi}(z|x) \| p(z)) -D_{KL}(q_{\phi}(h_x|x) \| p(h_x)) -D_{KL}(q_{\phi}(h_y|y) \| p(h_y))}_\text{Distribution matching terms} \\ 
& + \underbrace{\mathbb{E}_{q_{\phi}(z|x),q_{\phi}(h_x|x)} \log p_{\theta}(x|z,h_x) + \mathbb{E}_{q_{\phi}(z|x),q_{\phi}(h_y|y)} \log p_{\theta}(y|z,h_y)}_\text{Reconstruction loss terms} \\
& = \mathcal{L}_{Private}(x,y;\theta, \phi) \\ 
\end{aligned}
\end{equation*}

Readers familiar with variational autoencoders \cite{kingma2013auto} will likely recognize the form of these lower bounds as consisting of two parts.  The KL divergence terms match user-chosen priors to aggregated posteriors coming from encoder networks and the expected data log likelihood terms maximize the log probability of the observed data.  In practice the KL divergence terms require well behaved, differentiable expressions and the expected log probability of the reconstructions can be replaced with a reconstruction loss term (usually assumptions are made about the distributions to make these equivalent).

The \textit{goal}, then, of training is twofold: fit the aggregated posteriors to the user-chosen priors and minimize reconstruction error.  In \cite{makhzani2015adversarial}, Makhzani et al. show how the first goal can be better achieved using adversaries and introduce Adversarial Autoencoders (AAE).  Why?  First, they allow a much larger class of priors to be chosen from.  The restriction is no longer that a known, differentiable expression for the KL divergence between the aggregated posterior and prior is known.  Instead, the priors must simply be able to be sampled from.  Second, adversaries arguably do a better job of matching priors to posteriors than KL divergence does.  So, the primary research gap we aim to address in our work can be summarized as this: as AAE is to VAE, ACCA is to VCCA and ACCA-Private is to VCCA-Private.

Before moving on to a description of the model and how it is trained, we should mention the work of \cite{wang2019adversarial} since it is both inspired by VCCA and uses adversaries.  However, it is not a direct extension of VCCA to using adversaries.  It instead employs multiple adversarial autoencoders in addition to multiple cross-view autoencoders.  The aim of this work is to fill the gap described above.

%% file: sections/method.tex
\section{Adversarial Canonical Correlation Analysis}
\label{method}
In this section we present our ACCA model.

\subsection{ACCA Overview}
Before explaining the structure and mechanics of ACCA, we start with the simple adversarial autonencoder model of Fig. 3a.  In this model, inputs $x$ are encoded to $z$ using an encoder network $f(x,y;\phi)$ with parameters $\phi$.  We choose a distribution $p(z)$ from which we can sample and we wish to match $q_{\phi}(z|x)$ to $p(z)$ as closely as possible, as a function of the network parameters $\phi$.  A decoder network $g_x(z;\theta_x)$ parameterized by $\theta$ maps $z$ to $\hat{x}$ and we wish to minimize the difference between $x$ and $\hat{x}$.  

In VCCA, $z_i \sim \mathcal{N}(\mu(x_i),\sigma(x_i))$ with $\mu(x_i)$ and $\sigma(x_i))$ the actual outputs of the encoder.  In other words, the encoder outputs the parameters of the Gaussian that $z_i$ is sampled from, providing an additional source of variability to $z_i$ beyond the stochasticity of $x_i$.  In AAE \cite{makhzani2015adversarial}, Makhzani et al. explore this approach to the encoders, in addition to two others: deterministic and universal approximator posterior.  In the deterministic approach, $z_i$ is a deterministic function of $x_i$ and the only source of randomness comes from $x_i$.  In the universal approximator posterior approach, multiple samples from a noise distribution are added to the input, $x_i$, of the encoder and then averaged out.  They find that there is no noticeable difference between each version of $q_{\phi}(z|x)$ and use the deterministic approach.  We do the same for all encoders: $f(x,y;\phi)$ in ACCA and $f_z(x;\phi_z)$, $f_x(x;\phi_x)$ and $f_y(y;\phi_y)$ in ACCA-Private.

\subsection{Training ACCA}
The adversarial game being played between the encoder $f(x,y;\phi)$ and the discriminator networks $D(z;\psi)$ (parameterized by $\psi$) is used to match $q_{\phi}(z|x)$ to $p(z)$ and is a competition between the discriminator (which aims to differentiate between the two distributions) and the encoder (which aims to fool the discriminator, in addition to providing embeddings that allow good reconstructions though this requirement is outside the game and comes from the reconstruction loss).  The adversarial game between the encoder and discriminator can be written as:

\begin{equation}\label{eq:adversarial_game}
\begin{aligned}
\underset{\phi}{\text{min }} \underset{\psi}{\text{max }} \mathbb{E}_{z \sim p(z)} [\log D(z;\psi)] + \mathbb{E}_{x \sim p_{data}} [\log(1-D(f(x,y;\phi);\psi))]
\end{aligned}
\end{equation}

This game, which replaces the KL divergence terms of the VCCA loss, requires two separate training passes: a pass to update the discriminator parameters $\psi$ and a pass to update the encoder parameters $\phi$.  A third and final pass to update all the autoencoder parameters, $\phi$ and $\theta$, uses reconstruction loss only and corresponds to the data log likelihood terms of the VCCA ELBO.  We describe each of the three passes next.

\subsubsection{Identifying the Frauds: Discriminator Update}
In this pass, a batch $X_k$ is constructed containing $n$ samples from $X$ and then passed through the encoder to yield a batch of latent values which we denote $NEG_k$.  Similarly, we draw $n$ samples from $p(z)$ placed into a batch called $POS_k$.  The batch $NEG_k$ is considered the \textit{negative} batch to the discriminator $D(z;\psi)$ because it is trying to recognize elements of that batch as imposters and samples from $p(z)$ (the $POS_k$ batch) as the true or positive class.

$NEG_k$ and $POS_k$ are concatenated and fed to $D(z;\psi)$, which is a deep neural network with one output neuron followed by a sigmoid activation function.  Binary cross entropy loss is used with elements of $NEG_k$ using class label $y=0$ and elements of $POS_k$ using class label $y=1$.  The discriminator loss is then:

\begin{equation}\label{eq:discriminator_loss}
\begin{aligned}
\mathcal{L}_{disc} = \dfrac{-1}{2n} \sum_{i=1}^{i=2n} y_i\log D(z_i;\psi) + (1-y_i) \log (1 - D(z_i;\psi))
\end{aligned}
\end{equation}

\begin{flushleft}and the discriminator parameters $\psi$ \textit{alone} (not the encoder parameters $\phi$ even though the encoder was used in the forward pass) are updated as $\psi := \psi - \lambda \frac{d\mathcal{L}_{disc}}{d\psi}$.\end{flushleft}

\subsubsection{Fooling the Discriminator: Encoder Update}
In this pass, the encoder has the opposite objective as the discriminator did in the prior pass: its aim is to \textit{fool} the discriminator.  For this pass, a negative batch $NEG_k$ is constructed in the same manner.  However, the \textit{labels} for the batch are switched to $y=1$ and instead of updating the discriminator parameters, $\psi$, we update the parameters of the encoder, $\phi$, using loss

\begin{equation}\label{eq:generative_loss}
\begin{aligned}
\mathcal{L}_{gen} = \dfrac{1}{2n}\sum_{i=1}^{i=n}y_i\log D(z_i;\psi)
\end{aligned}
\end{equation}

and encoder parameters $\phi$ alone (not the discriminator parameters $\psi$ even though the network was used in the forward pass) are updated as follows $\phi := \phi - \lambda \frac{d\mathcal{L}_{gen}}{d\phi}$.

Notice the change in sign in the loss function.  The encoder's aim is to \textit{maximize} the likelihood of fooling the discriminator into mistaking negative batches as positive.

\subsubsection{Maximizing Data Log Likelihood (Minimizing Reconstruction Loss): Full Autoencoder Update}
The adversarial game described above served only to replace the KL divergence terms of the VCCA ELBO and does not in any way seek to maximize the data log likelihood terms of the ELBO.  So, to address those data log likelihood terms, a third and final pass is made through the full autoencoder.  Batches $X_k$ and $Y_k$ are sent through encoder $f(x,y;\phi)$ and then decoders $g_x(z;\theta_x)$ and $g_y(z;\theta_y)$ to yield $\hat{X_k}$ and $\hat{Y_k}$.  Reconstruction loss is then calculated as:

\begin{equation}\label{eq:recon_loss}
\begin{aligned}
\mathcal{L}_{recon} = \| X - \hat{X} \|_k + \| Y - \hat{Y} \|_k
\end{aligned}
\end{equation}

with $k\in \{1,2\}$.

\begin{flushleft}and both encoder $\phi$ and decoder $\theta$ parameters are updated as $\phi := \phi - \lambda \frac{d\mathcal{L}_{recon}}{d\phi}$ and $\theta := \theta - \lambda \frac{d\mathcal{L}_{recon}}{d\theta}$.\end{flushleft}

\begin{figure*}
    \centering
    \begin{subfigure}[t]{0.31\textwidth} 
        \centering
        \includegraphics[width=\textwidth]{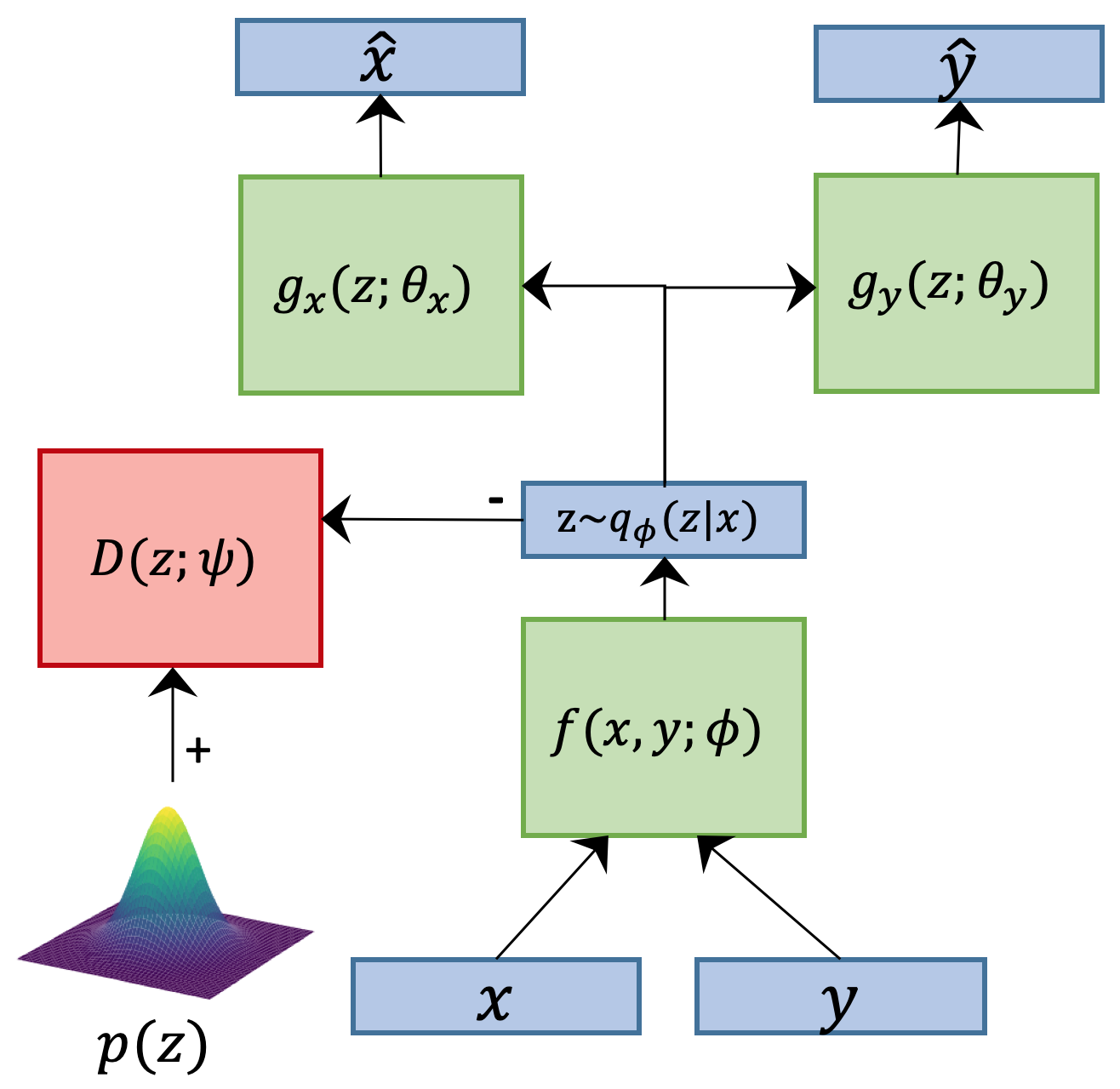}
        \caption{ACCA network structure}
        \label{fig:acca_network_structure}
    \end{subfigure}\hspace{5mm}
    \begin{subfigure}[t]{0.61\textwidth} 
        \centering 
        \includegraphics[width=\textwidth]{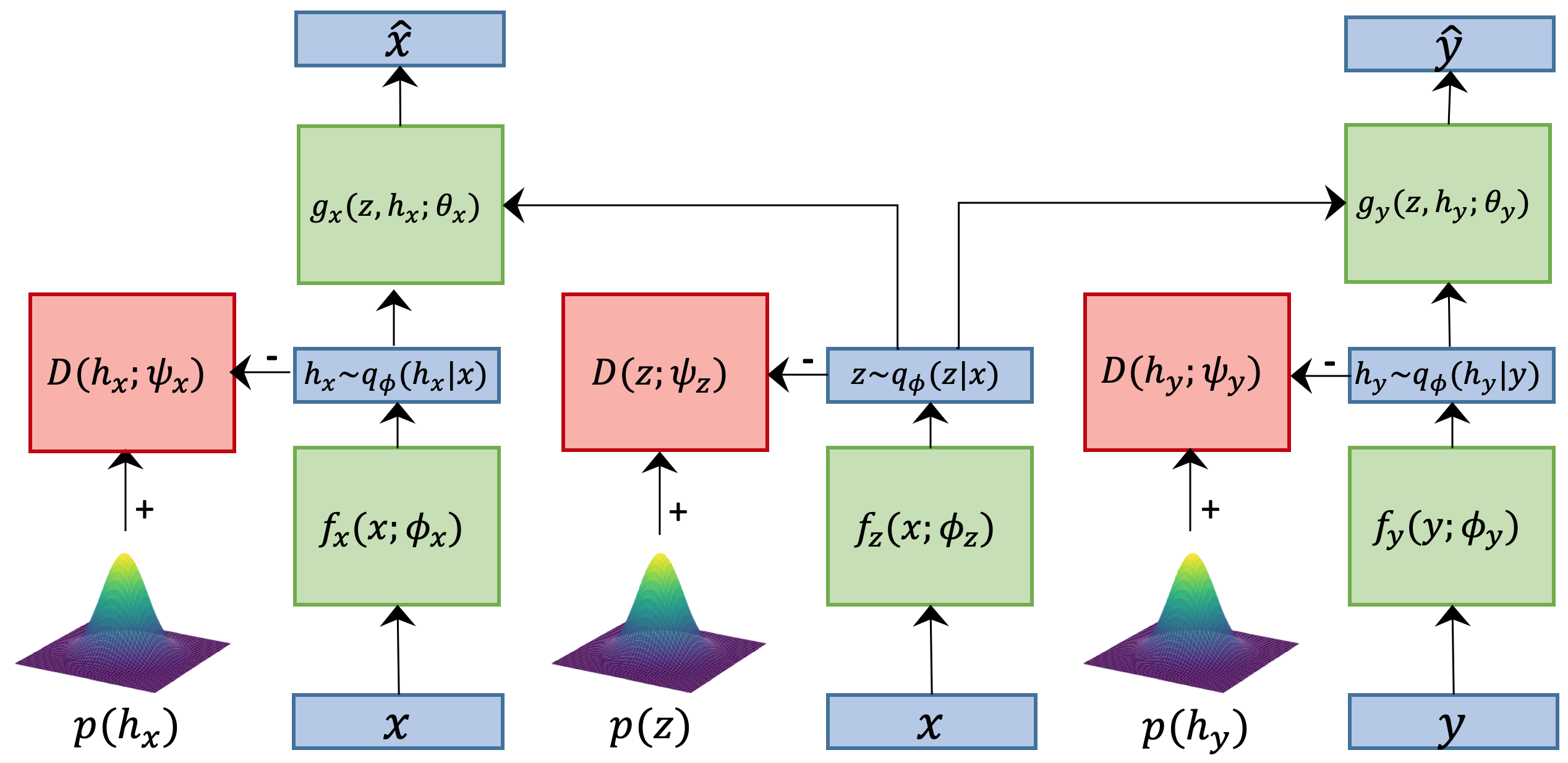}
        \caption{ACCA-Private network structure}
        \label{fig:acca_Private_network_structure}
    \end{subfigure}
    \caption{The network structures of (a) ACCA and (b) ACCA-Private.  Gaussian priors are shown everywhere just for convenience.  Also, notice that the encoder for $z$ in ACCA-Private is chosen to be a function of view $x$ alone.  We do this for comparability with VCCA-Private, since they only derive an ELBO for VCCA-Private using $z$ as a function of a single view and do not derive an ELBO for $z$ as a function of both views.  This is less problematic than it is for VCCA, though, because view specific information for $y$ is still accounted for in $h_y$. It is straightforward to make the encoder for ACCA a function of both views.}
\end{figure*}

\subsection{ACCA-Private Training}
Training of ACCA-Private proceeds in the same manner as ACCA except multiple embeddings are computed for each batch, corresponding to $z$, $h_x$, and $h_y$.  These happen in parallel during the first two passes since each of the encoders and discriminators are independent of each other.  The last stage, where reconstruction loss is calculated, is entirely feedforward but not independent: the decoder for $x$ relies on $z$ and $h_x$, and the decoder for $y$ relies on $z$ and $h_y$.  

\subsection{ACCA and ACCA-Private Validation Criteria}
One of the challenges with representation learning algorithms, in general, is how to design a principled validation criteria that is not overly task-biased.  In \cite{wang2016deep}, Wang et. al use classification accuracy from a linear SVM as the validation criteria for VCCA and VCCA-Private, arguing that it is a common use case for a representation learned on that dataset and the limited expressive power of the classifier does not itself aid the representation.  We seek a less task-specific bias.  Because VCCA and VCCA-Private have a single loss value and representation learning is an unsupervised task, we use the best loss value during training as the stopping criteria.  

The adversarial game of equation \ref{eq:adversarial_game} is minimized when $p(z)$ is matched to $q(z|x)$, that is, when the discriminator loss (equation \ref{eq:discriminator_loss}) and generator loss (equation \ref{eq:generative_loss}) are both equal at random chance, which reduces to $-\log(0.5) \approx .693147$.  And, all things being equal, we want to minimize reconstruction error.  Because of this, we use the following as the validation criteria on ACCA in order to both minimize deviations from the solution to equation \ref{eq:adversarial_game} and reconstruction error. We find it to be a good validation criteria in practice that balances goodness of fit to priors as well as general information content in the representations:

\begin{equation}\label{eq:acca_validation}
\begin{aligned}
 \texttt{ValACCA} = 
| -\log(0.5)-\mathcal{L}_{disc}| + |-\log(0.5)-\mathcal{L}_{gen}| + \mathcal{L}_{recon}
\end{aligned}
\end{equation}

\begin{flushleft}For ACCA-Private, this becomes: \end{flushleft}

\begin{equation}\label{eq:acca_private_validation}
\begin{aligned}
 \texttt{ValACCAP} = \mathcal{L}_{recon} + \dfrac{ \sum_{var \in \{z,h_x,h_y\}} | -\log(0.5)-\mathcal{L}_{disc_{var}}| + |-\log(0.5)-\mathcal{L}_{gen_{var}}|}{3}
\end{aligned}
\end{equation}

%% file: sections/dataset.tex
\section{The Tangled MNIST Dataset}

\begin{figure}
    \centering
    \begin{subfigure}[t]{.47\columnwidth} 
        \centering
        \includegraphics[width=\textwidth]{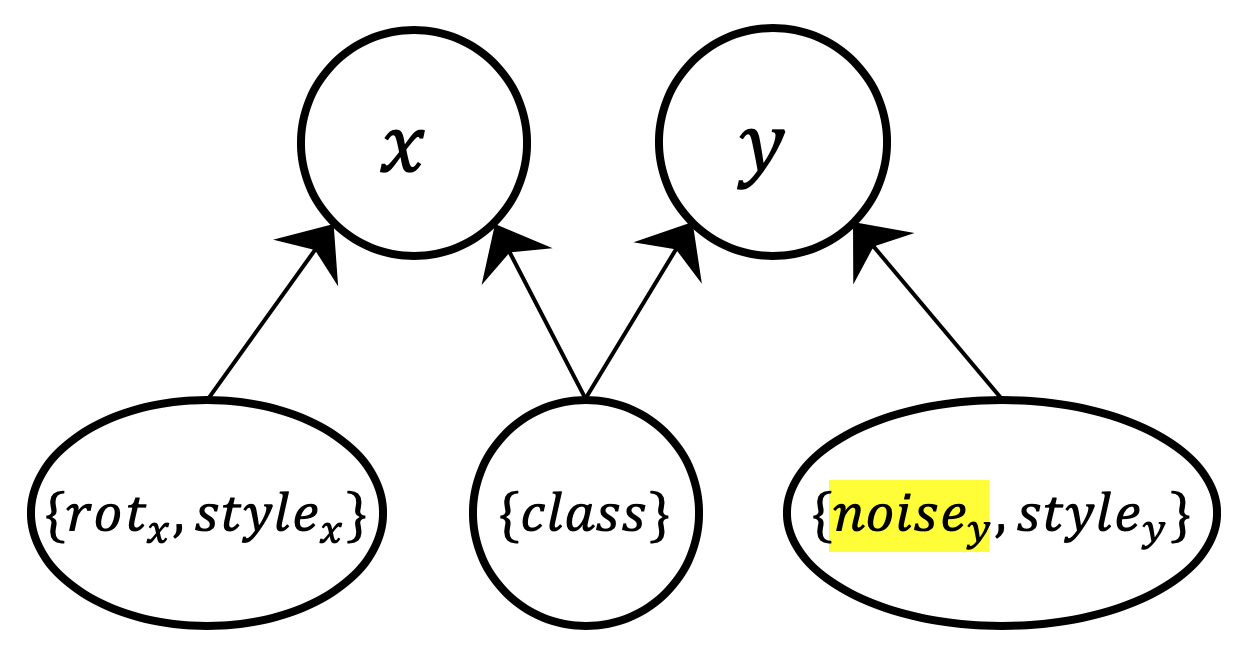}
        \caption{The generative model for Noisy MNIST.  $\texttt{noise}_y$ is incompressible (784 dimensions, the same number as the images) additive Gaussian noise. Difference from Tangled MNIST is highlighted.}
        \label{fig:noisy_mnist}
    \end{subfigure}\hspace{2mm}
    \begin{subfigure}[t]{0.47\columnwidth} 
        \centering 
        \includegraphics[width=\textwidth]{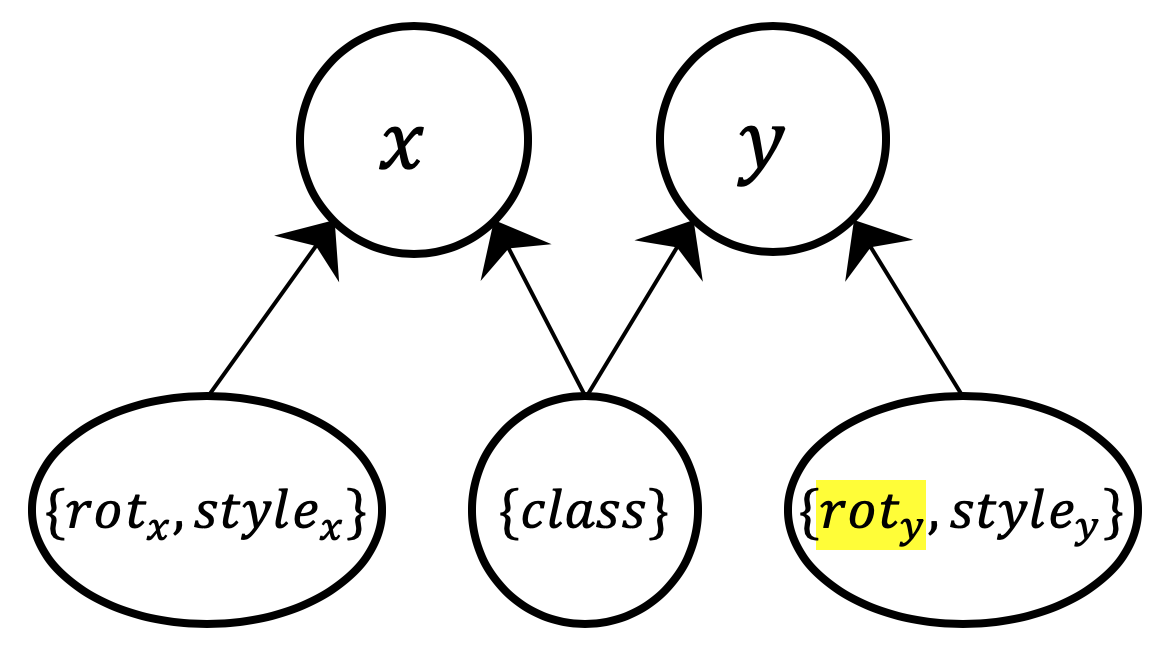}
        \caption{The generative model for Tangled MNIST.  $\texttt{noise}_y$ is replaced with $rot_y$, a one-dimensional factor of variation. Difference from Noisy MNIST is highlighted.}
        \label{fig:tangled_mnist}
    \end{subfigure}\hspace{2mm}
    \begin{subfigure}[t]{\columnwidth} 
        \centering 
        \includegraphics[width=\textwidth]{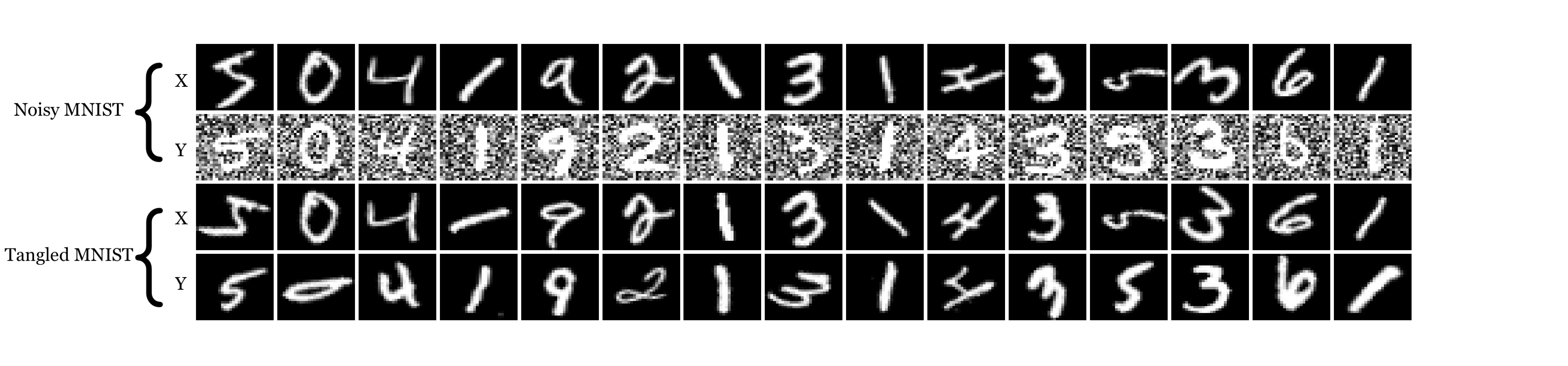}
        \caption{Samples from Noisy MNIST and Tangled MNIST.  Rows 1 and 2 are samples from view $x$ and $y$ of Noisy MNIST, respectively.  Rows 3 and 4 are samples from view $x$ and $y$ of Tangled MNIST, respectively. In both datasets, each view is a random sample from the same class and view $x$ is rotated a random angle of rotation sampled uniformly from $(\dfrac{-\pi}{4},\dfrac{\pi}{4})$. They differ in view $y$. In Noisy MNIST, incompressible independent noise is added to each pixel, which makes quantitative analysis from a disentangling perspective challenging. In Tangled MNIST, we replace the incompressible noise with an independent angle of rotation sampled uniformly from the same set as view $x$, $(\dfrac{-\pi}{4},\dfrac{\pi}{4})$. This yields a low dimensional underlying factor of variation for view $y$ which we have hope to capture in learned representations.}
        \label{fig:dataset_samples}
    \end{subfigure}
    
    \caption{The Noisy MNIST and Tangled MNIST datasets.}
\end{figure}

Wang et al. \cite{wang2015deep} introduced a dataset they call Noisy MNIST in order to study the behavior of DCCAE, CorrAE, and DistAE.  It is also used in \cite{wang2016deep} to study VCCA and VCCA-Private.  The dataset is constructed as follows: first, scale MNIST to $[0,1]$.  Then, for each element of MNIST, rotate the image a random amount by sampling the angle of rotation uniformly from $(\dfrac{-\pi}{4},\dfrac{\pi}{4})$.  Consider this view $x$.  To generate view $y$, first randomly choose another MNIST digit of the same class.  Then, for each pixel of the second image, add independent noise sampled uniformly from $[0,1]$ and then truncate the resulting values back to $[0,1]$.  

There is one major limitation to using this dataset for multi-view representation learning analysis: because the noise is independent per dimension in view $y$, it is incompressible and, in models such as VCCA-private and ACCA-private, we have no hope of recovering it in $h_y$ unless it has enough dimensions to capture the number of independent noise dimensions in $y$ (784 in MNIST).  Because we believe that the PGM introduced by VCCA-Private, which contains view-specific information, sets an important precedent for MVRL research, we think it is important to use a dataset that has some known information for each of $z$, $h_x$, and $h_y$ to contain of lower dimensions to make analysis more tractable.  For Noisy MNIST, class information is common between views and should reside in $z$ and the angle of rotation for view $x$ is specific to that view, so should reside in $h_x$.  Because of this, we introduce a new dataset that is a minor variation to Noisy MNIST that will allow us to explore some of the disentangling properties of VCCA, VCCA-Private, ACCA, and ACCA-Private that we call Tangled MNIST.  It is constructed in the same manner as Noisy MNIST except that no noise is added to view $y$.  Instead, view $y$ is rotated by an independent angle of rotation sampled from $(\dfrac{-\pi}{4},\dfrac{\pi}{4})$.  Because it is sampled independently from the angle of rotation for view $x$, this information should reside in $h_y$, giving known information to investigate in each of $z$ (class information), $h_x$ (angle of rotation for view $x$), and $h_y$ (angle of rotation for view $y$).  


In Tangled MNIST, there are 5 independent factors of variation, 3 of which are known and 2 of which are theoretical (the style dimensions which are of unknown dimensionality):

\begin{enumerate}
    \item class (common)
    \item $x$ angle of rotation, $\texttt{rot}_x$
    \item $y$ angle of rotation, $\texttt{rot}_x$
    \item $x$ style (intra-class coordinates for $x$)
    \item $y$ style (intra-class coordinates for $y$)
\end{enumerate}

The generative model for Noisy MNIST is shown in Fig. \ref{fig:noisy_mnist}.  The underlying factors of variation are organized by whether they are view-common or view-specific.  The generative model for Tangled MNIST is likewise shown in Fig. \ref{fig:tangled_mnist}.  Observe that the only difference between the two figures is that $\texttt{noise}_y$ from Noisy MNIST is replaced with $\texttt{rot}_y$ in Tangled MNIST. Example samples from each dataset are shown in Fig. \ref{fig:dataset_samples}.

Because of these advantages, all the following experiments were run on Tangled MNIST.

%% file: sections/experiments.tex
\section{Experiments}
\label{experiments}

In the spirit of reproducibility, all code for the experiments and figures generated in this section are available for download\footnote[1]{\url{https://github.com/bcdutton/AdversarialCanonicalCorrelationAnalysis}}.  


\input{sections/experiment_1.tex}

\input{sections/experiment_2.tex}

\input{sections/experiment_3.tex}

%% file: sections/experiment_1.tex
\subsection{Experiment 1: ACCA and VCCA on Tangled MNIST}

We compare ACCA to VCCA in this section using the Tangled MNIST dataset from the previous section. All experimentation in this section uses the network structure of Fig. \ref{fig:vcca_network_double} where the encoder is a function of both views and is based on the ELBO derived in section 2.4.2.

To start, we first compare the general performance and training behavior of the models when using 5 dimensions for $z$. One of the main objectives here is to show that\textit{ACCA's efforts to match $q(z|x,y)$ to $p(z)$ act as a stronger regularizer than the KL-Divergence term of VCCA}. This means better adherence of the aggregate posterior to the prior, but at the expense of reduced information content unless network capacity is increased.

Tangled MNIST has 5 independent factors of variation (3 of which are known), so we choose $z$-dim=5.  We use the same network architectures as \cite{wang2016deep} except that we do not use dropout: four fully connected hidden layers in both the encoder and decoder with 1024 units each followed by ReLU activation functions except on the layer leading to $z$ (where no activation function is used) and the final decoder layer (which uses sigmoid activation functions since the training data is all in $[0,1]$). All networks for all experiments were trained 100 epochs and the best model was selected using the validation criteria from section 3.4.

Fig. \ref{fig:51_Reconstructions} shows the general reconstruction and generation performance at these settings. Fig. \ref{fig:51_Losses} shows the difference between the loss functions for VCCA and ACCA. Having just a single loss function, VCCA training is more stable in practice and ACCA must balance converge in the adversarial game of equation \ref{eq:adversarial_game} between the encoder and discriminator with the overall information content of $z$ optimized during the reconstruction training phase. As can be seen from \ref{fig:ACCA_5_Losses}, the adversarial game has converged by epoch 20. 

In Fig. \ref{fig:51_Information}, we show the corresponding information curves during training for both methods. After each epoch, we train a linear SVM from scikit-learn \cite{scikit-learn} on $z$ to predict each of the known underlying factors of variation (class, $\texttt{rot}_x$, and $\texttt{rot}_y$) and use the accuracy values for class and coefficient of determination for $\texttt{rot}_x$ and $\texttt{rot}_y$ as measures of the information content in the representation for each of these factors of variation. In this figure, you can observe that both ACCA and VCCA have similar overall information content across each of the FOVs, \textit{however}, the overall scores for VCCA are higher. We observed this at these settings across multiple runs and hypothesized that this is due to the stronger regularizing behavior of ACCA's discriminator as compared to the KL-Divergence term in VCCA's loss.

To explore this hypothesis, we seek to observe the behavior of each method in lower dimensions so that we can observe the posteriors of each method in relation to its information capacity. To do this, we drop $z$-dim down to 2. Because there are 5 independent FOVs and only 2 representation dimensions, we expect a competition over information content in the representations. In order to ease this a bit, we increase the network capacity by adding two additional fully connected ReLU layers to the decoder so that even complicated representations in $z$ space have a better chance at interpretable reconstructions. In Fig. \ref{fig:51_Losses_2}, we can observe the losses under these new settings for each method. ACCA's adversarial game has converged by epoch 15 or so. In Fig. \ref{fig:51_Information_2}, we can see some of the differences in terms of information content in each of the representations. We should note that, under repeated runs of this experiment, there is variability in these graphs due to the constrained nature of the situation. However, class information consistently dominates. We believe this is because it is most helpful in reconstruction.  

The behavior of the posteriors becomes clear when we look at the embeddings of Tangled MNIST. In Fig. \ref{fig:51_Embeddings_2}, we show these embeddings for both ACCA and VCCA. In part (a), we show the embeddings of the training data from Tangled MNIST three times for VCCA and color each of the embeddings based on a different underlying FOV. On the left, we color by class. In the middle, we color by $\texttt{rot}_x$, and on the right, we color by $\texttt{rot}_y$.  In part (b), we do the same for ACCA. In part (c), we focus on the differences in goodness of fit of the posteriors. On the left are contours from the log probability of the prior $p(z) \sim \mathcal{N}(0,I)$. For our posteriors, we estimate the log probabilities using Kernel Density Estimation from scikit-learn \cite{scikit-learn} with a Gaussian kernel of bandwidth 0.2. The plot of the log probabilities of $q(z|x,y)$ for VCCA is in the middle and the plot for ACCA is on the right. The behavior shown here was consistently observed across runs of this experiment. The KL Divergence loss term for VCCA consistently allows ``fissures'' or holes in $q(z|x,y)$ as a compromise with reconstruction error. The discriminator from ACCA, however, identifies and punishes holes in $q(z|x,y)$, leading to a better fit between $p(z)$ and $q(z|x,y)$, even though it is often at the expense of overall information content. In other words, this is strong evidence that the discriminator from ACCA is a stronger regularizer on learned representations than the KL Divergence loss term from VCCA. 

In Fig. \ref{fig:51_Generations_2}, we provide additional insight into how information is represented in $z$. Starting from $z$ space, we iterate over a grid from $(-4,4) \times (-4,4)$ with step size 0.25. At each step, at the center of the grid, we send $z$ through the decoder for each of the views. In the top row are the two resulting generated views for $x$ and $y$, respectively, for VCCA. In the bottom row are the resulting generated views for ACCA. The differences in the density of the space between VCCA and ACCA can more closely be observed here, as well as the way in which rotation and style information is grouped into class clusters.

%% file: sections/experiment_2.tex
\subsection{Experiment 2: ACCA-Private and VCCA-Private on Tangled MNIST}

In this section, we explore the differences between ACCA-Private (see Fig. \ref{fig:acca_Private_network_structure} for the network structure) and VCCA-Private using Tangled MNIST. Because three separate representations are learned instead of one, visualization of aggregate posterior behavior is more natural and we start off using $z$-dim=$h_x$-dim=$h_y$-dim=2.  Each model is trained for 100 epochs using the same network structure as Wang et al. \cite{wang2016deep} (see Fig. \ref{fig:vcca_private_network_single}). In each encoder, four fully connected hidden layers of 1,024 units are used with ReLU activations except at the final layer before the representation output where no activation is used. For both decoders, the same layers are used except the final layers which produce the reconstructions use sigmoid activations instead of no activations. No dropout was used anywhere and, during training, each model was selected using the validation criteria described in section 3.4.  

Reconstructions from this experiment can be found in Fig. \ref{fig:52_Reconstructions}. In Fig.\ref{fig:52_Losses}, we show the loss curves during training for each method. Note the additional training complexity for ACCA-Private since there are now \textit{three} adversarial games being played between three separate encoders and discriminators. The corresponding information curves are shown in Fig. \ref{fig:52_Information}. Each curve corresponds to a pairing of representation (from $\{z,h_x,h_y\}$) and known factor of variation ($\{\texttt{class},\texttt{rot}_x, \texttt{rot}_y\}$). Notice how the blips in the adversarial losses from Fig. \ref{fig:ACCA_Private2_Loss} correspond to information reshuffling in \ref{fig:ACCA_Private2_Information}.

This is an important figure, though, for much larger reasons. In section 2.4, we propose a perspective on what disentangling means in multiview settings. In particular, we argue that disentangling should be understood as occurring at two conceptual layers: view and then dimension. In the first layer, each view is understood as containing a set of factors of variation. In a slight abuse of notation, lets call these sets $X$ and $Y$. In section 1, we describe the consensus and complementary principles of \cite{xu2013survey} and show how \textit{information overlap} between $X$ and $Y$ is a governing principle of multiview data. In disentangling language, we argue that this means that $X \cap Y \neq \emptyset$. Furthermore, the complementary principle suggests that $(X \cup Y) \setminus (X \cap Y)$ might be nonempty because either $X \setminus Y \neq \emptyset$ or $Y \setminus X \neq \emptyset$. By measuring information content in this way, using linear models on the representation to predict the underlying factor of variation, we are following precedents in single view disentangling metrics, such as the DCI metric of \cite{eastwood2018framework} and the SAP metric of \cite{kumar2017variational}.

With this paradigm in mind, we can say that in both ACCA-Private and VCCA-Private, we observe strong view-level disentangling. In both models, $z$ classification accuracy is the highest predicted value, followed by $h_x$ $\texttt{rot}_x$ accuracy, followed by $h_y$ $\texttt{rot}_y$ accuracy. This where we \textit{want} information to go based. Because class information is shared by both views, we want it in $z$ since it represents common information. Since $\texttt{rot}_x$ is a factor of variation particular to view $x$, we want it in $h_x$. And, since $\texttt{rot}_y$ is a factor of variation particular to view $y$, we want it in $h_y$. Having said that, there is significant room for improvement that can be observed in this figure (and the next two): class information can observed in $h_x$ and $h_y$. In ACCA-Private, the values for both are around 30\% and in VCCA-Private, both values are around 23\%.

In Fig. \ref{fig:52_VCCA_Embeddings} and \ref{fig:52_ACCA_Embeddings}, we can observe the embeddings of $z$, $h_x$, and $h_y$ over the training data of Tangled MNIST. In part (a), we show a 3$\times$3 plot. The first row shows the embeddings for $z$, the second row shows the embeddings for $h_x$, and the third row shows the embeddings for $h_y$. Each column is colored by one of the factors of variation. Column one is colored by class, column two is colored by $\texttt{rot}_x$, and column three is colored by $\texttt{rot}_y$. It is clear from each of these figures that view disentangling is happening naturally: along the diagonal of this plot, there is clear color differentiation. On the off diagonals, colors are highly intermixed. This should not be surprising considering the values just observed in  \ref{fig:ACCA_Private2_Information}. It is also easy to observe how some class information has spilled into $h_x$ and $h_y$. In particular, observe how the class represented by blue has is distinguishable from the other colors.

In part (b) of Fig. \ref{fig:52_VCCA_Embeddings} and Fig. \ref{fig:52_ACCA_Embeddings}, we show the estimated log probabilities of the prior and each of the aggregate posteriors: $q(z|x)$, $q(h_x|x)$, and $q(h_y|y)$, respectively. Once again, as in the previous subsection, we observe that aggregate posteriors for VCCA and VCCA-Private can ``fissure'' when underlying factors of variation are discrete. This offers further evidence that the regularization provided by the discriminator in ACCA and ACCA-Private is stronger and does a better job at matching the aggregate posterior to the prior than VCCA and VCCA-Private.

Lastly, we show how increasing the representational capacity can easy information reshuffling in VCCA-Private and ACCA-Private. In Fig. \ref{fig:52_Information_4}, we increased $z$-dim, $h_x$-dim, and $h_y$-dim to size 4 and plotted the information curves. All other parameters remain the same. You can see how higher representation capacity, though theoretically less interesting and grounded, does provide higher prediction values and more stable training behavior in ACCA-Private.

%% file: sections/experiment_3.tex
\subsection{Arbitrary Priors}

One of the main advantages that ACCA and ACCA-Private offer over VCCA and VCCA-Private is the ability to match their aggregate posteriors to relatively arbitrary priors.  We emphasize this fact in this section because we believe it is important groundwork for future research in multiview representation learning based on the strategic choice of priors, such as those that offer disentangled representations. To demonstrate the flexibility offered by ACCA and ACCA-Private in this regard, we construct a complicated prior by wrapping a uniform distribution over an S-manifold in 3-d.  Samples from the prior can be seen in Fig. \ref{fig:ACCA_S_Prior} (note: coloring is chosen here simply to highlight the shape of the manifold).

We train ACCA using $z$-dim=3 on Tangled MNIST with the S-manifold distribution acting as $p(z)$.  The best model is chosen using the validation criteria discussed earlier for ACCA in section 3.4.  The loss and information curves are shown in Fig. \ref{fig:ACCA_S_Loss} and Fig. \ref{fig:ACCA_S_Info}. Once again, because $z$-dim is less than the number of factors of variation, we observe some volatility in information content during training with occasional reshuffling.  

Complicated priors can also be used with ACCA-Private and chosen independently for each representation, as shown in Fig. \ref{fig:ACCA_Private_S_Embeddings}, where $z$-dim=3, $h_x$-dim=$h_y$-dim=2, $p(h_x)=p(h_y)=\mathcal{N}(0,I)$, and $p(z)$ is the S-manifold distribution above.  The embeddings (colored by their corresponding FOV information) are shown.

%% file: sections/conclusion.tex
\section{Conclusion}
\label{conclusion}

In this work, we show how adversaries can be used in multi-view represention learning using the probability models of VCCA and VCCA-Private.  We show that these approaches 

\begin{enumerate}
    \item Act as more powerful regularizers, matching aggregate posteriors to priors more effectively
    \item Allow a broader class of more complicated distributions to be used without complicated KL Divergence derivations
\end{enumerate}

We also propose a new perspective on disentangling in multiview settings and explore the multi-level disentangling properties of VCCA-Private and ACCA-Private, where disentangling factors of variation can be understood as occuring at two levels: at the view level and then at the dimension level. At the view level, information on underlying factors of variation is siphoned into multidimensional representations based on whether they are common to all views or not. Subsequent, dimensional disentangling follows existing single view disentangling research where information on factors of variation is sought in individual dimensions of learned representations. For ACCA-Private and VCCA-Private, we observe that, while view disentangling is occurring, there is some bleed over with common information spilling over into the view-specific representations and needs further research and refinement. We also design a validation criteria for ACCA and ACCA-Private that works well in practice, bypassing the need for manual investigation during training.  Lastly, we derive a variational lower bounds for VCCA using $q(z|x,y)$, allowing $z$ to store both view's view-specific information and not just $x$. 

\begin{acknowledgements}
This work was funded by the NNSA Office of Defense Nuclear Nonproliferation R\&D through the Consortium for Nonproliferation Enabling Capabilities (CNEC). The author would like to acknowledge his advisor, Raju Vatsavai, for his guidance and support.
\end{acknowledgements}

%% file: sections/appendix.tex
\clearpage
\section{Appendix}
\subsection{Additional Plots from Section 5.1}

\begin{figure}[ht]
    \centering
    \begin{subfigure}[t]{0.245\textwidth}
        \centering
        \includegraphics[width=\textwidth]{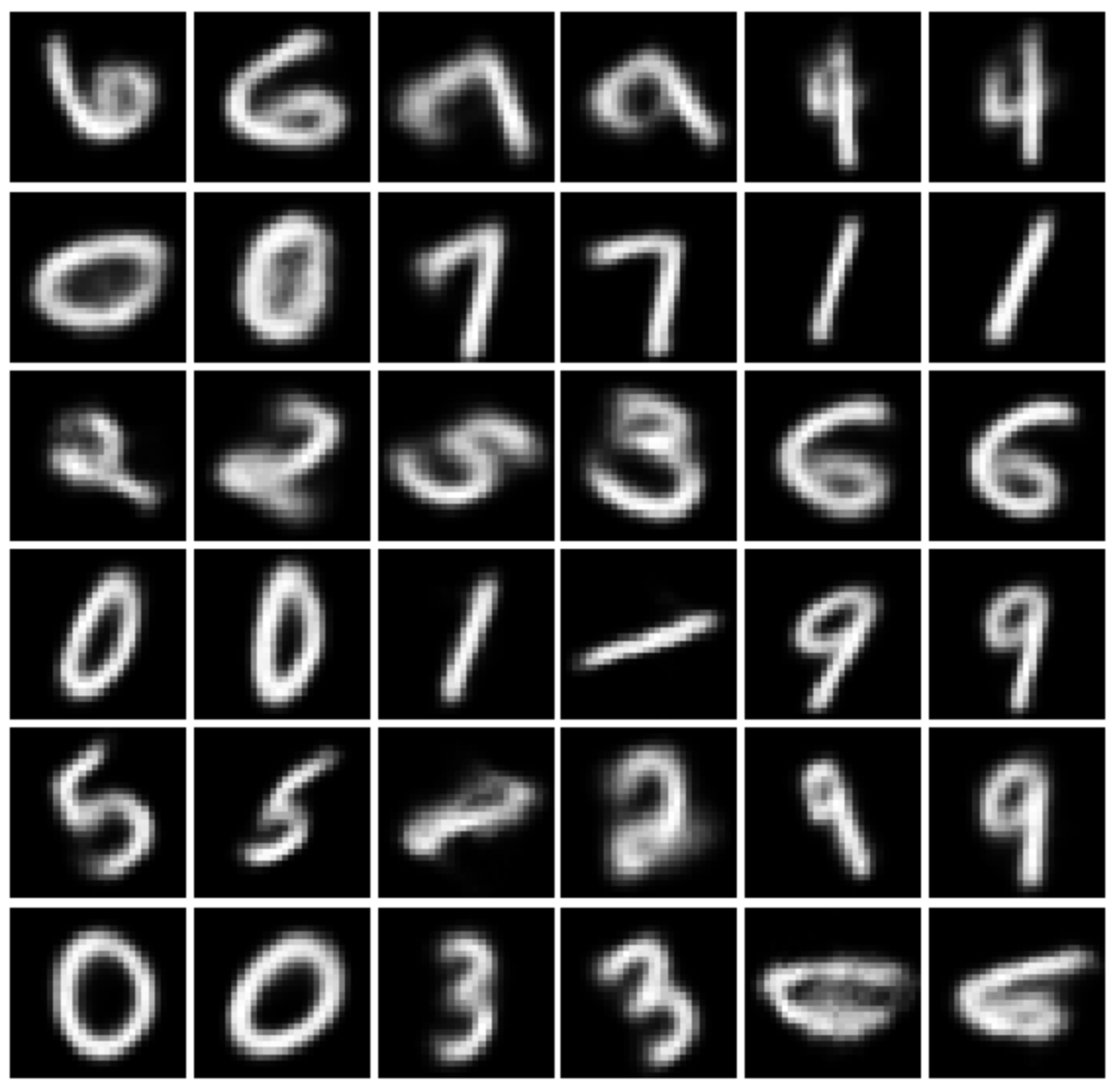}
        \caption{36 random generations from VCCA}
        \label{fig:VCCA_5_Generations}
    \end{subfigure}
    \quad
    \begin{subfigure}[t]{0.25\textwidth}  
        \centering 
        \includegraphics[width=\textwidth]{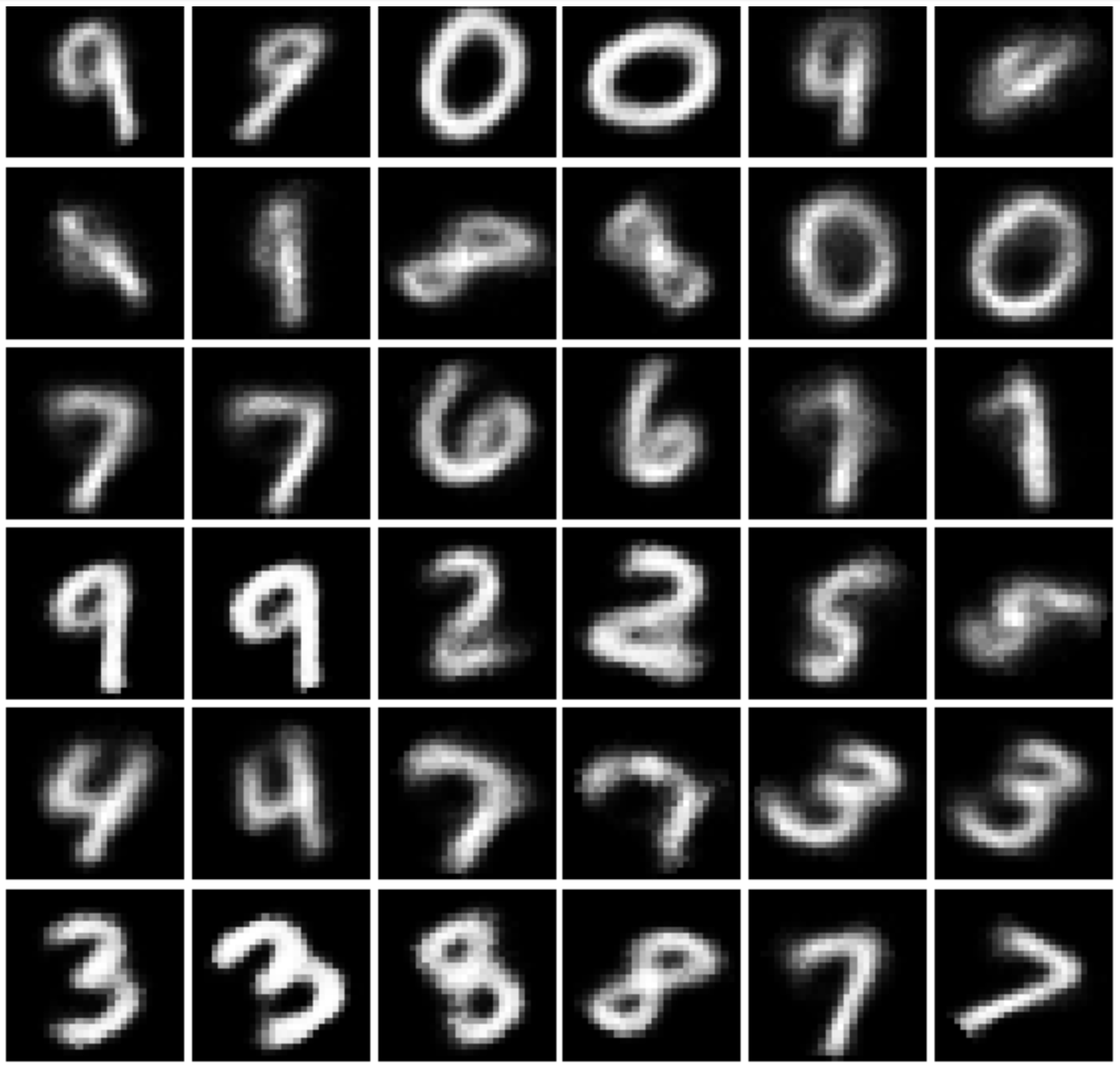}
        \caption{36 random generations from ACCA}
        \label{fig:ACCA_5_Generations}
    \end{subfigure}
    \quad
    \begin{subfigure}[t]{0.195\textwidth}   
        \centering 
        \includegraphics[width=\textwidth]{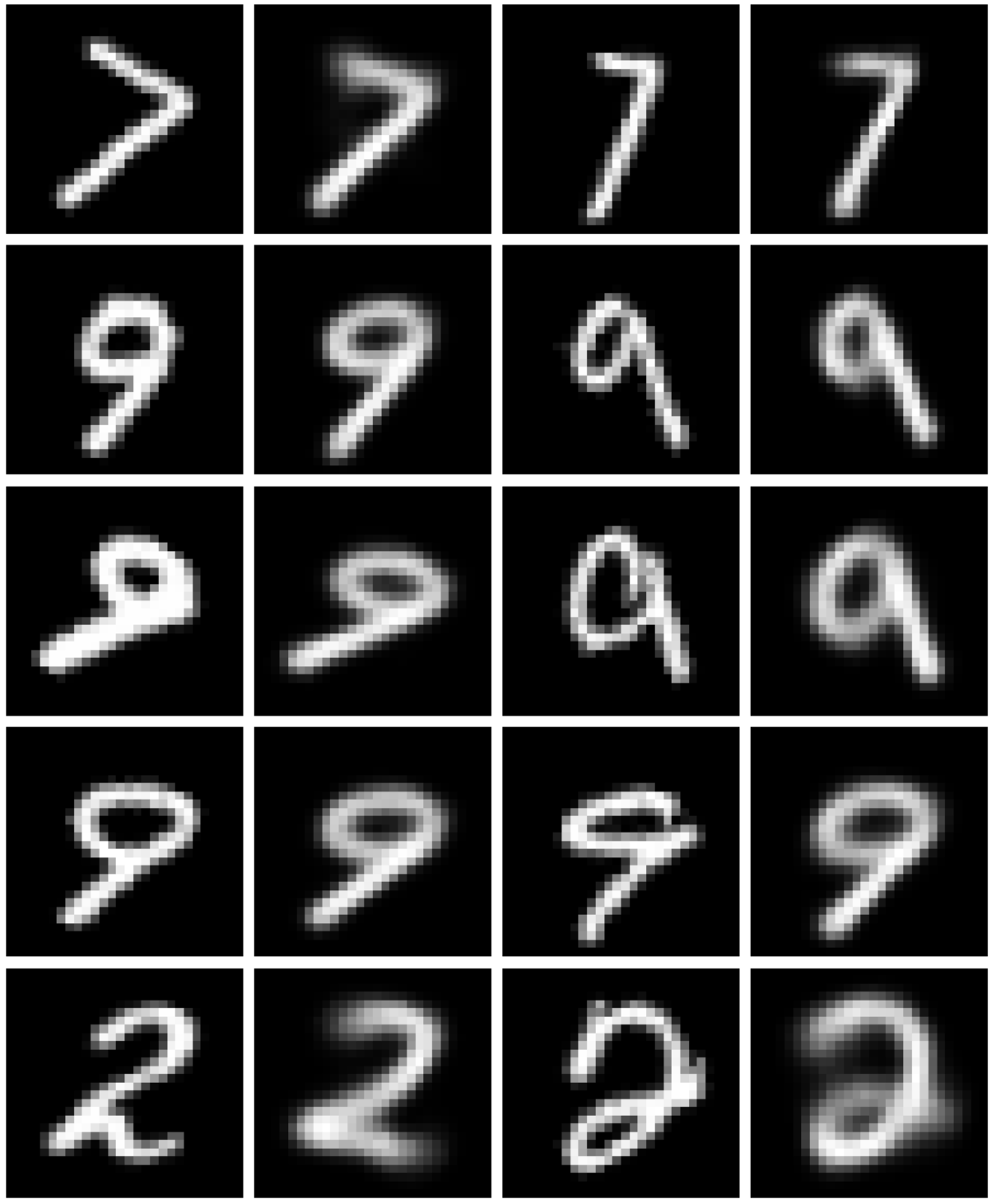}
        \caption{5 reconstructions for both views $x$ and $y$ from VCCA. Each row contains $x$, $\hat{x}$, $y$, and $\hat{y}$, respectively.}
        \label{fig:VCCA_5_Reconstructions}
    \end{subfigure}
    \quad
    \begin{subfigure}[t]{0.195\textwidth}   
        \centering 
        \includegraphics[width=\textwidth]{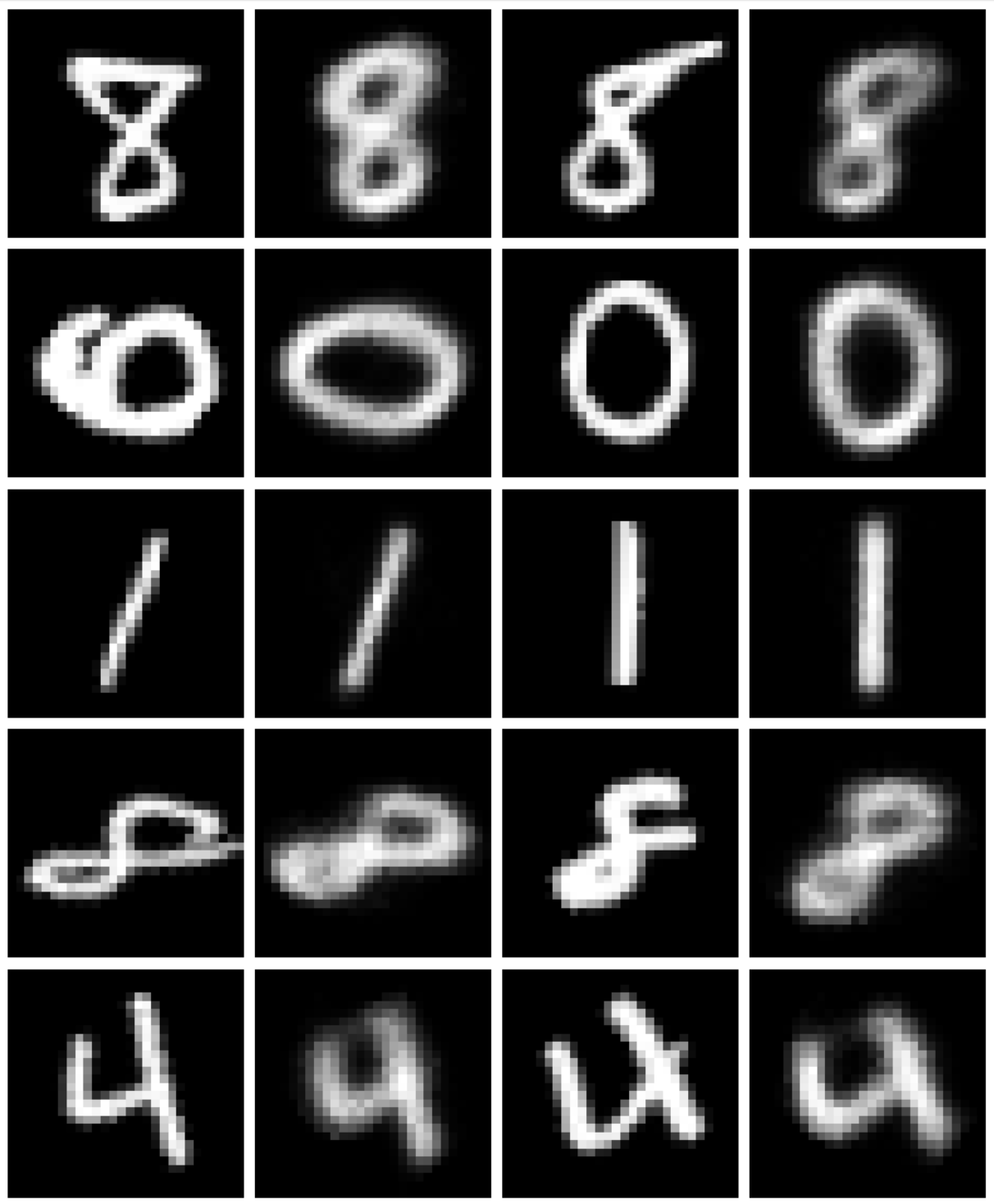}
        \caption{5 reconstructions for both views $x$ and $y$ from ACCA. Each row contains $x$, $\hat{x}$, $y$, and $\hat{y}$, respectively.}
        \label{fig:ACCA_5_Reconstructions}
    \end{subfigure}
    \caption{\small VCCA and ACCA generations and reconstructions on Tangled MNIST with $z$-dim=5}
    \label{fig:51_Reconstructions}
\end{figure}

\begin{figure}[ht]
    \centering
    \begin{subfigure}[t]{0.975\textwidth}
        \centering
        \includegraphics[width=\textwidth]{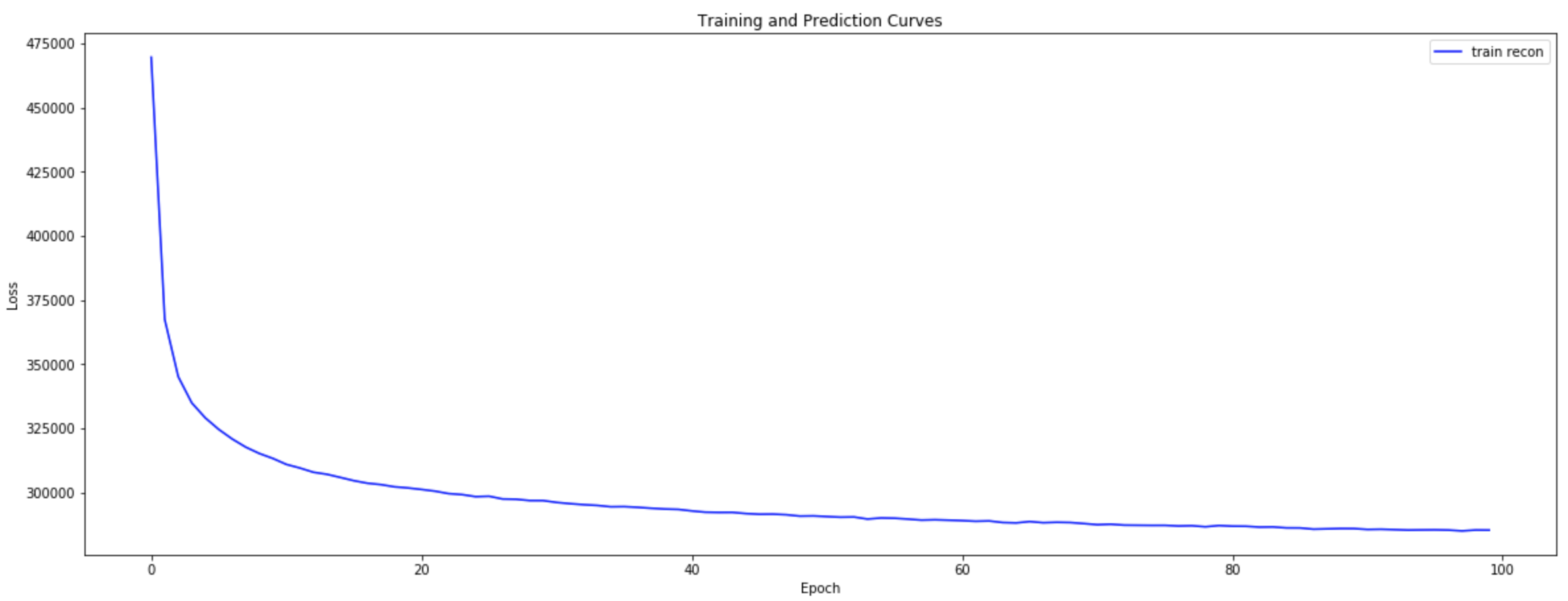}
        \caption{Loss during training}
        \label{fig:VCCA_5_Losses}
    \end{subfigure}
    \begin{subfigure}[t]{0.975\textwidth}  
        \centering 
        \includegraphics[width=\textwidth]{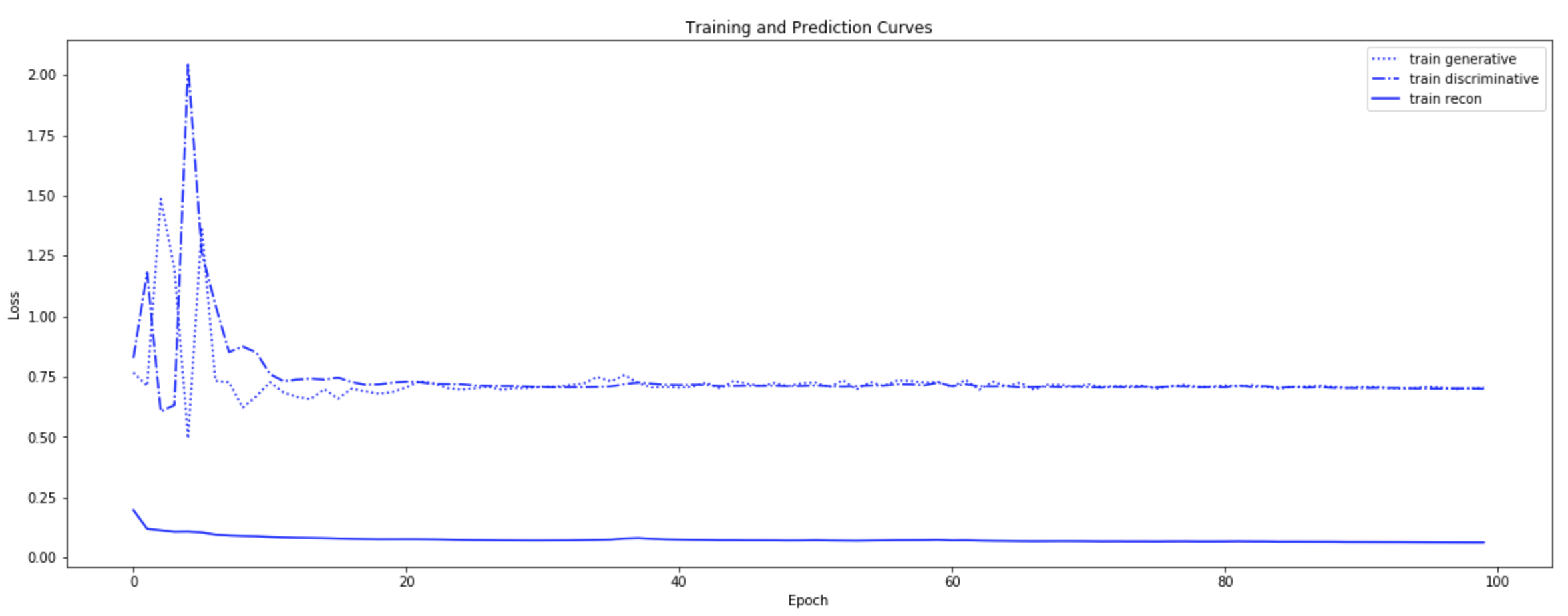}
        \caption{Losses during training}
        \label{fig:ACCA_5_Losses}
    \end{subfigure}
    \caption{VCCA and ACCA loss curves during training on Tangled MNIST for 100 epochs with $z$-dim=5. VCCA, having just a single loss curve and training pass exhibits more training stability than ACCA, which requires three losses and three separate training passes. In ACCA, the adversarial game between the discriminator and the encoder converges at $-\log(0.5) \approx .693147$. Here you see ACCA's adversarial game converge around roughly epoch 20.}
    \label{fig:51_Losses}
\end{figure}

\begin{figure}[ht]
    \begin{subfigure}[t]{0.975\textwidth}
        \centering
        \includegraphics[width=\textwidth]{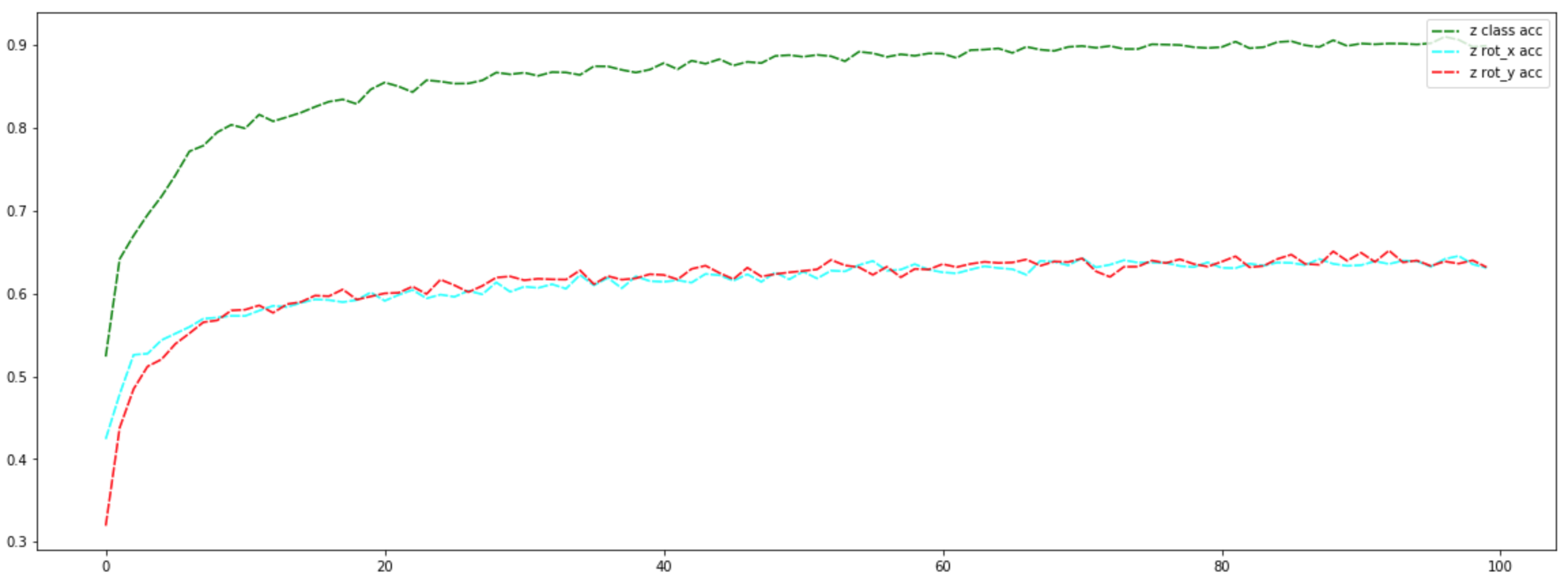}
        \caption{Information content in $z$ during training}
        \label{fig:VCCA_5_Info}
    \end{subfigure}
    \begin{subfigure}[t]{0.975\textwidth}  
        \centering 
        \includegraphics[width=\textwidth]{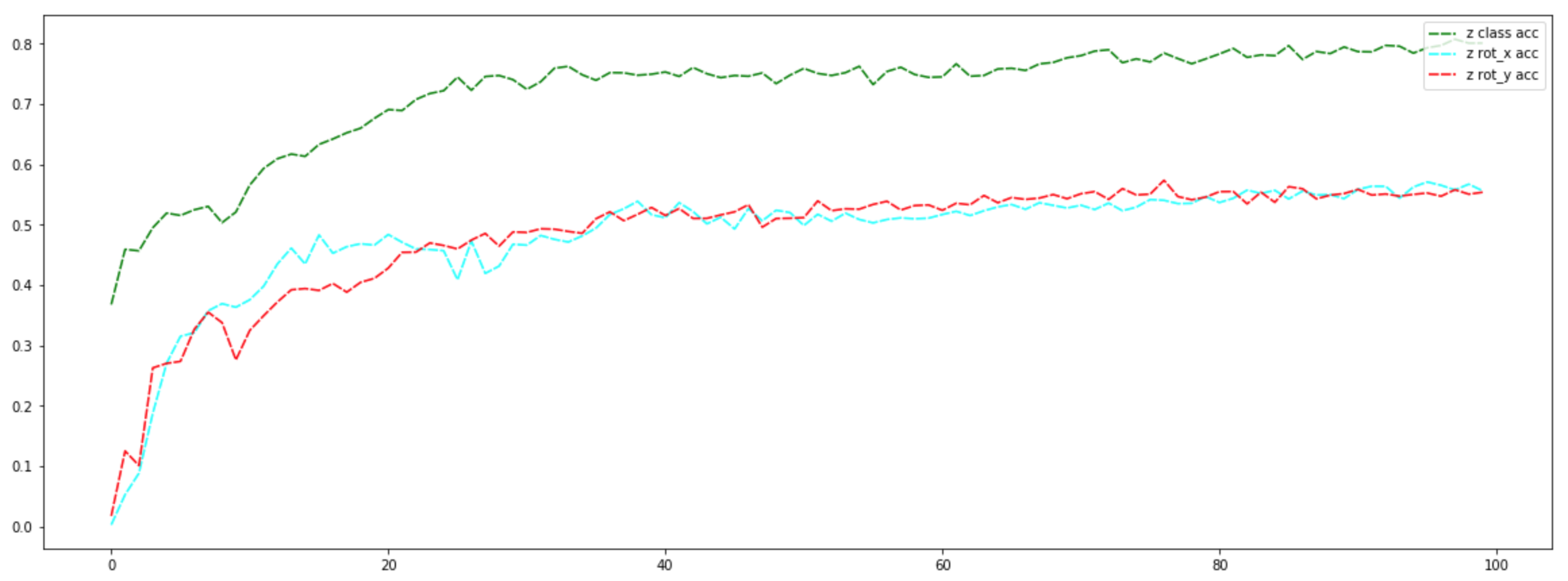}
        \caption{Information content in $z$ during training}
        \label{fig:ACCA_5_Info}
    \end{subfigure}
    \caption{VCCA and ACCA information curves during training on Tangled MNIST for 100 epochs with $z$-dim=5. Information content is with respect to each of the three known factors of variation: $\texttt{class}$, $\texttt{rot}_x$, and $\texttt{rot}_y$. Both models contain roughly the same \textit{proportion} of information (dominated by class information with view-specific rotation information lagging, but roughly equal). However, the \textit{overall} information content is higher for VCCA, as can be seen by the higher accuracy numbers. We believe this is due to the stronger regularizing effect of the discriminator in ACCA and explore this hypothesis in subsequent experiments.}
    \label{fig:51_Information}
\end{figure}

\begin{figure}[ht]
    \centering
    \begin{subfigure}[t]{0.975\textwidth}
        \centering
        \includegraphics[width=\textwidth]{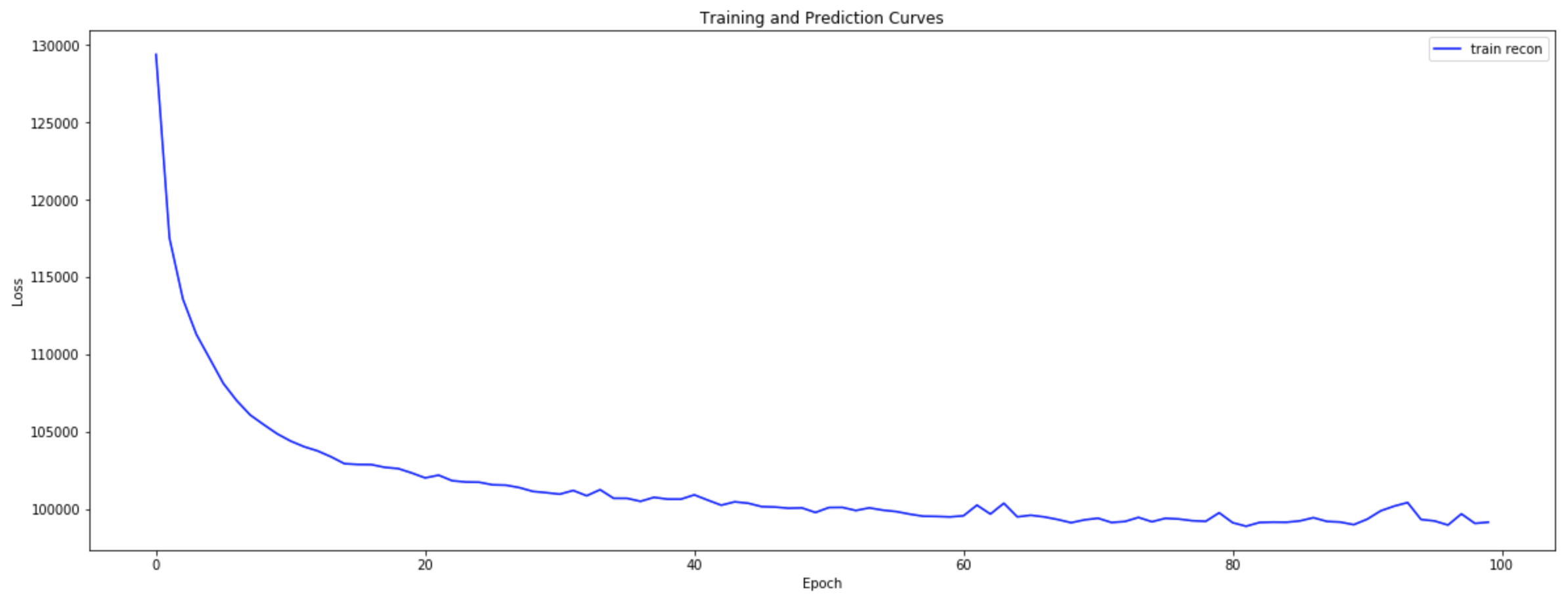}
        \caption{Loss during training for VCCA}
        \label{fig:VCCA_2_Loss}
    \end{subfigure}
    \begin{subfigure}[t]{0.975\textwidth}  
        \centering 
        \includegraphics[width=\textwidth]{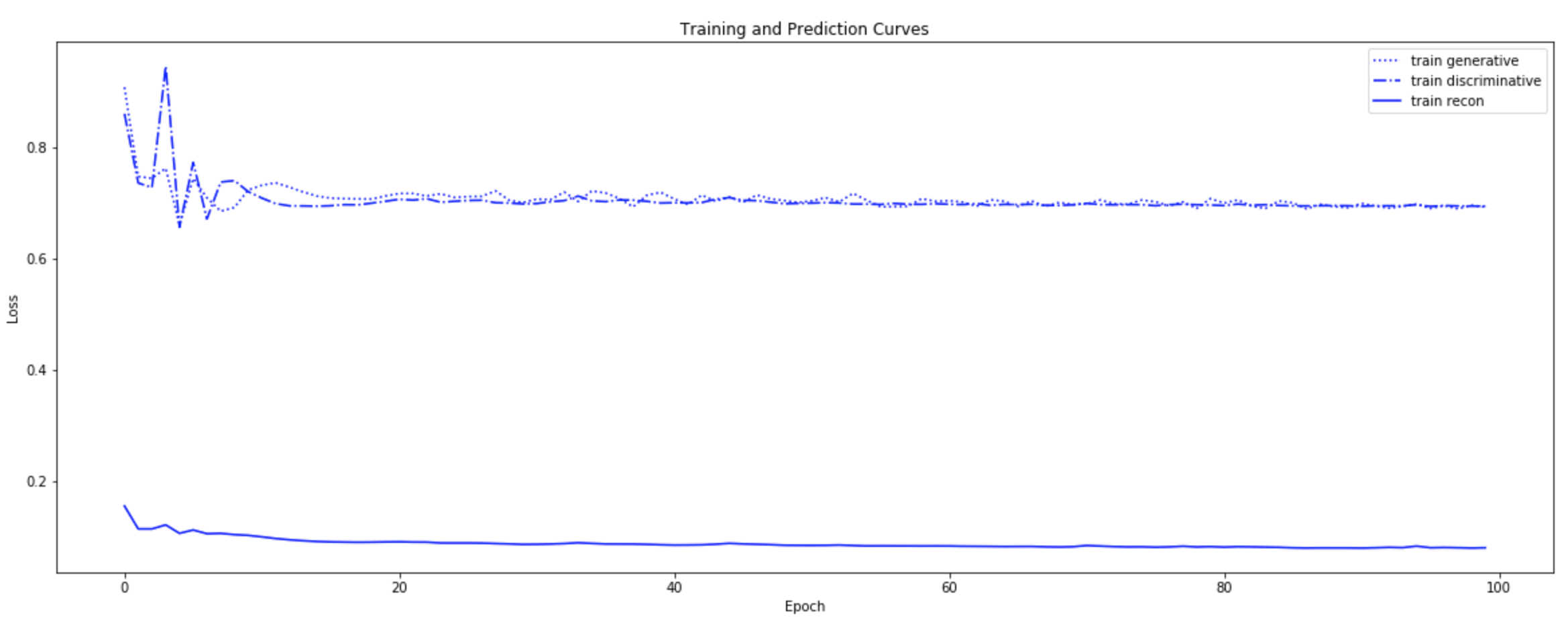}
        \caption{Losses during training for ACCA}
        \label{fig:ACCA_2_Loss}
    \end{subfigure}
    \caption{To explore the regularizing effects of the KL divergence term VCCA and the adversary in ACCA, we constrain the representational power of $z$ by dropping it to two dimensions to exaggerate the information bottleneck and competition between the representational power of $z$ and the match between $p(z)$ and $q(z|x,y)$. However, we also add two additional fully connected ReLU layers to the decoder to improve reconstruction power on the back end of the network. Shown here are the loss curves for this setup during training on Tangled MNIST for 100 epochs with $z$-dim=2.}
    \label{fig:51_Losses_2}
\end{figure}

\begin{figure}[ht]
    \begin{subfigure}[t]{0.975\textwidth}  
        \centering 
        \includegraphics[width=\textwidth]{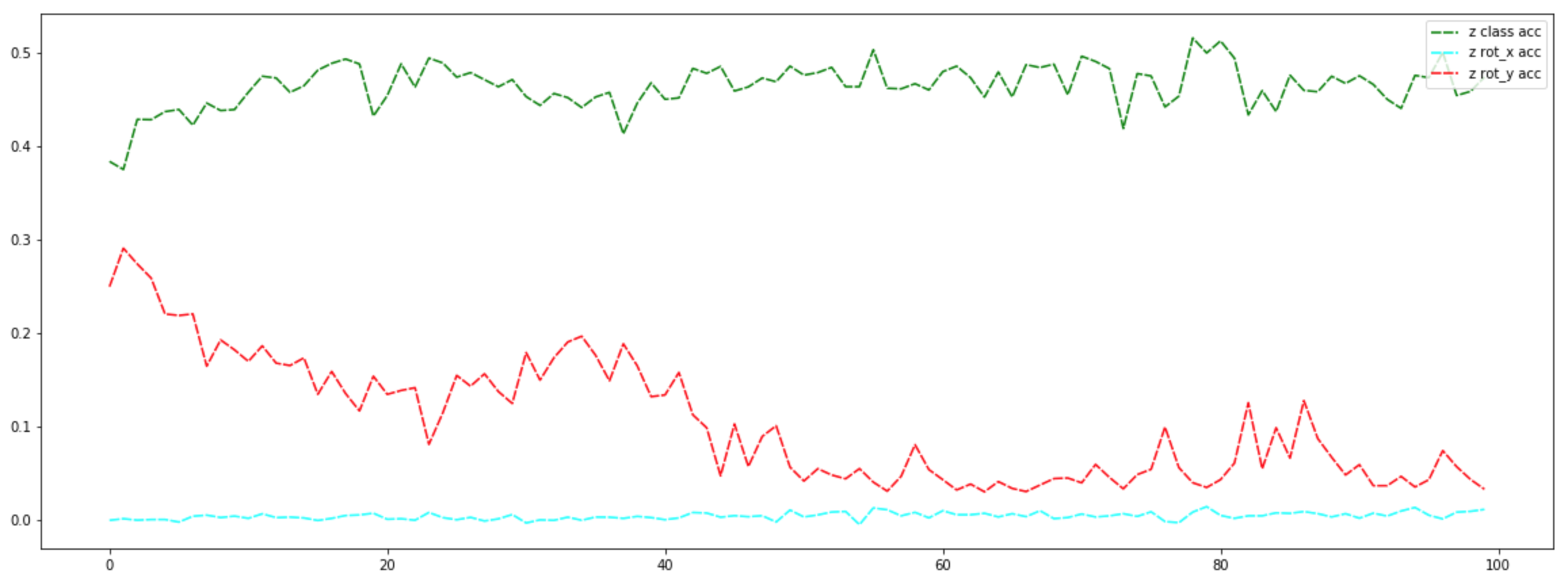}
        \caption{Information content in $z$ during training for VCCA}
        \label{fig:VCCA_2_Information}
    \end{subfigure}
    \begin{subfigure}[t]{0.975\textwidth}  
        \centering 
        \includegraphics[width=\textwidth]{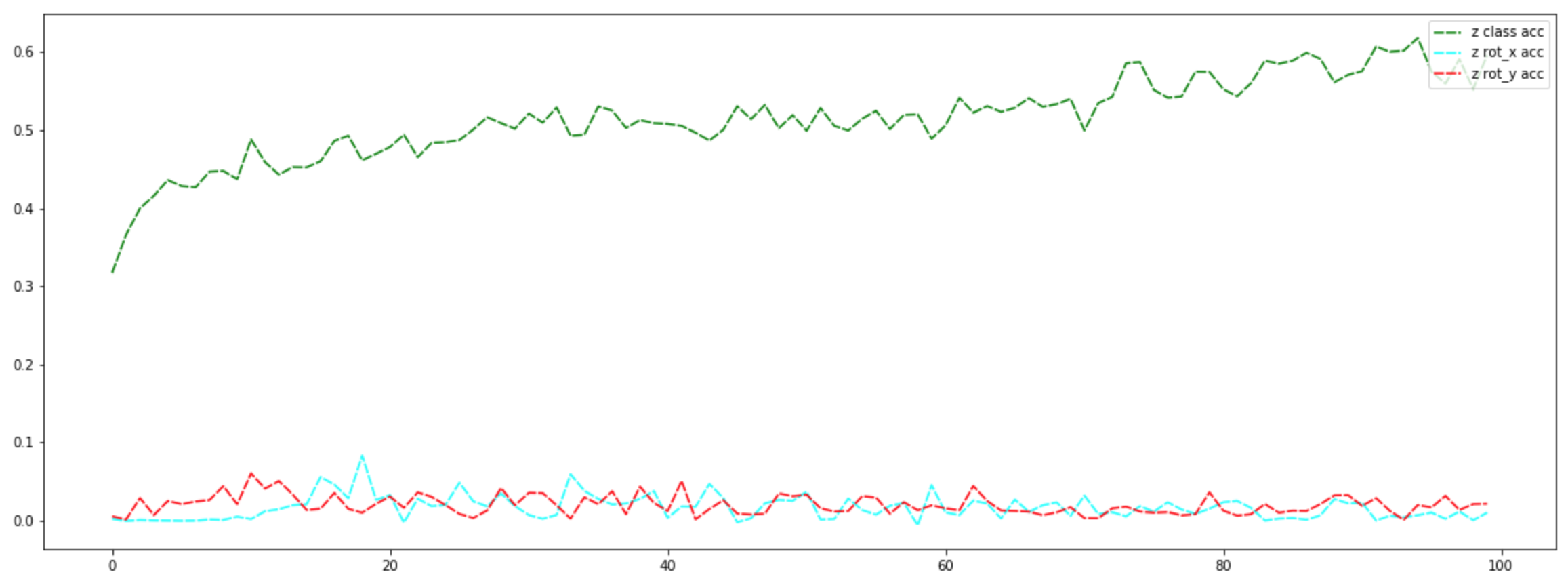}
        \caption{Information content in $z$ during training for ACCA}
        \label{fig:ACCA_2_Information}
    \end{subfigure}
    \caption{The information content in $z$ during training with $z$-dim=2 and two additional, fully connected ReLU decoder layers. As expected, there is some competition for information content at this size of representation. Under repeated experiments at this setting, we observed consistent volatility in the kind of information stored in $z$ for both ACCA and VCCA. In the runs shown, in VCCA, $z$ initially contains information about $\texttt{rot}_y$ and $\texttt{class}$, but sheds $\texttt{rot}_y$ in favor of class information. In the run shown for ACCA, class information dominates with very little $\texttt{rot}_y$ or $\texttt{rot}_x$ information. However, it is not clear which method globally preserves the \textit{most} overall information.}
    \label{fig:51_Information_2}
\end{figure}

\begin{figure}[ht]
    \centering
    \begin{subfigure}[t]{0.975\textwidth}
        \centering
        \includegraphics[width=\textwidth]{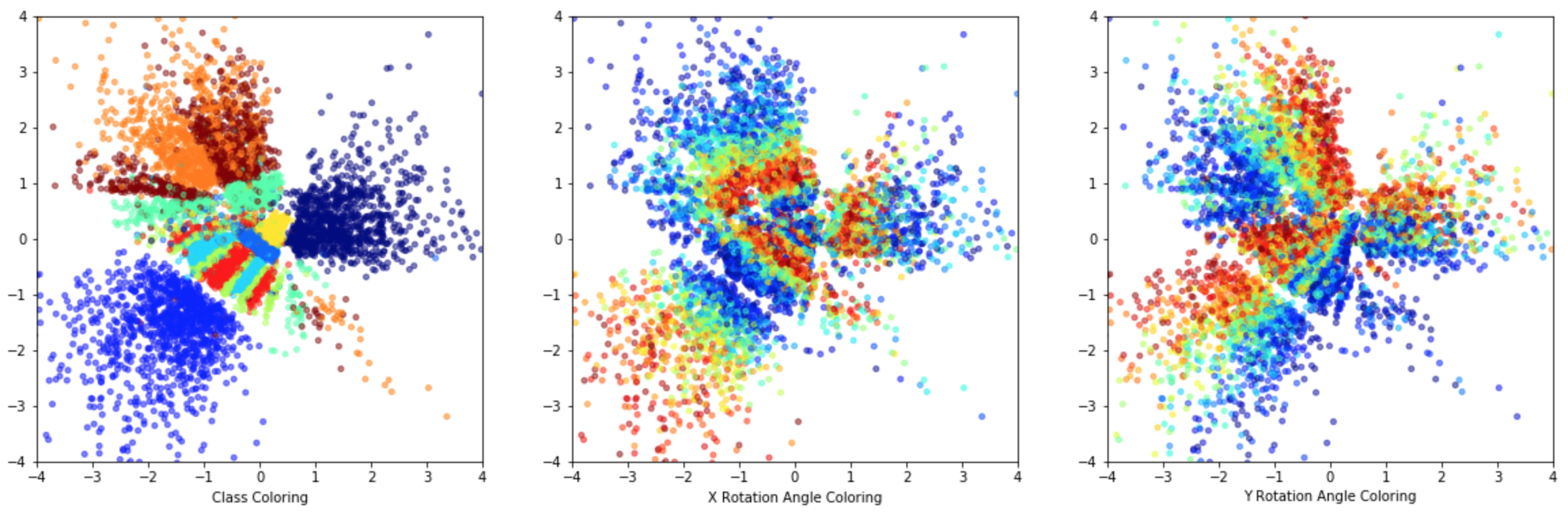}
        \caption{$z$ representations on training data for VCCA. Colorings correspond to class information (left), $\texttt{rot}_x$ information (middle) and $\texttt{rot}_y$ information (right).}
        \label{fig:VCCA_2_Embedding}
    \end{subfigure}
    \begin{subfigure}[t]{0.975\textwidth}  
        \centering 
        \includegraphics[width=\textwidth]{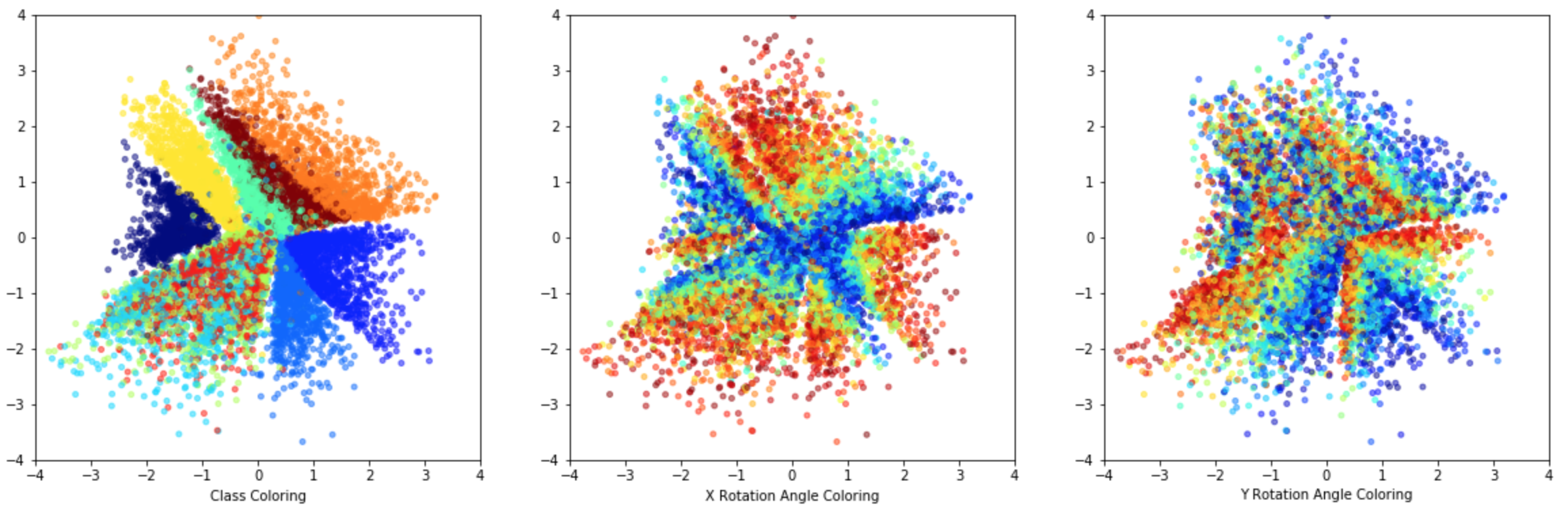}
        \caption{$z$ representations on training data for ACCA. Colorings correspond to class information (left), $\texttt{rot}_x$ information (middle) and $\texttt{rot}_y$ information (right).}
        \label{fig:ACCA_2_Embedding}
    \end{subfigure}
    \begin{subfigure}[t]{0.975\textwidth}  
        \centering 
        \includegraphics[width=\textwidth]{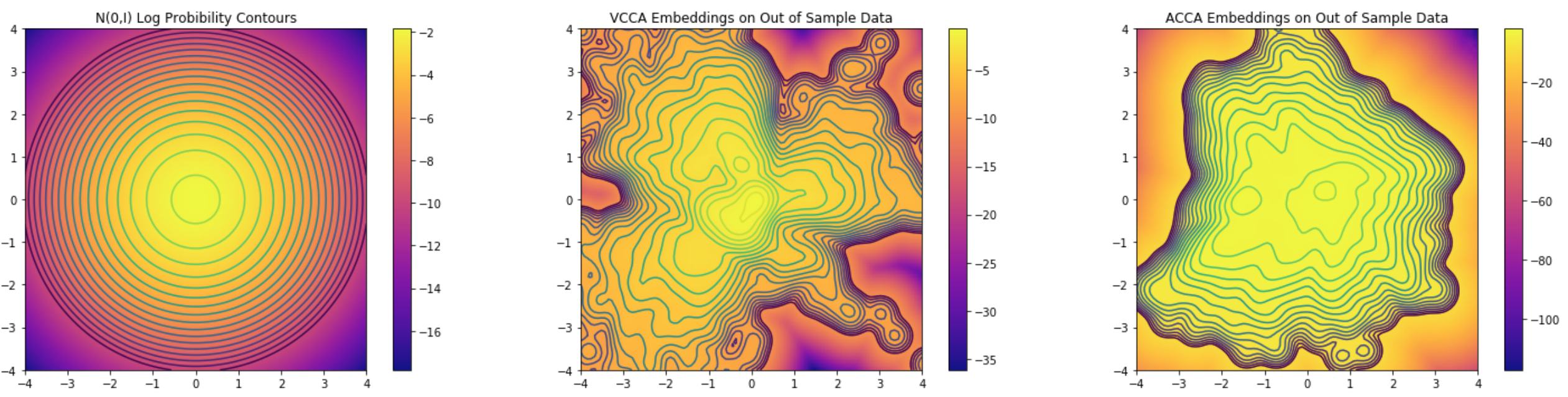}
        \caption{In the embeddings above, to highlight goodness of fit differences between VCCA and ACCA, we estimate the log probability densities for $q(z|x,y)$ for both ACCA and VCCA. We do this via Kernel Density Estimation. We use the scikit-learn library \cite{scikit-learn} with a Gaussian kernel with a bandwith of 0.2. In (a), we show the log probabilities of $p(z) \sim \mathcal{N}(0,I)$, in (b) we show the estimated log probabilities for $q(z|x,y)$ for VCCA, and in (c) we show the estimated log probabilities of $q(z|x,y)$ for ACCA. It is clear that the embeddings for ACCA are much closer to $p(z)$ than those for VCCA. A pattern we observed consistently is that the embeddings for VCCA allow ``fissures'' more consistently when class information (a categorical random variable) is predominately stored, while embeddings for ACCA rarely have gaps or holes. We believe this effect highlights the additional regularization power that the adversary of ACCA has over the KL Divergence loss term in VCCA. We also believe that the looser regularizer of VCCA is what allows higher classification scores on the learned representations since class boundaries are more separable - something often desired in practice, even though it comes at the expense of higher deviations of $q(z|x,y)$ from $p(z)$.}
        \label{fig:2d_Goodness}
    \end{subfigure}
    \caption{$z$ embeddings for VCCA and ACCA with $z$-dim=2 and two additional ReLU fully connected layers in the decoder. Notice the tighter fit that ACCA gives between $q(z|x,y)$ and $p(z) \sim \mathcal{N}(0,I)$. This is behavior we consistently observed when the information represented is categorical in nature. While this looser regularization may sometimes be desirable in practice, this experiment highlights the additional regularization power of ACCA, such that \textit{even categorical} information rarely contains ``holes'' or ``fissures'' in $q(z|x,y)$.}
    \label{fig:51_Embeddings_2}
\end{figure}

\begin{figure}[ht]
    \centering
    \begin{subfigure}[t]{0.475\textwidth}
        \centering
        \includegraphics[width=\textwidth]{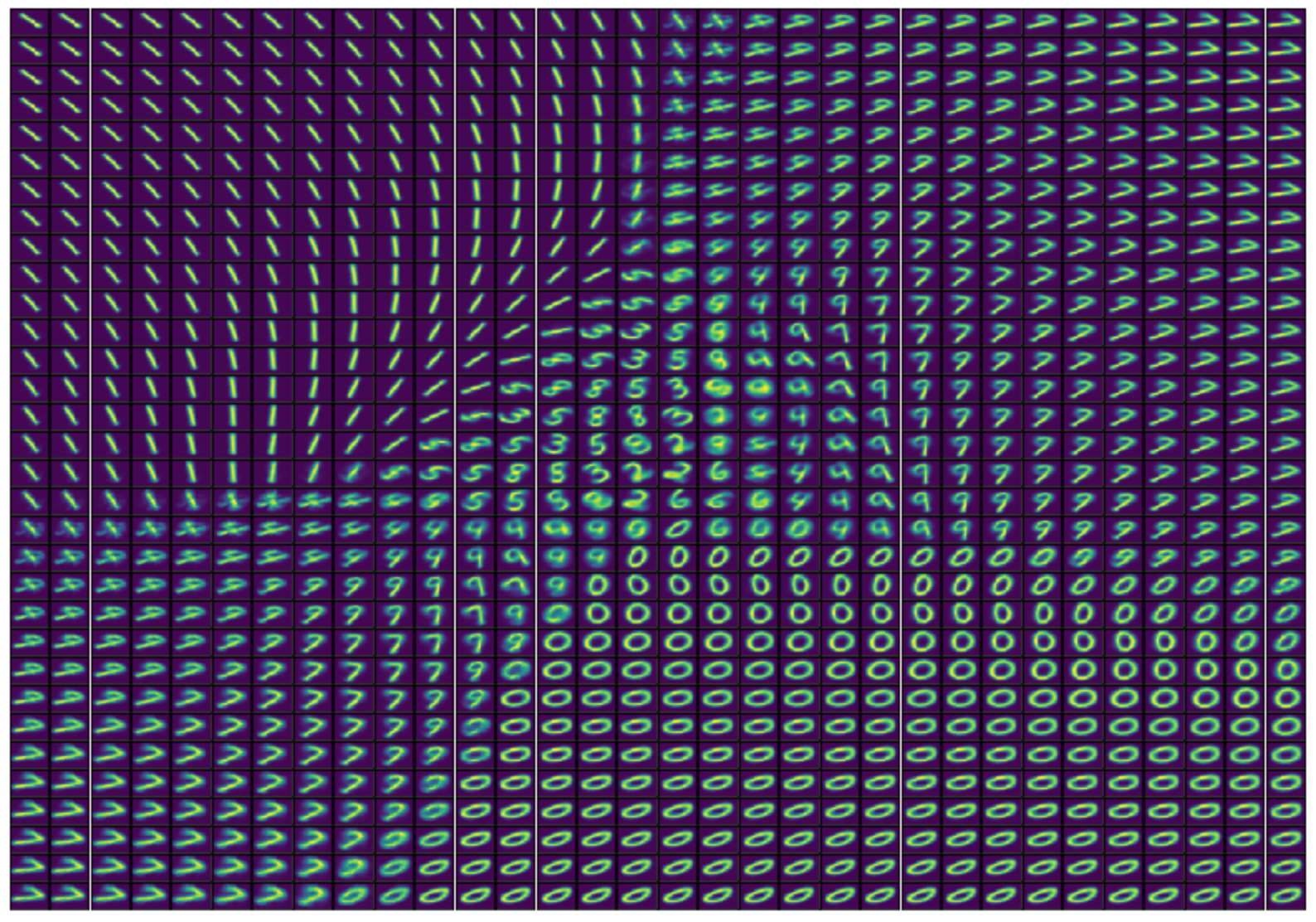}
        \caption{View $x$ generations in VCCA}
        \label{fig:VCCA_2_Generations_x}
    \end{subfigure}
    \begin{subfigure}[t]{0.475\textwidth}  
        \centering 
        \includegraphics[width=\textwidth]{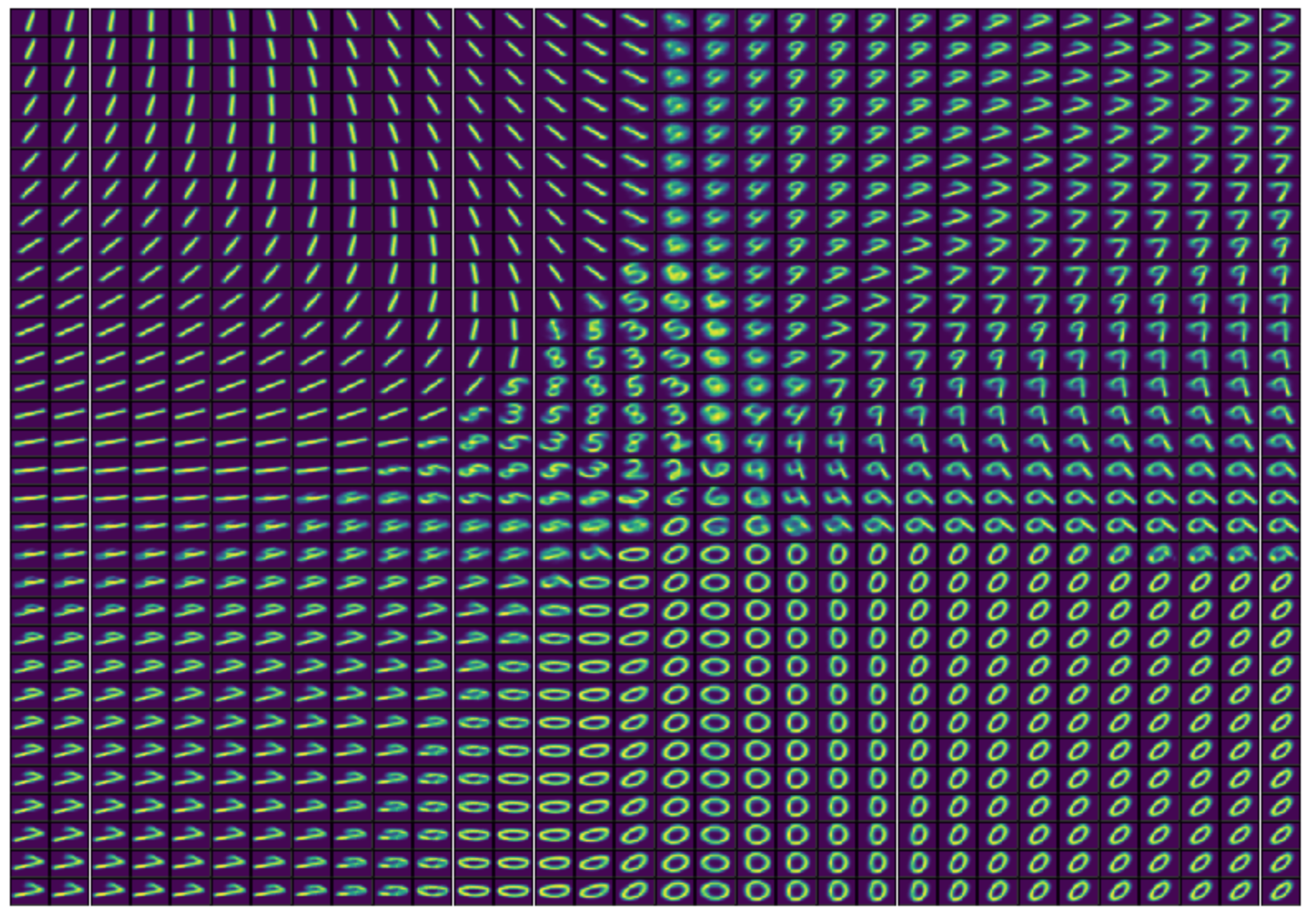}
        \caption{View $y$ generations in VCCA}
        \label{fig:VCCA_2_Generations_y}
    \end{subfigure}
    \begin{subfigure}[t]{0.475\textwidth}  
        \centering 
        \includegraphics[width=\textwidth]{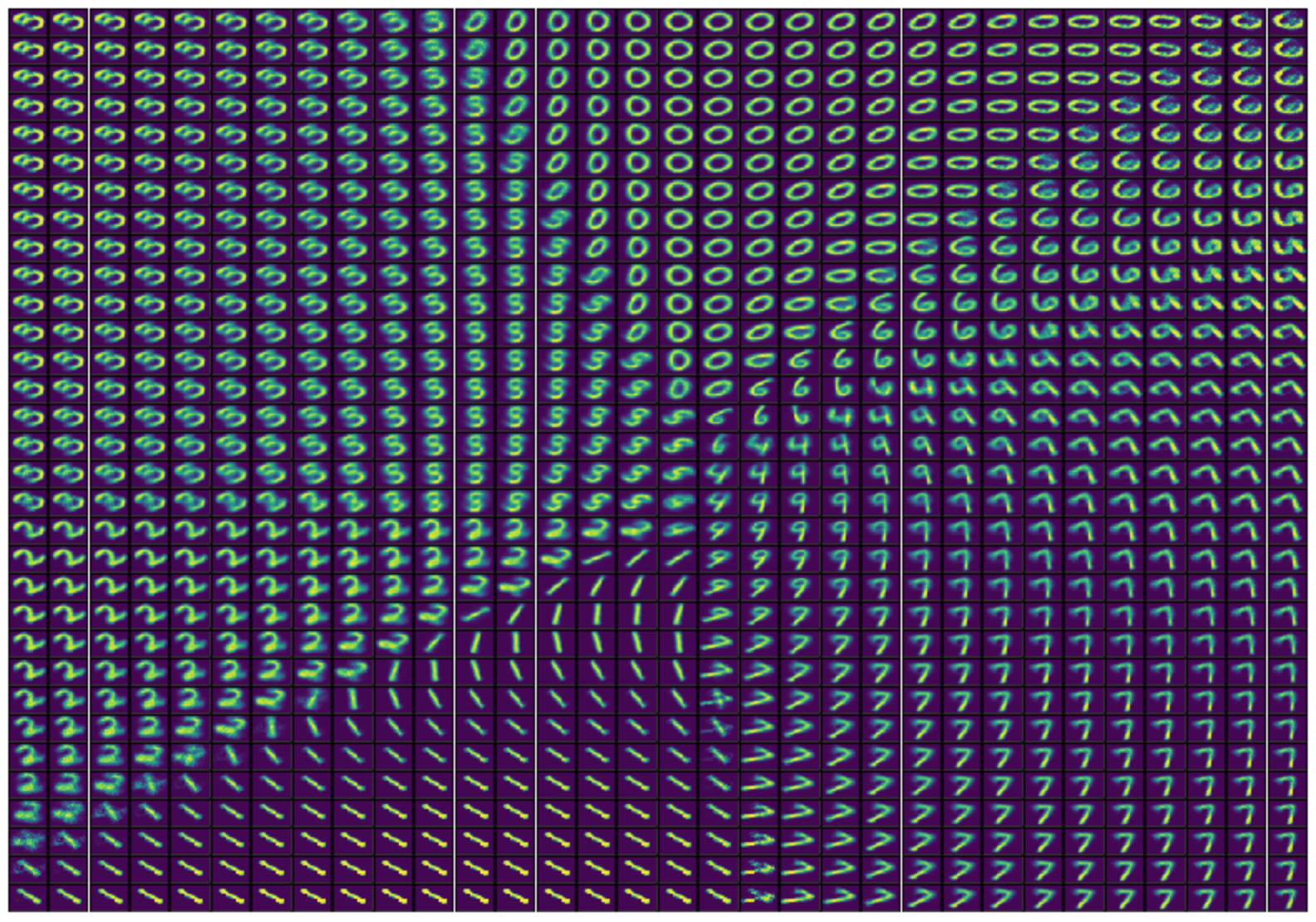}
        \caption{View $x$ generations in ACCA}
        \label{fig:ACCA_2_Generations_x}
    \end{subfigure}
    \begin{subfigure}[t]{0.475\textwidth}  
        \centering 
        \includegraphics[width=\textwidth]{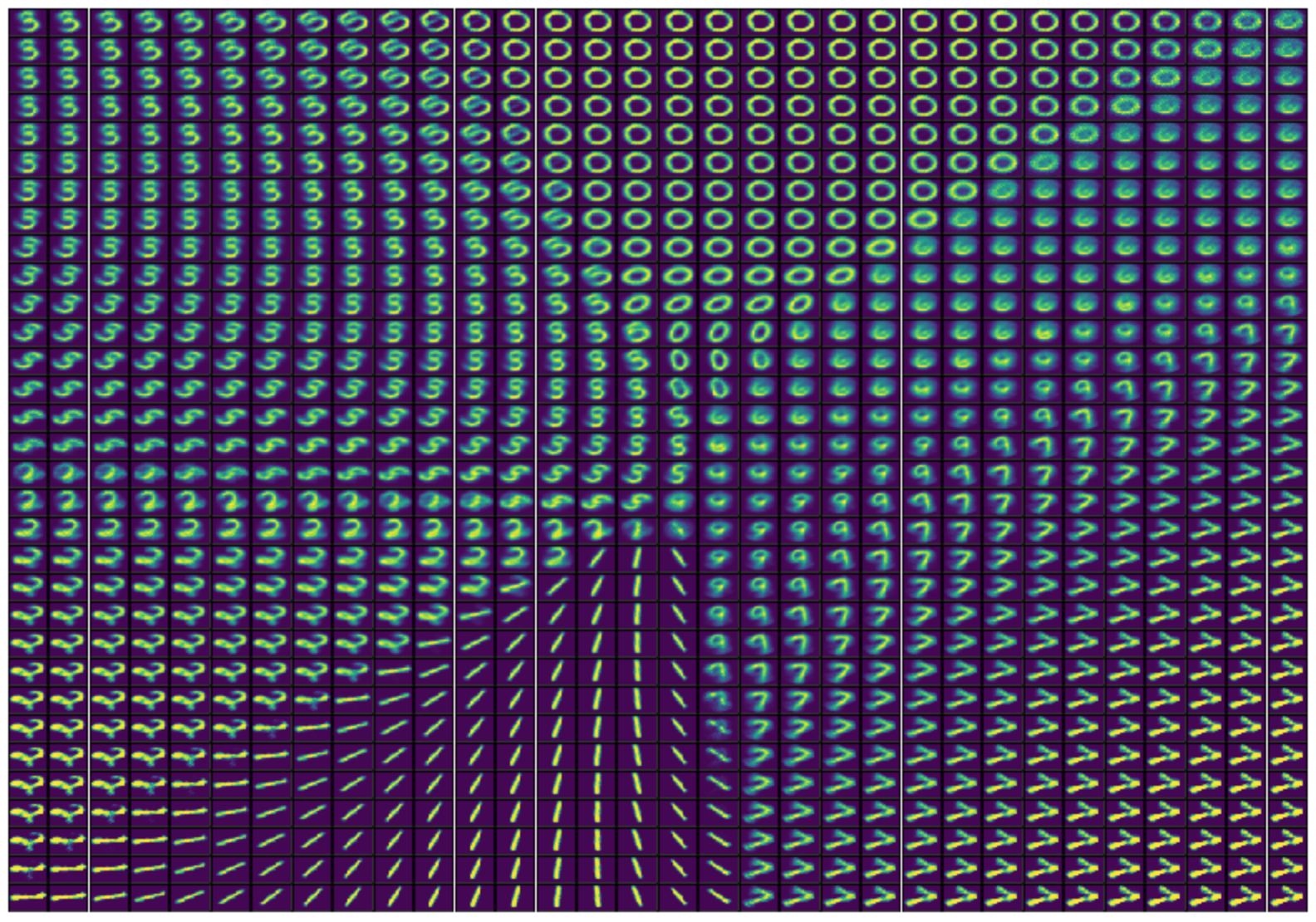}
        \caption{View $y$ generations in ACCA}
        \label{fig:ACCA_2_Generations_y}
    \end{subfigure}
    \caption{Here we explore the information content in $z$ as a function of the location for $z$-dim=2 and two additional, fully connected ReLU layers in the decoder. We walk $z$ over grid $(-4, 4) \times (-4,4)$ with step size 0.25 and decode the center of each cell into generations for views $x$ and $y$.  Subfigures (a) and (b) show resulting generations for views $x$ and $y$, respectively, for VCCA.  Subfigures (c) and (d) show resulting generations for views $x$ and $y$, respectively, for ACCA. It is clear that, in both VCCA and ACCA, class information is clustered in $z$ and rotation and style information act as intraclass coordinates within each cluster, as can be seen in the colorings from the previous figure.}
    \label{fig:51_Generations_2}
\end{figure}

\clearpage
\subsection{Additional Plots from Section 5.2}
\begin{figure}[ht]
    \centering
    \begin{subfigure}[t]{0.475\textwidth}
        \centering
        \includegraphics[width=\textwidth]{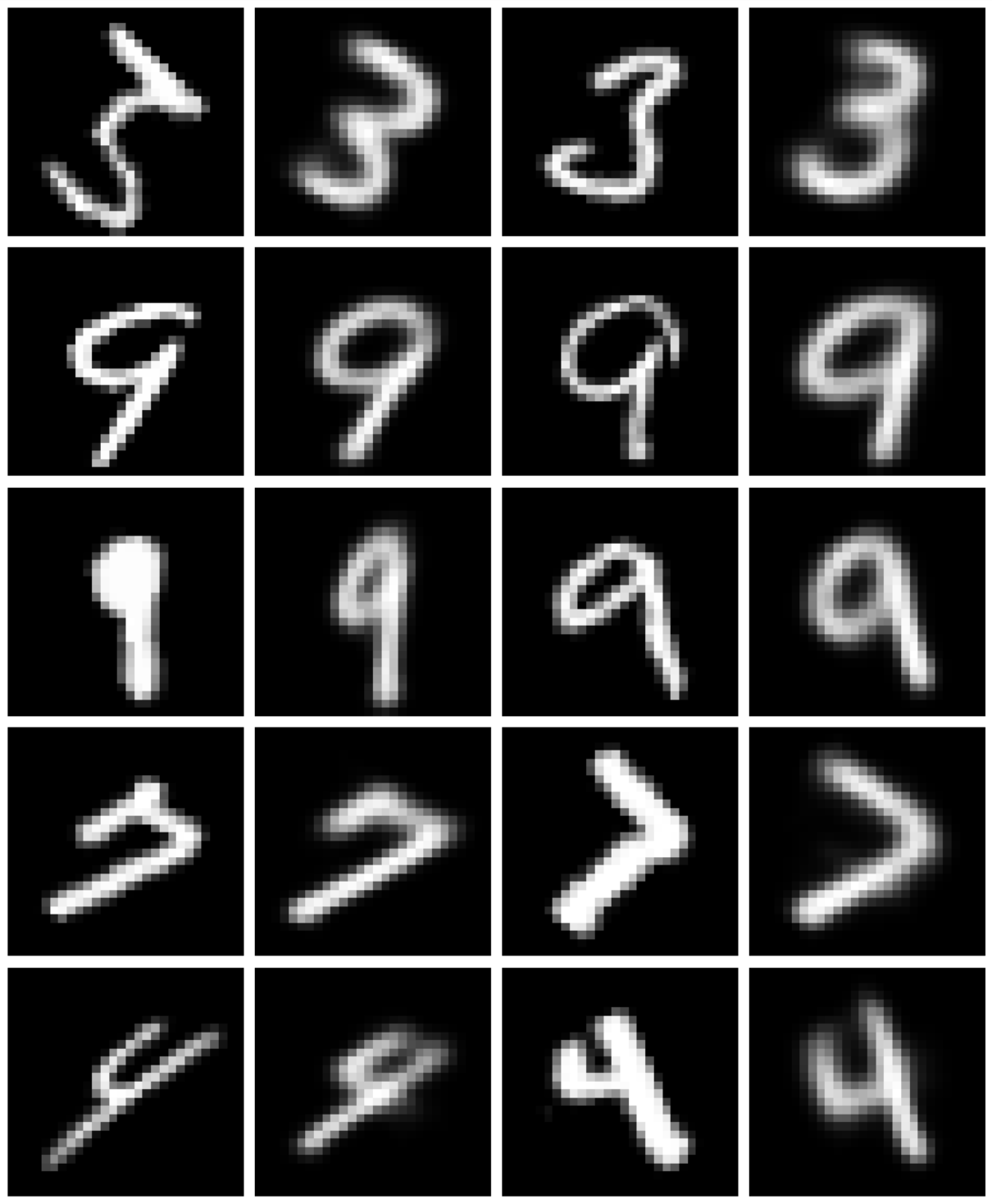}
        \caption{VCCA-Private}
        \label{fig:VCCA_Private2_Reconstructions}
    \end{subfigure}
    \quad
    \begin{subfigure}[t]{0.475\textwidth}  
        \centering 
        \includegraphics[width=\textwidth]{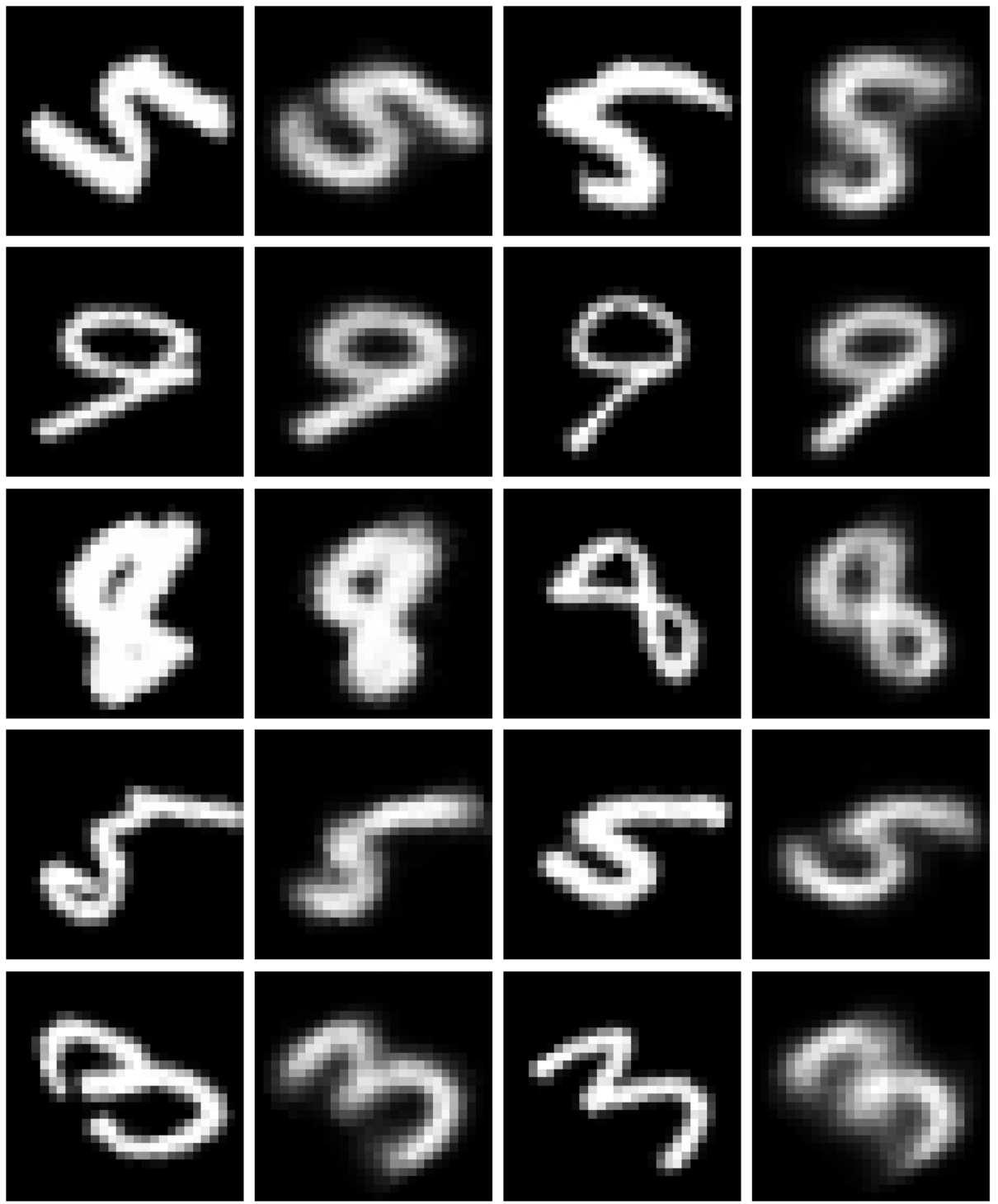}
        \caption{ACCA-Private}
        \label{fig:ACCA_Private2_Reconstructions}
    \end{subfigure}
    \caption{5 random reconstructions for VCCA-Private and ACCA-Private with $z$-dim=$h_x$-dim=$h_y$-dim=2. Each column is, respectively, $x$, $\hat{x}$, $y$, and $\hat{y}$. There is one additional degree of freedom in this representation since there are 5 underlying factors of variation and 6 dimensions for representation.}
    \label{fig:52_Reconstructions}
\end{figure}

\begin{figure}[ht]
    \centering
    \begin{subfigure}[t]{0.975\textwidth}
        \centering
        \includegraphics[width=\textwidth]{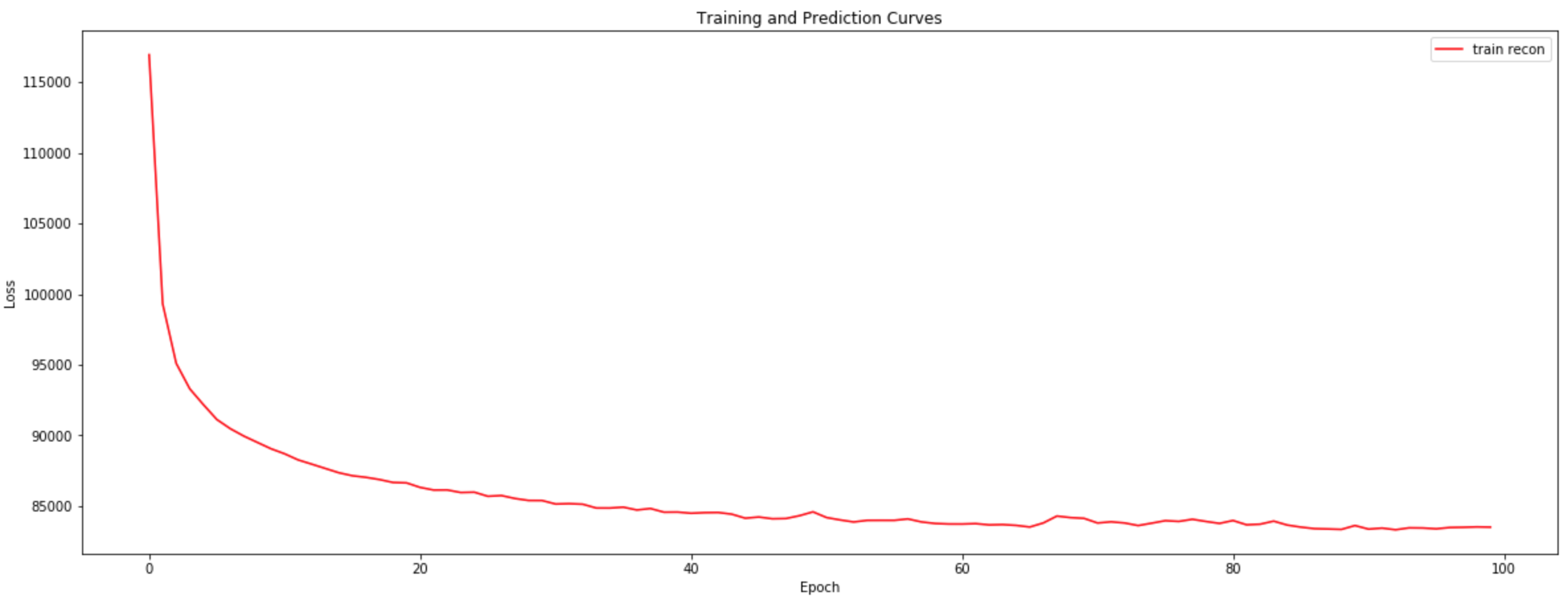}
        \caption{VCCA-Private}
        \label{fig:VCCA_Private2_Loss}
    \end{subfigure}
    \begin{subfigure}[t]{0.975\textwidth}  
        \centering 
        \includegraphics[width=\textwidth]{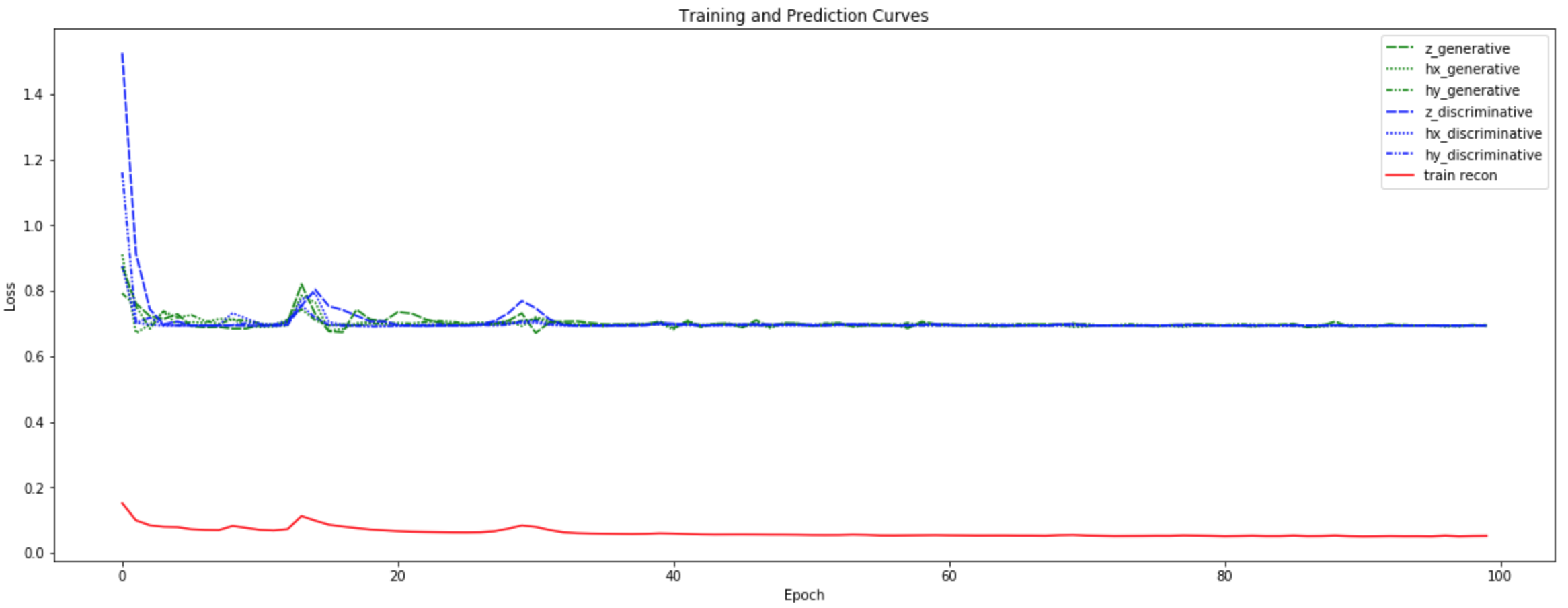}
        \caption{ACCA-Private}
        \label{fig:ACCA_Private2_Loss}
    \end{subfigure}
    \caption{Loss curves during training for VCCA-Private and ACCA-Private on Tangled MNIST with $z$-dim=$h_x$-dim=$h_y$-dim=2. The adversarial games of ACCA-Private all converge by roughly epoch 33. View the loss curves from ACCA-Private next to its corresponding information content curves from Fig. \ref{fig:ACCA_Private2_Information} to see how the blips in each of the adversarial games being played correspond to a reshuffling of information content.}
    \label{fig:52_Losses}
\end{figure}

\begin{figure}[ht]
    \begin{subfigure}[t]{0.975\textwidth}  
        \centering 
        \includegraphics[width=\textwidth]{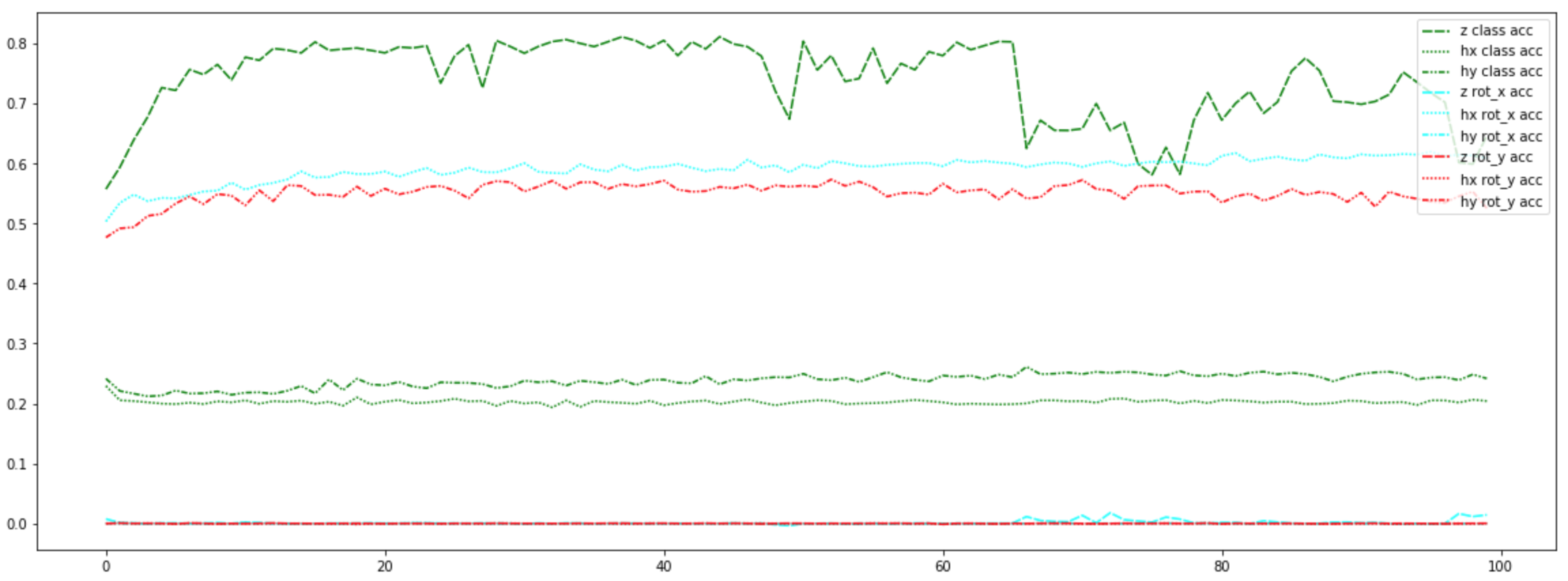}
        \caption{VCCA-Private}
        \label{fig:VCCA_Private2_Information}
    \end{subfigure}
    \begin{subfigure}[t]{0.975\textwidth}  
        \centering 
        \includegraphics[width=\textwidth]{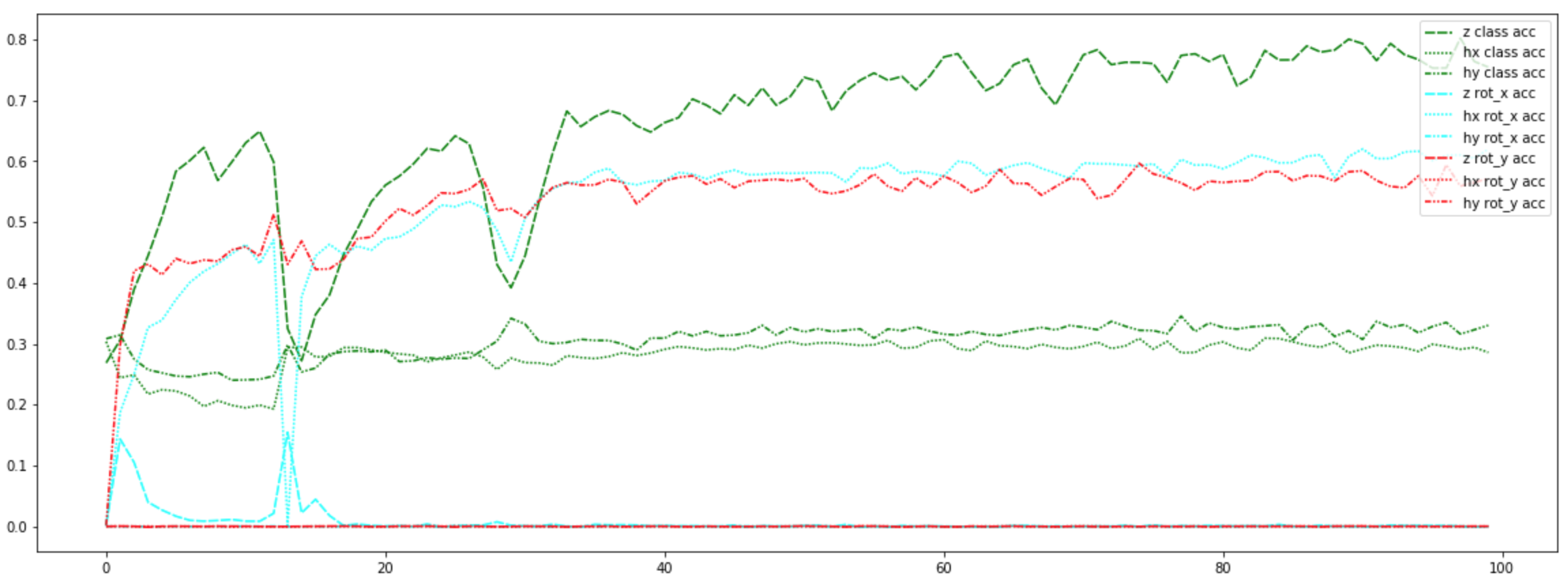}
        \caption{ACCA-Private}
        \label{fig:ACCA_Private2_Information}
    \end{subfigure}
    \caption{Information content during training for VCCA-Private and ACCA-Private on Tangled MNIST with $z$-dim=$h_x$-dim=$h_y$-dim=2. Colors correspond to information type (class vs $\texttt{rot}_x$ vs $\texttt{rot}_y$) and line style corresponds to representation type ($z$ vs $h_x$ vs $h_y$). A few things to observe: VCCA behavior is much more stable over training than ACCA owing to the more complicated training procedure and time required for the adversarial game between the encoder and discriminator to converge (see equation \ref{eq:adversarial_game}). Subfigure (b) should be seen alongside Fig. \ref{fig:ACCA_Private2_Loss} to see how blips in the losses for ACCA-Private correspond to information reshuffling between representations. The three highest curves in each figure show that class information is primarily going to $z$, $\texttt{rot}_x$ information is primarily going to $h_x$ and $\texttt{rot}_y$ information is primarily going to $h_y$. This is good disentangling behavior under the paradim we describe in section 2 for both ACCA and VCCA. However, from these figures, we can also observe the degree of class information spillover into $h_x$ and $h_y$ (roughly 23\% accuracy in VCCA for both and roughly 30\% accuracy in ACCA for both. This is not good multiview disentangling behavior under our paradigm because we seek to have no information overlap in our representations.}
    \label{fig:52_Information}
\end{figure}

\begin{figure}[ht]
    \centering
    \begin{subfigure}[t]{0.975\textwidth}
        \centering
        \includegraphics[width=\textwidth]{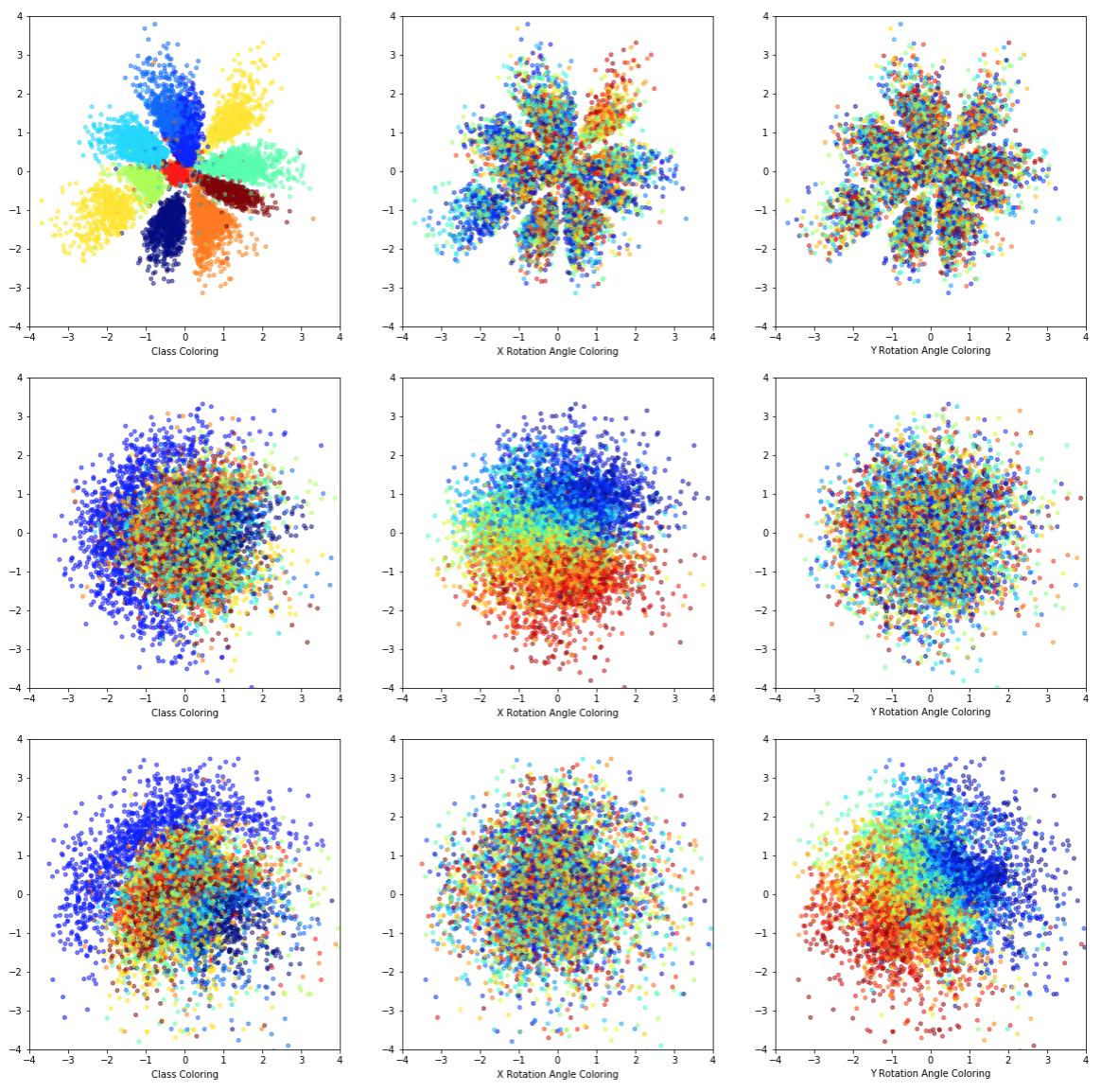}
        \caption{Embeddings from VCCA-Private, colored by an independent factor of variation. Each row is the embedded representation of training data from Tangled MNIST for $z$, $h_x$, and $h_y$, respectively. Column 1 is colored with class information, column 2 is colored with $\texttt{rot}_x$ information and column 3 is colored with $\texttt{rot}_y$ information. Ideally, we want the plots on the diagonal to be clearly separable by color so that the learned representations correspond to the desired information content. It is clear from these plots that VCCA-Private does disentangle at this view level (as distinguished from dimensional disentangling): class information is clearly separable in $z$, $\texttt{rot}_x$ information is clearly separable in $h_x$, and $\texttt{rot}_y$ is clearly separable in $h_y$. However, it is also clear that additional future work is needed because there is clear information spill over: both $\texttt{rot}_x$ and $\texttt{rot}_y$ contain a decent amount of class information.}
        \label{fig:VCCA_Private2_Embeddings}
    \end{subfigure}
    \begin{subfigure}[t]{0.975\textwidth}  
        \centering 
        \includegraphics[width=\textwidth]{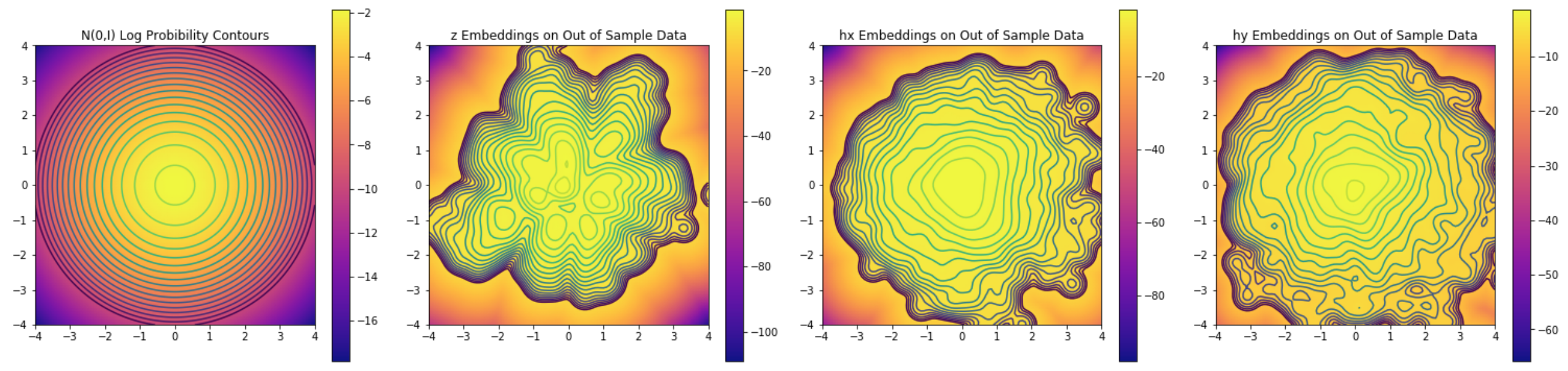}
        \caption{VCCA-Private log probabilities for $p(z) = p(h_x) = p(h_y) \sim \mathcal{N}(0,I)$, $q(z|x,y)$, $q(h_x|x)$ and $q(h_y|y)$. The densities for the learned representations were estimated using Kernel Density Estimation from scikit-learn \cite{scikit-learn} using a Gaussian kernel with bandwidth of 0.2. In $q(z|x,y)$, we can once again see that VCCA allows large fissures in learned representations when the underlying distribution is categorical. We previously discussed this is Fig. 11. However, learned representations for continuous information, such as $\texttt{rot}_x$ and $\texttt{rot}_y$ fit $p(h_x)$ and $p(h_y)$ well.}
        \label{fig:VCCA_Private2_Goodness}
    \end{subfigure}
    \caption{VCCA-Private embeddings of training data from Tangled MNIST with $z$-dim=$h_x$-dim=$h_y$-dim=2.}
    \label{fig:52_VCCA_Embeddings}
\end{figure}

\begin{figure}[ht]
    \begin{subfigure}[t]{0.975\textwidth}  
        \centering 
        \includegraphics[width=\textwidth]{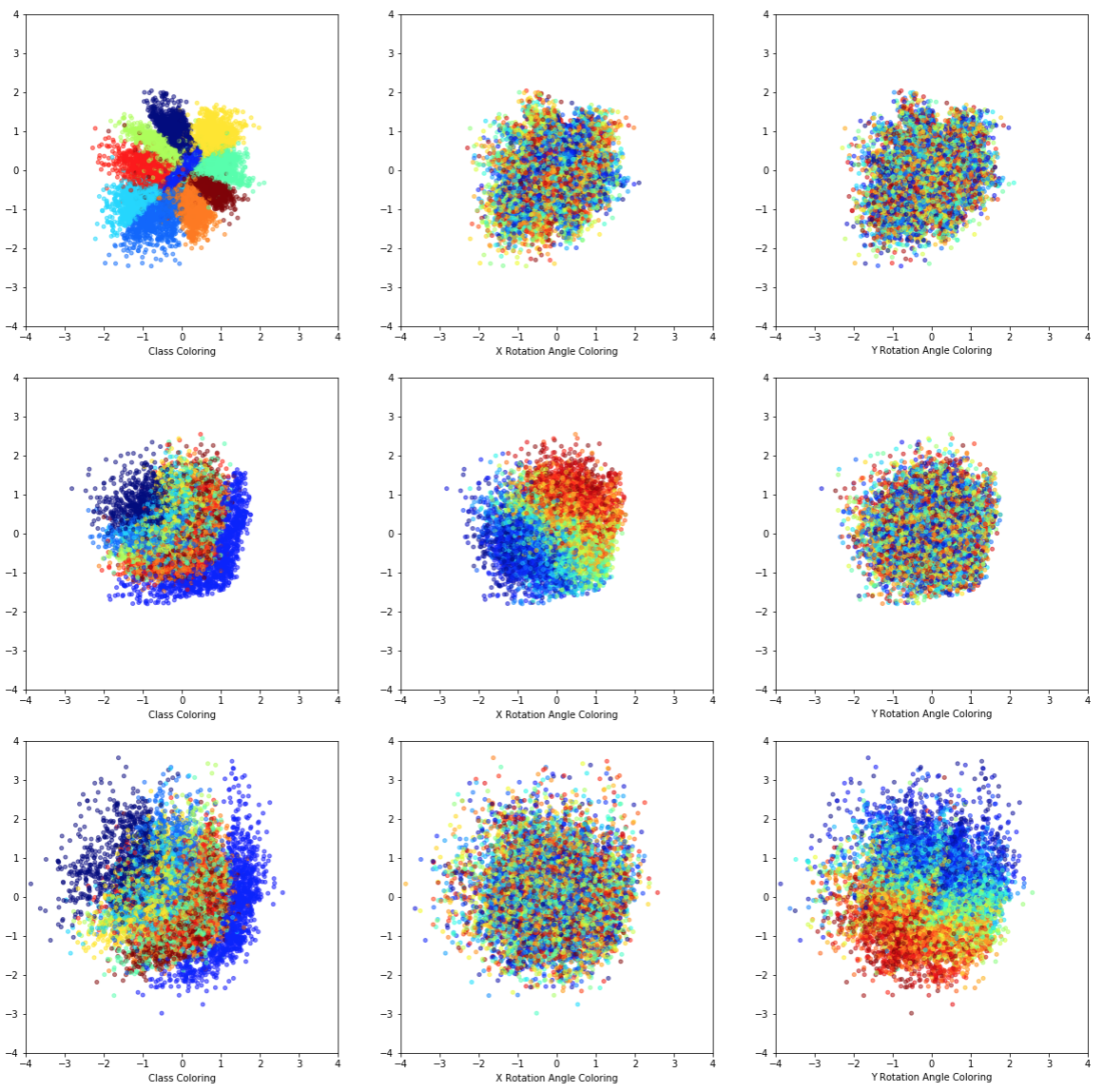}
        \caption{Embeddings from ACCA-Private, colored by an independent factor of variation. Each row is the embedded representation of training data from Tangled MNIST for $z$, $h_x$, and $h_y$, respectively. Column 1 is colored with class information, column 2 is colored with $\texttt{rot}_x$ information and column 3 is colored with $\texttt{rot}_y$ information. We observe similar view level disentangling behavior to VCCA. Class information is clearly separable in $z$, $\texttt{rot}_x$ information is clearly separable in $h_x$, and $\texttt{rot}_y$ information is clearly separable in $h_y$. We also see similar spill over: class information does show up in $h_x$ and $h_y$, so future work is needed for more faithful disentangling. We also observe the stronger regularization ability of ACCA over VCCA here in $q(z|x,y)$, not allowing any holes or gaps in the representation because they would be exploited by the adversary.}
        \label{fig:ACCA_Private2_Embeddings}
    \end{subfigure}
    \begin{subfigure}[t]{0.975\textwidth}  
        \centering 
        \includegraphics[width=\textwidth]{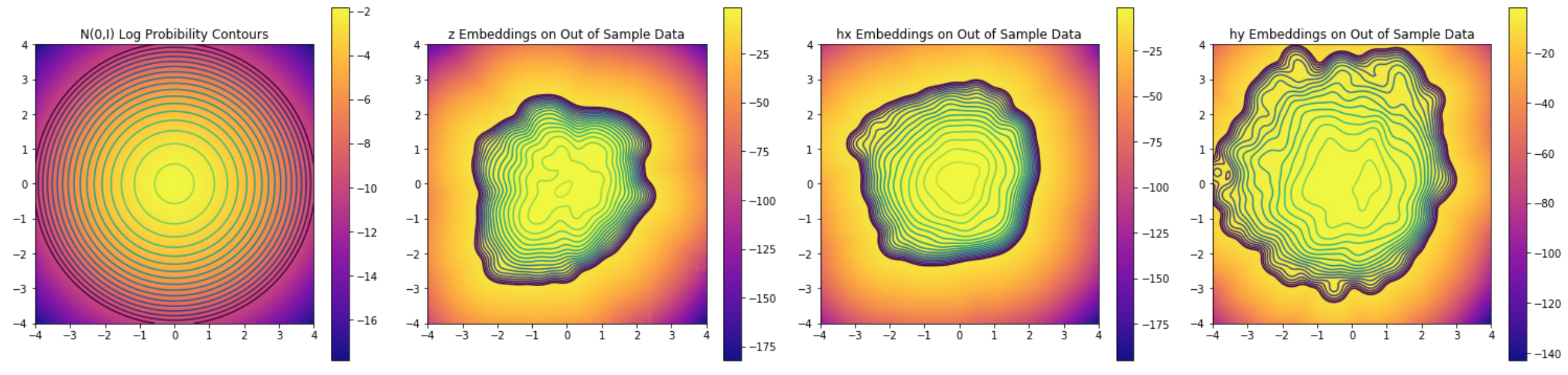}
        \caption{ACCA-Private log probabilities for $p(z) = p(h_x) = p(h_y) \sim \mathcal{N}(0,I)$, $q(z|x,y)$, $q(h_x|x)$ and $q(h_y|y)$. The densities for the learned representations were estimated using Kernel Density Estimation from Scikit-learn \cite{scikit-learn} using a Gaussian kernel with bandwidth of 0.2. In $q(z|x,y)$, we can further see the regularization power of the adversaries and the closeness between $p(z)$ and $q(z|x,y)$.}
        \label{fig:ACCA_Private2_Goodness}
    \end{subfigure}
    \caption{ACCA-Private embeddings of training data from Tangled MNIST with $z$-dim=$h_x$-dim=$h_y$-dim=2.}
    \label{fig:52_ACCA_Embeddings}
\end{figure}

\begin{figure}[ht]
    \centering
    \begin{subfigure}[t]{0.975\textwidth}
        \centering
        \includegraphics[width=\textwidth]{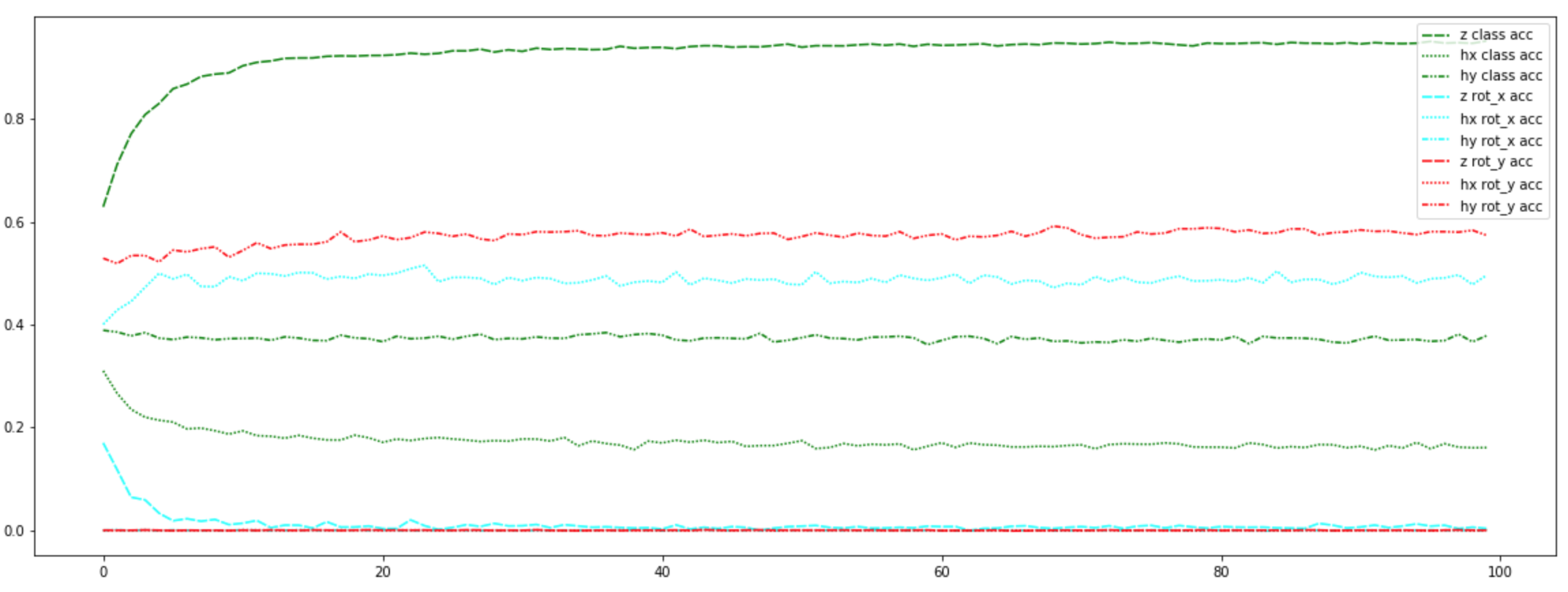}
        \caption{VCCA-Private}
        \label{fig:VCCA_Private4_Information}
    \end{subfigure}
    \begin{subfigure}[t]{0.975\textwidth}  
        \centering 
        \includegraphics[width=\textwidth]{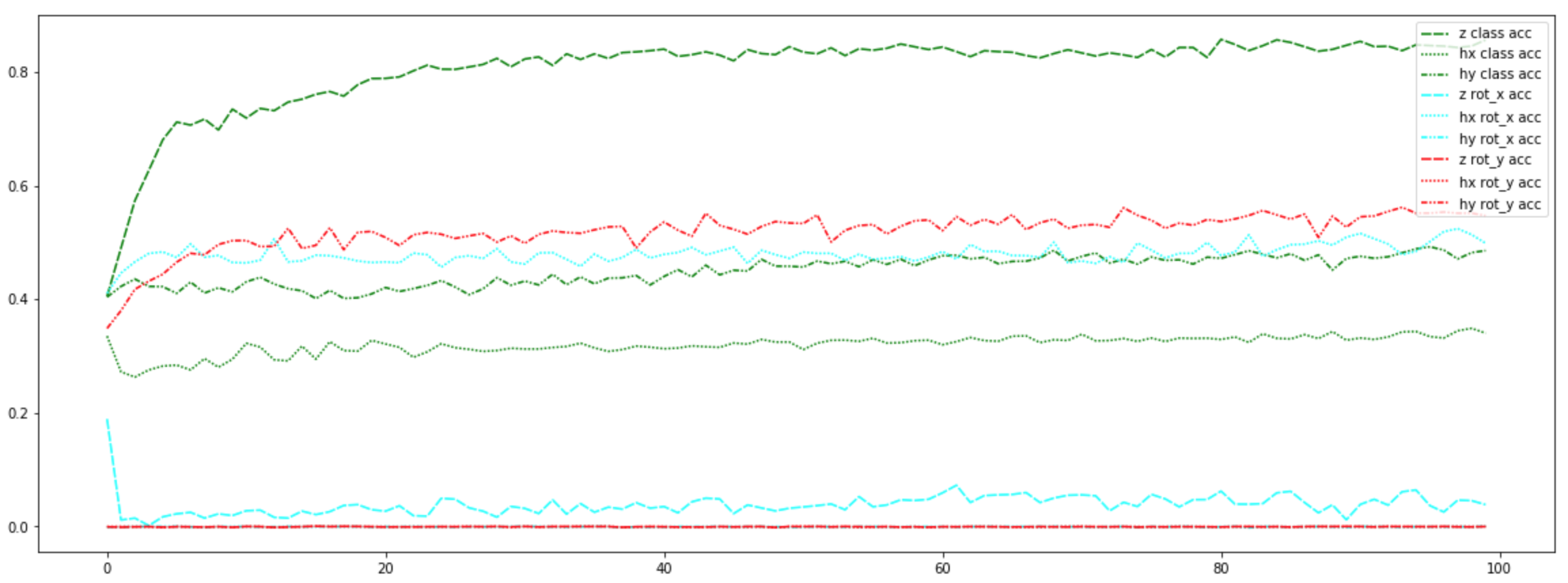}
        \caption{ACCA-Private}
        \label{fig:ACCA_Private4_Information}
    \end{subfigure}
    \caption{Here we keep the network structure the same as the previous experiments with VCCA-Private and ACCA-Private but increase the size of each representation in order to increase the information capacity of the representations: $z$-dim=$h_x$-dim=$h_y$-dim=4.  We observe less chaotic information reshuffling (\textit{between} representations, not necessarily \textit{within} representations) during training and higher overall accuracies.}
    \label{fig:52_Information_4}
\end{figure}

\clearpage
\subsection{Additional Plots from Section 5.3}

\begin{figure}[H]
    \centering
    \begin{subfigure}[t]{0.475\textwidth}
        \centering
        \includegraphics[width=\textwidth]{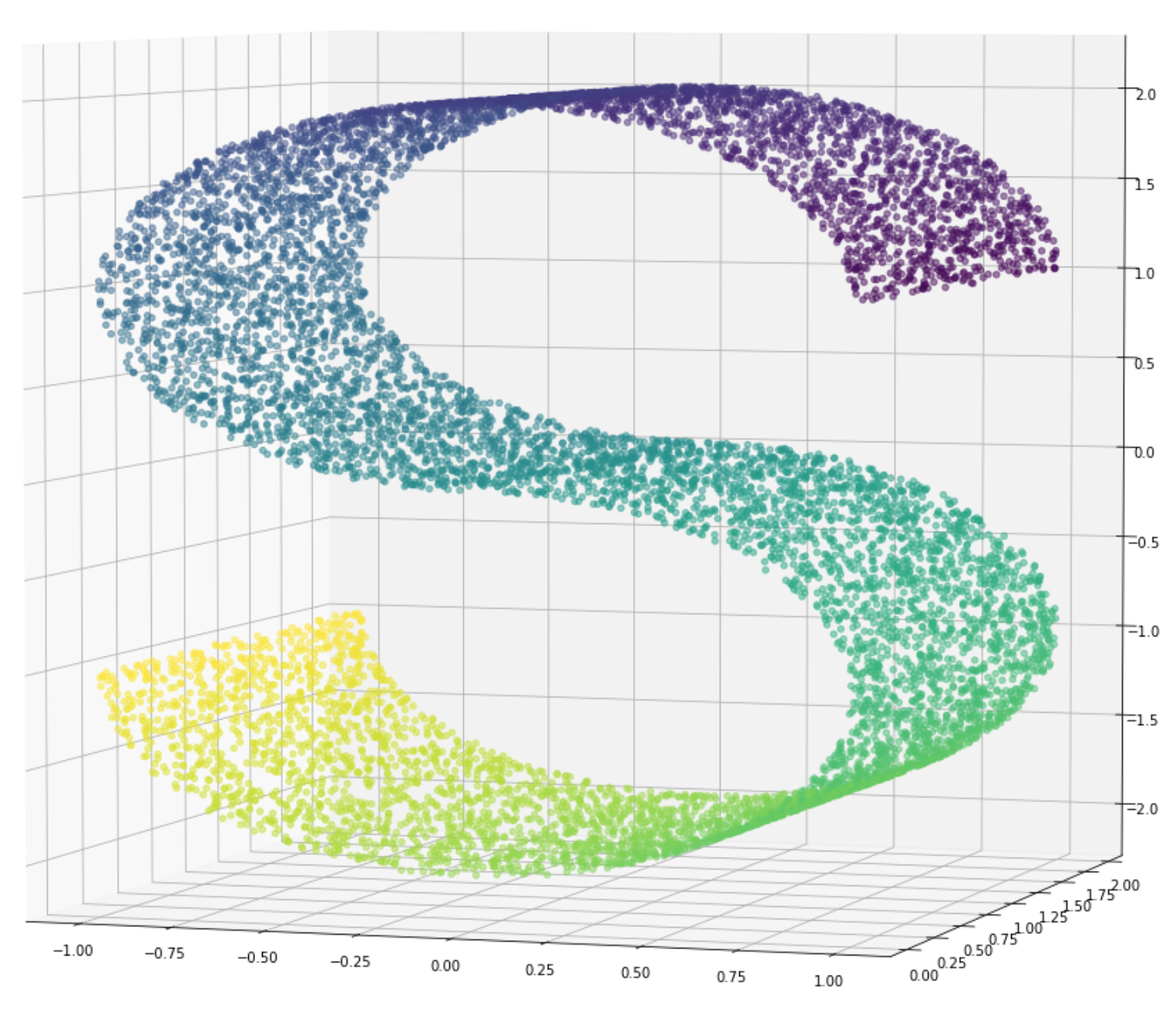}
        \caption{S-manifold prior.  A uniform 2d distribution is wrapped over the S-manifold in 3d.  Colors shown here are chosen to highlight the shape of the manifold only.}
        \label{fig:ACCA_S_Prior}
    \end{subfigure}
    \begin{subfigure}[t]{0.475\textwidth}  
        \centering 
        \includegraphics[width=\textwidth]{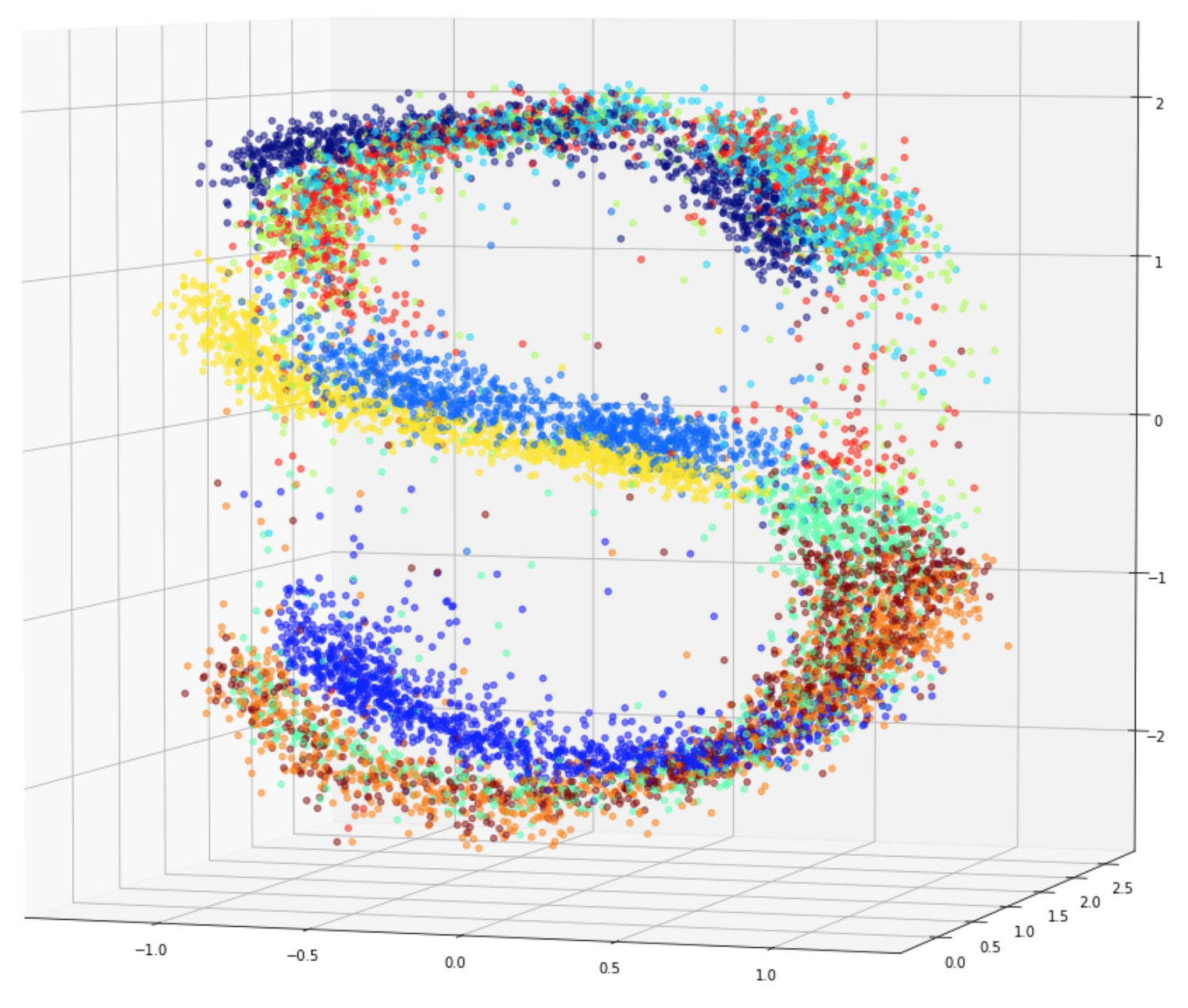}
        \caption{Embeddings from ACCA colored by class.}
        \label{fig:ACCA_S_Class}
    \end{subfigure}
    \begin{subfigure}[t]{0.475\textwidth}  
        \centering 
        \includegraphics[width=\textwidth]{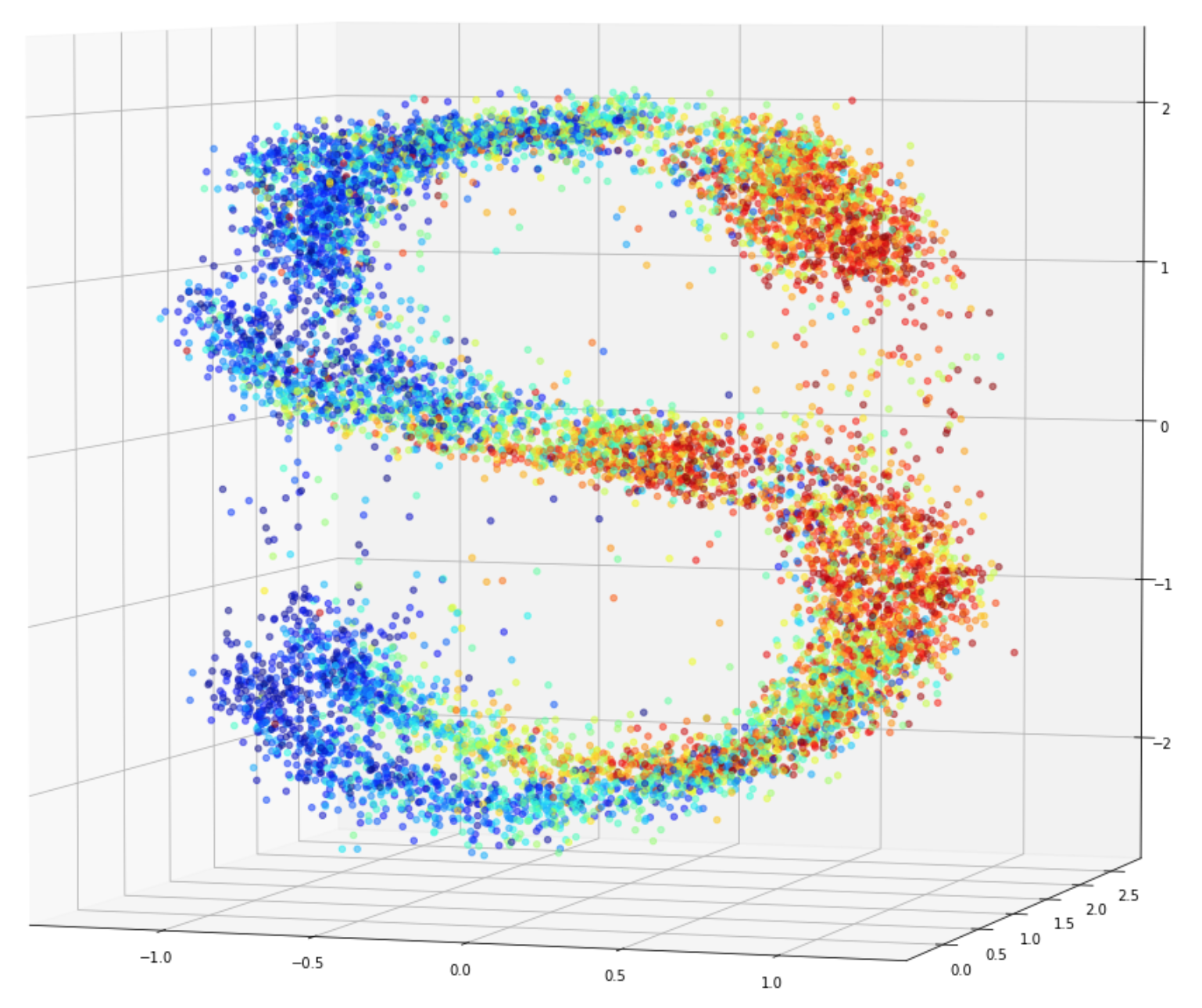}
        \caption{Embeddings from ACCA colored by $\texttt{rot}_x$}
        \label{fig:ACCA_S_rotx}
    \end{subfigure}
    \begin{subfigure}[t]{0.475\textwidth}  
        \centering 
        \includegraphics[width=\textwidth]{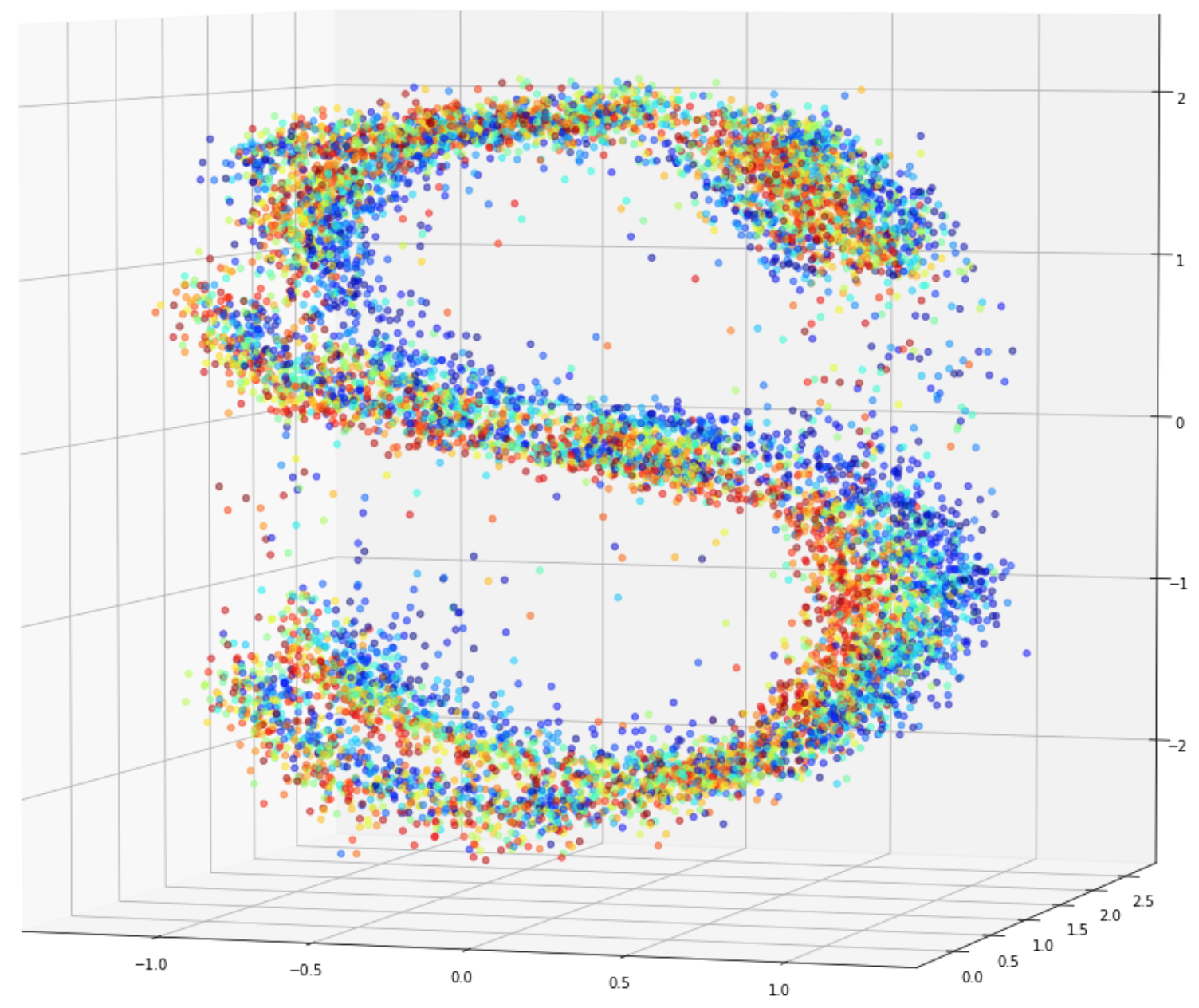}
        \caption{Embeddings from ACCA colored by $\texttt{rot}_x$}
        \label{fig:ACCA_S_roty}
    \end{subfigure}
    \caption{ACCA allows arbitrary priors ($p(z)$) to be chosen, setting a basis for new research directions in multiview representation learning. To illustrate this, we construct a prior by wrapping a uniform 2d distribution over an S-manifold (a). The embedding from ACCA is shown in (c-d) with colorings corresponding to different underlying factors of variation.}
\end{figure}

\begin{figure}[ht]
    \centering
    \begin{subfigure}[t]{0.975\textwidth}
        \centering
        \includegraphics[width=\textwidth]{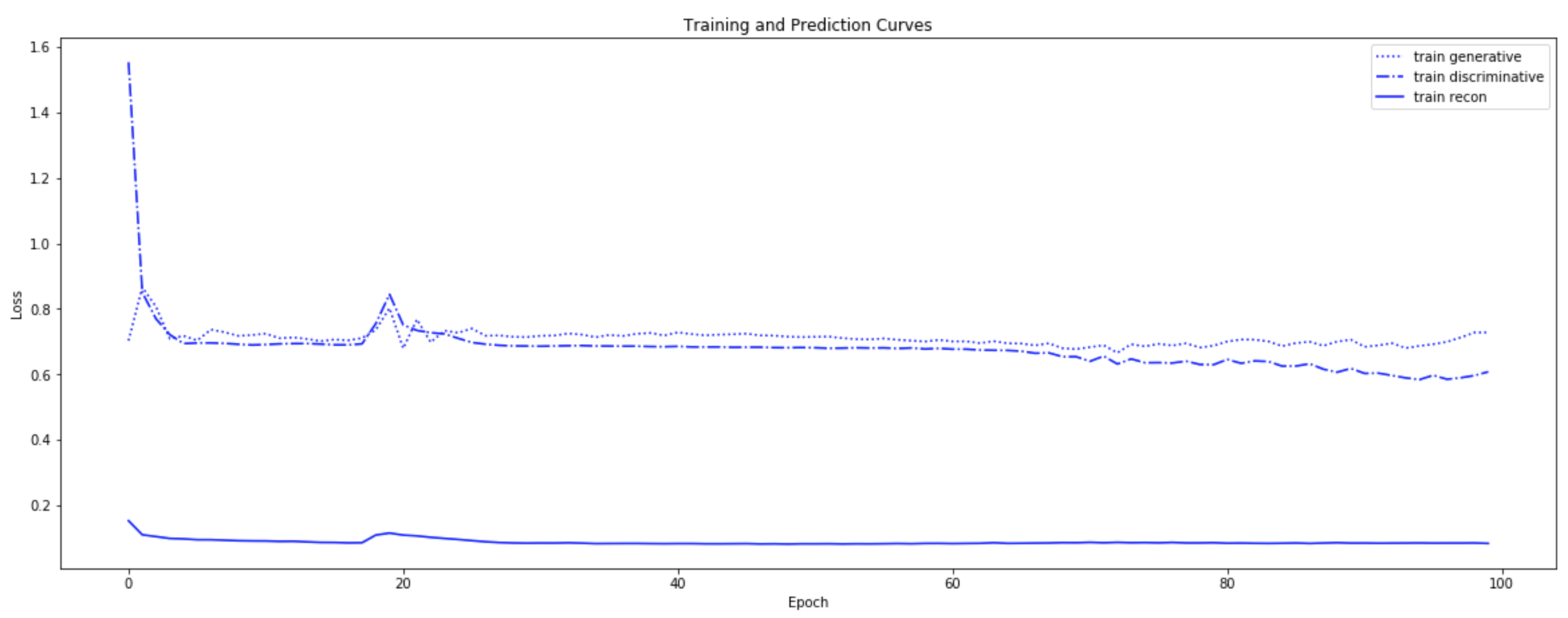}
        \caption{Loss curves}
        \label{fig:ACCA_S_Loss}
    \end{subfigure}
    \begin{subfigure}[t]{0.975\textwidth}  
        \centering 
        \includegraphics[width=\textwidth]{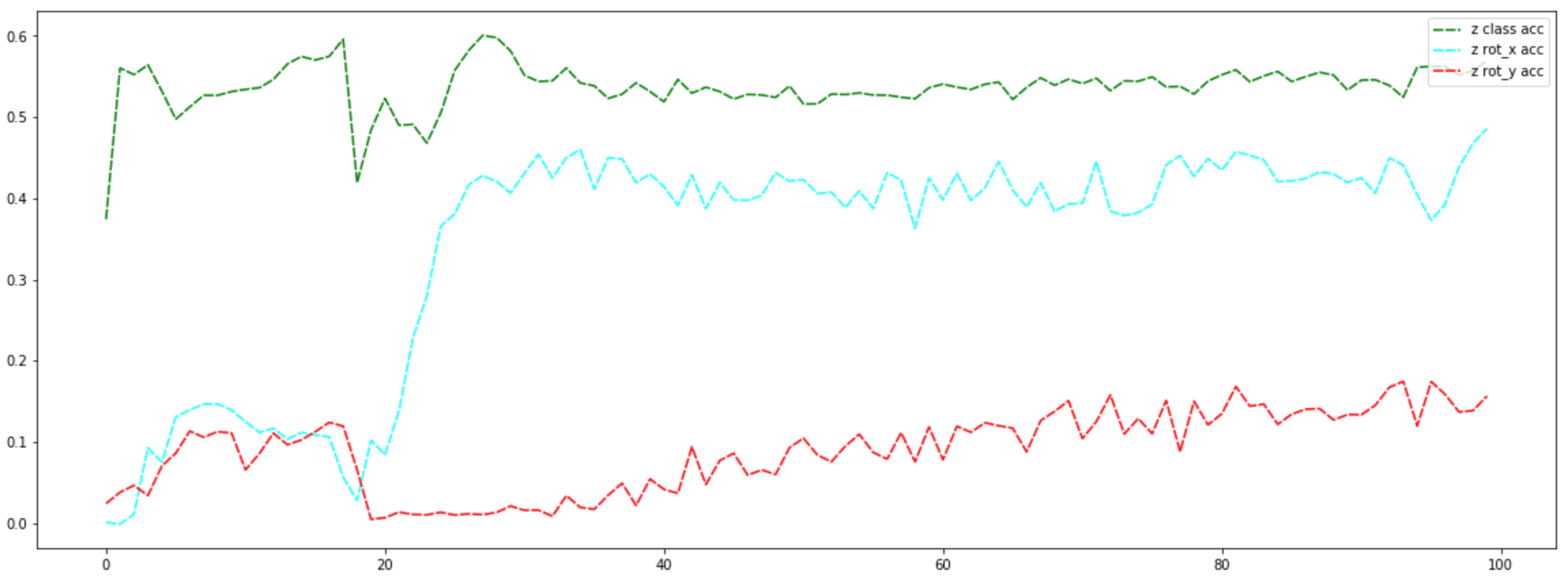}
        \caption{Information curves}
        \label{fig:ACCA_S_Info}
    \end{subfigure}
    \caption{Loss (a) and information (b) curves for ACCA with $z$-dim=3, trained on Tangled MNIST for 100 epochs with the S-manifold prior.}
    \label{fig:53_LossInformation}
\end{figure}

\begin{figure}[ht]
    \centering
    \begin{subfigure}[t]{\textwidth}
        \centering
        \includegraphics[width=\textwidth]{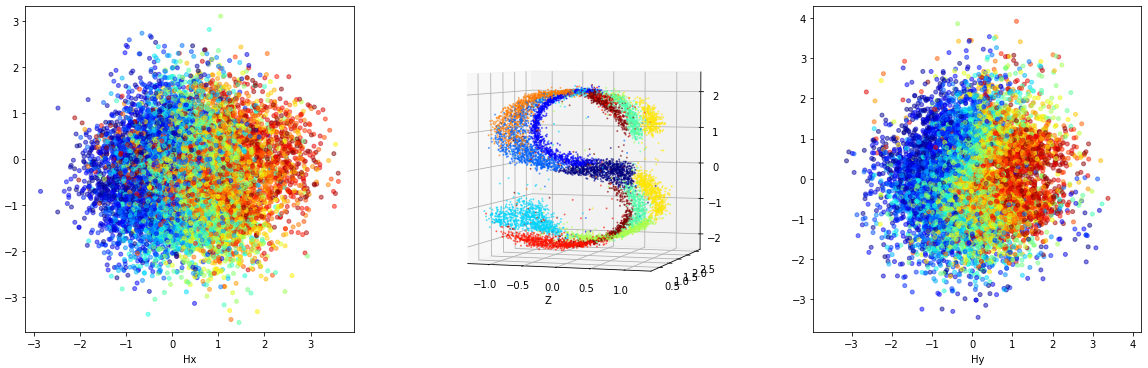}
    \end{subfigure}
    \caption{Arbitrary priors can be used independently for individual representations. Here we use the S-manifold prior on $z$ alone with $\mathcal{N}(0,I)$ on $h_x$ and $h_y$, trained on Tangled MNIST and colored by the desired information content: $\texttt{rot}_x$ for $h_x$, class for $z$, and $\texttt{rot}_y$ for $h_y$.}
    \label{fig:ACCA_Private_S_Embeddings}
\end{figure}